\documentclass[10pt]{article} 

\usepackage[utf8]{inputenc}
\usepackage{amsmath}
\usepackage{amsfonts}
\usepackage{amssymb}
\usepackage{mathtools}
\usepackage{listings}
\usepackage{mathrsfs}
\usepackage{times}
\usepackage{float}
\usepackage{color}
\usepackage{setspace}
\usepackage{color,soul}

\usepackage{amsmath,amsfonts,amsthm,eucal}
\usepackage{enumerate}
\usepackage[letterpaper,margin=3cm]{geometry}
\usepackage{float}
\usepackage[title]{appendix}
\usepackage{color}
\usepackage{tabularx}
\usepackage{siunitx}
\usepackage[tableposition=below]{caption}
\usepackage{array}

\usepackage[
pdfstartview=XYZ,
bookmarks=true,
colorlinks=true,
linkcolor=blue,
urlcolor=blue,
citecolor=blue,
pdftex,
bookmarks=true,
linktocpage=true,   
hyperindex=true
]{hyperref}

\usepackage{lipsum}
\usepackage{lineno}

\usepackage[FIGBOTCAP,TABTOPCAP,bf,tight]{subfigure}

\usepackage{natbib}
\setcitestyle{authoryear}

\usepackage[pdftex]{graphicx}
\usepackage{epstopdf}
\usepackage{booktabs} 
\usepackage{tikz,mathpazo}
\usetikzlibrary{shapes.geometric, arrows}
\usepackage{caption}

\newcommand{\tensor}[1]{\ensuremath{\boldsymbol{#1}}}

\usepackage{bm}

\usepackage{algorithm}
\usepackage{algorithmicx}
\usepackage[noend]{algpseudocode}

\theoremstyle{remark}

\newtheorem{remark}{Remark}
\newtheorem{theorem}{Theorem}

\renewcommand{\vec}[1]{\ensuremath{\boldsymbol{#1}}}
\theoremstyle{definition}
\newtheorem{definition}{Definition}

\usepackage{algorithm}
\usepackage[noend]{algpseudocode}
\newcolumntype{M}[1]{>{\centering\arraybackslash}m{#1}}

\newcommand{\mV} {{\bm V}}  
\newcommand{\bomega} {{\bm \omega}}  
\newcommand{\bX} {{\bm X}}  
\newcommand{\bY} {{\bm Y}}  
\newcommand{\mE} {\bm E}   
\newcommand{\vU} {\vec{U}} 
\newcommand{\bx} {{\bm x}}  
\newcommand{\bz} {{\bm z}}   
\newcommand{\bi} {{\bm i}}  
\newcommand{\bW} {{\bm W}}  
\newcommand{\bU} {{\bm U}}  
\newcommand{\bb} {{\bm b}}  
\newcommand{\bh} {{\bm h}}  
\newcommand{\br} {{\bm r}}  
\newcommand{\by} {{\bm y}}

\newcommand{\indep}{\rotatebox[origin=c]{90}{$\models$}}

\makeatletter
\let\@fnsymbol\@arabic
\makeatother

\date{}

\title{Data-driven discovery of interpretable causal relations for deep learning material laws with uncertainty propagation} 
\begin{document}

\author{
Xiao Sun \footnote{Department of Applied Mathematics and Statistics, Johns Hopkins University, Baltimore, MD.} 
\and   Bahador Bahmani \footnote{Department of Civil Engineering and Engineering Mechanics, Columbia University, New York, NY.}
\and Nikolaos N. Vlassis $^2$
    \and  WaiChing Sun  $^2$ \thanks{Corresponding author: \textit{wsun@columbia.edu} }
    \and Yanxun Xu  $^1$ \thanks{Corresponding author: \textit{yanxun.xu@jhu.edu} }
}

\maketitle

\begin{abstract}
This paper presents a computational framework that generates ensemble predictive mechanics models 
with uncertainty quantification (UQ). 
We first develop a causal discovery algorithm to infer causal relations among time-history data measured during each representative volume element (RVE) simulation through a directed acyclic graph (DAG). With multiple plausible sets of causal relationships estimated from multiple RVE simulations, the predictions are propagated in the derived causal graph while using a deep neural network equipped 
with dropout layers as a Bayesian approximation for uncertainty quantification. 
We select two representative numerical examples (traction-separation laws for frictional interfaces, elastoplasticity models for granular assembles) to examine the accuracy and robustness of the proposed 
causal discovery method for the common material law predictions in civil engineering applications.
\end{abstract}

\section{Introduction}
\label{intro}

Computer simulations for mechanics problems often require 
material (constitutive) laws that replicate the local constitutive responses 
of the materials. These material laws can be used to replicate
the responses of an interface 
(e.g., traction-separation laws or cohesive zone models)
or bulk materials (e.g., elastoplasticity models for solids, porosity-permeability relationship and water retention curve). 
A  computer model 
is then completed by incorporating these local 
constitutive laws into a discretized form of balance principles 
(balance of mass, linear momentum and energy) 
where discretized numerical solutions can be sought by a proper solver.

 Constitutive laws, such as stress-strain relationship for bulk materials, 
 traction-separation laws for interface, porosity-permeability for porous media, 
 are often derived following a set of axioms and rules \citep{truesdell2004non}.  
 In these hand-crafted models, 
 phenomenological observations are incorporated into constitutive laws (e.g., critical state theory for soil mechanics \citep{schofield1968critical, sun2013unified, bryant2019micromorphically, na2019configurational}, 
 void-growth theory for ductile damage \citep{gurson1977continuum}). 
 While the earlier simpler models are often amended by 
 newer and more comprehensive models \citep{dafalias1984modelling}
 in order to improve the performance (e.g. accuracy, more realistic interpretation of mechanisms), 
 these improvements are often a trade-off that may unavoidably increase 
 the number of parameters, leading to increasing difficulty for 
 the calibration, verification and validation processes, as well as the uncertainty  quantification \citep{dafalias1984modelling, borja2007estimating, borja2008coseismic,  clement2013uncertainty, wang2019meta, wang2016identifying}. 
 
 The rise of  big data and the great promises of machine learning have led to a new generation of approaches that either bypass the usages of constitutive laws via model-free data-driven methods (e.g., \citet{kirchdoerfer2016data, he2020physics, bahmani2021kd, karapiperis2021data}) or replace parts of the modeling efforts/components with 
 models generated from supervised learning 
 (e.g., \citet{furukawa1998implicit, lefik2003artificial, wang2018multiscale, wang2019updated, zhang2020ai, vlassis2020geometric, vlassis2021sobolev, logarzo2021smart, masi2021thermodynamics}). 
 However, one critical issue of these machine learning and data-driven approaches is the lack of  sufficient 
 interpretability of predictions. 
 While there is no universally accepted definition of interpretability, 
 we will herein employ the definition used in \citet{miller2019explanation} 
 which refers interpretability as the degree to which a human can understand the cause of the prediction. 
 
 One possible way to boost the interpretability is to introduce a proper medium to represent knowledge  
 that can be understood by human \citep{sussillo2013opening}. 
 Causal graph, also known as causal Bayesian network,
 is one such  medium in which the causal relations among different 
 entities are mathematically represented by a directed graph.  
 In the application of computational mechanics 
 \cite{wang2018multiscale, wang2019updated, heider2020so} have derived a mathematical framework 
 to decompose a complex prediction task into multiple easier predictions represented by subgraphs within graphs. 
 More recent work such as \citet{wang2019meta, wang2019cooperative} 
 introduce a deep reinforcement learning approach that employs the Monte Carlo Tree Search (MCTS) to assemble a directed graph 
that generates a sequence of interconnected predictions of physical quantities to emulate a hand-crafted constitutive law.  
However, these directed graphs are generated to optimize a given performance metric (e.g. accuracy, calculation speed and forward prediction robustness), but not necessarily reveal the underlying causal relations among physical quantities. 
 
 Discovering causal relations from observational data is an important problem with applications in many fields of science, such as social science  \citep{oktay2010causal}, finance \citep{gong2017causal}, and biomedicine \citep{shen2020challenges}. The standard way to discover causality is through randomized controlled experiments. However, conducting such experiments can be either impractical, unethical, and/or very expensive in many disciplines \citep{xu2012bayesian, xie2020bayesian}.
For mechanics problems, the major issues include the time and labor cost for physical experiments, the lack of facilities or equipment to complete the required tests and the difficulties to obtain specimens 
\citep{mitchell2005fundamentals, powrie2018soil, wood1990soil}. As a result, an alternative approach, which is adopted in this study, is to use sub-scale simulations as the digital representation that generates auxiliary 
data sets to build material laws or forecast engine for the macroscopic 
material responses \citep{liu2016determining, ma2020computational, frankel2019predicting, wang2019updated}.
Classic methods for causal discovery are based on probabilistic graphical modeling \citep{pearl2000causality}, the structure of which is a  directed acyclic graph (DAG) with nodes representing random variables and edges representing conditional dependencies between variables. 
 Learning a DAG from observational data is highly challenging since the number of possible DAGs is super-exponential to the number of nodes. There are two main approaches for causal discovery: the constraint-based approach and the score-based approach.  The constraint-based approach  aims to  recover a Markov equivalence class through inferring conditional independence relationships among the variables, and the resulting Markov equivalence class may contain multiple DAGs that indicate the same conditional independence relationships \citep{le2016fast, cui2016copula}. On the other hand, the score-based approach uses a scoring function, such as the Bayesian Information Criterion (BIC), to search for the DAG that best fits the data \citep{heckerman1995learning, huang2018generalized}.  
 
 In this paper, we aim to discover causal relations that can explain the underlying mechanism of a history-dependent macroscopic constitutive law upscaled from direct numerical simulations at the meso-scale. 
 The most common method for constructing the causal relations from time-series data is Granger causality \citep{granger1969}, which assumes a number of lagged effects and analyzes the data in a unit no more than that number of lags. See \cite{runge2018causal} for a review of causal discovery methods on time-series data. 
However, most of these causal discovery methods assume that the data are generated from a stationary process, meaning that the data are generated by a distribution that does not change with time. Such an assumption does not hold in many  physical processes, in which the mechanisms or parameters in the causal model may change over time.   Several methods have been proposed recently to tackle time-varying causal relations in non-stationary processes  \citep{ghassami2018multi, huang2019causal, huang2020causal}. However, they either assume linear causal models, or does not offer the flexibility of incorporating known physical knowledge, limiting their applicability to the nonlinear path-dependent relations in learning material constitutive laws.  
 
 In this work, we offer two major innovations. First, we introduce a new decoupled discovery/training approach where the discovery of causal relations represented by a DAG is enabled by 
 a causal discovery algorithm  that deduces plausible causal relations from non-stationary time series data and incorporates known physical knowledge. Second, we leverage the obtained causal graph as the representation of mechanics knowledge and 
 adopt a Bayesian approximation using the dropout layer technique first introduced by \citet{gal2016dropout} 
 to propagate 
 epistemic uncertainty in the causal graph and generate quantitative predictions with uncertainty quantification.

 The rest of the paper is organized as follows. Section \ref{sec:constitutive}  first introduces the two data sets (learning traction-separation law and hypo-plasticity of granular materials) used for our numerical experiments.  
 This is followed by the description of theory and implementation of the proposed causal discovery algorithm used to deduce the causal relations from non-stationary time series data (Section \ref{sec:discovery}). The setup of the deep neural network model for the prediction tasks and the uncertainty propagation   are included in Sections  \ref{sec:neuralnetworktraining} and \ref{sec:uncertainty} respectively. The proposed framework 
 is then tested against two numerical experiments (Section \ref{sec:examples}), which is then followed by the conclusion.

\section{Causal relations and constitutive laws} \label{sec:constitutive}

As demonstrated in previous studies such as \citet{wang2018multiscale, wang2019meta, wang2019updated, wang2019cooperative, heider2020so, vlassis2020geometric}, 
the  relationships in a constitutive model can be represented by 
 a network of unidirectional information flow, i.e., a DAG $G=(\mV, \mE)$ where $\mV$ represents a vertex set and $\mE$ denotes an edge set.  With appropriate assumptions that will be discussed later,  the DAG can be identified as a causal graph \citep{pearl2000causality}.  
The causal relations are not only useful to explain the underlying mechanism of a process but also provide us a basis to formulate multi-step transfer learning to
predict constitutive responses. 
This strategy can be beneficial because one can leverage more data gathered from physical numerical experiments to train the prediction model. For instance,
while a black-box prediction of stress-strain curves only leverages the stress-strain pair for supervised learning, the introduction of knowledge graphs may introduce multiple supervised learning tasks where measurements of porosity, fabric tensors or any other physical and geometrical attributes measured during the experiments can be leveraged to improve the training. For completeness, we briefly describe the procedure to consider the data set as vertex sets in graphs and the causal discovery process used to create 
the directed edge set in a knowledge graph through two examples. 

Note that many of the physical quantities that become the vertices in the knowledge graphs are 
graph metrics obtained from analyzing the connectivity topology of the granular system. 
For brevity, we will not provide a review on the applications of graph theory for granular matter here. 
Interested readers may refer to Appendix \ref{sec:graph_metrics} for the definitions of the graph metrics and \citet{satake1978constitution, bagi1996stress, walker2010topological, tordesillas2010force, o2011particulate, kuhn2015stress} and \citet{vlassis2020geometric}
 for reviews on the graph theory applied to particulate and granular systems. 
 
\subsection{Dataset for traction-separation law}
In the first example, our goal is to conduct a numerical experiment 
to verify whether the causal discovery algorithm is able to 
re-discover the well-known causal relation that links the 
plastic dilatancy and contraction to the  
frictional behaviors \citep{scholz1998earthquakes, popov2010contact}
with a small data set. 

Following \citet{wang2018multiscale, wang2019meta, wang2019updated, wang2019cooperative}, we consider the vertex set consists of five elements, 
 the displacement jump/separation $\vec{U}$, the traction $\vec{T}$, and 
 three geometric measures, i.e.,
\begin{enumerate}
\item Displacement jump $\vec{U}$, the relative displacement of an interface of two in-contact bodies. 
\item Porosity $\phi$, the ratio between the volume of the void and the total volume of RVE.
\item Coordination number (averaged)  $CN=N_{\text{contact}}/N_{\text{particle}}$ where $N_{\text{contact}}$ is the number of particle contacts and $N_{particle}$ is the number of particles in the RVE.
\item Fabric tensor $\mathbf{A}_f={\frac{1}{N_{\text{contact}}}}\sum_{c=1}^{N_{\text{contact}}}\mathbf{n}^c\otimes\mathbf{n}^c$ where $\mathbf{n}^c$ is the normal vector of a particle contact $c$ in the RVE.  The symbol `$\otimes$' denotes a juxtaposition of two vectors 
(e.g.,\ $\vec{a} \otimes \vec{b} = a_{i}b_{j}$)
or two symmetric second order tensors 
[e.g.,\ $(\tensor{\alpha} \otimes \tensor{\beta})_{ijkl} = \alpha_{ij}\beta_{kl}$]. 
\item Traction $\vec{T}$, the traction vector acts on the interface. 
\end{enumerate}

To generate a machine learning based traction-separation law, we identify the displacement jump as the root and the traction as the leaf of the causal graph. The causal graph is a DAG $G=(\mV, \mE)$ where 
$\mV$ is the set consisting of the physical quantities $\vec{U}, \phi, CN, \tensor{A}_{f}$ and $\vec{T}$. Meanwhile, $\mE \subseteq \mV\times \mV$ is 
 a set of directed edges that connect any two elements from $\mV$, and $\mE$ is 
 determined from the causal discovery  algorithm outlined in Section \ref{sec:discovery}. 

The dataset is generated using an open-source code YADE. In total, there are  100 traction-separation law simulations run with different loading paths performed on the same RVE. This RVE consists of spherical particles with radii between $1\pm0.3$ mm with uniform distribution. The RVE has a height of 20 mm in the normal direction of the frictional surface and is initially consolidated to an isotropic pressure of 10 MPa. The inter-particle interaction is controlled by Cundall's elastic-frictional contact model \citep{cundall1979discrete} with an inter-particle elastic modulus of $E_{eq} = 1 \, GPa$, a ratio between shear and normal stiffness of $k_s / k_n = 0.3$, a frictional angle of $\phi = 30 ^{\circ}$, a density $\rho = 2600 \, kg/m^3$, and a Cundall damping coefficient $\alpha_{\text{damp}} = 0.2$.
For brevity, the generation and setup of the simulations are not included in this paper. Interested readers please refer to \citet{wang2019meta} for more information. The data required to replicate the results of this paper and for 3rd-party validation can be found in the Mendeley Data repository \citep{wang2019discrete}. 

\subsection{Dataset for hypo-plasticity of granular materials}

While the first data set is used to determine whether the causal graph 
algorithm may re-discover known physical relations in the literature, the second problem is designed to test whether 
the causal graph algorithm may successfully investigate new plausible causal relations not known {\it a prior} in the literature. 

For this purpose, we run 60 discrete element simulations and use 30 of them for calibrations and 30 for blind forward predictions. 
In addition to the conventional microstructural attributes (e.g., porosity and fabric tensor) typically used for hand-crafted constitutive laws \citep{manzari1997critical, dafalias2004simple, sun2013unified, bryant2019micromorphically, na2019configurational}, 
we have also recorded the evolution of the particle contact pairs 
in each incremental time step of the discrete element simulations. 
The particle contact connectivity is itself an undirected graph 
$G_{\text{contact}} = (\mV_{\text{particle}}, \mE_{\text{contact}})$ 
where $\mV_{\text{particle}}$ is the set of particles and 
$\mE_{\text{contact}}$ is the set of particle contacts, 
one for each contact between two contacting convex particle represented.  They are undirected edges. 
To facilitate new discovery, we compute 15 different graph
metrics of $G_{\text{contact}}$ (see Appendix \ref{sec:graph_metrics} for definition) that have not been used for composing constitutive laws and see if (1) whether the causal discovery algorithm may discover causal  relations among these new physical quantities and (2) whether the new discovery helps improve the accuracy, robustness and consistency of the forward predictions 
enabled by neural networks trained according to the discovered causal relations. 

In total, there are 11 types of time-history data in which 3 of them are second-order tensors (strain, stress and the strong fabric tensor), and the rest are scalar (porosity, coordination number graph density, graph local efficiency, graph average clustering, graph degree assortativity coefficient, graph transitivity and graph clique number). As such there are 11 elements in the vertex set and the goal of the causal discovery is to establish the edge set to complete the causal graph. A sequence of supervised learning is then used to generate predictions via deep learning.

\section{Causal discovery and knowledge graph constructions}  
\label{sec:discovery}

\subsection{Notations and assumptions}
\label{sec:assumption}
Let $G=(\mV, \mE)$ be a DAG containing only directed edges and has no directed cycles. For each $V_i\in \mV$, let $\mathrm{PA}^{i}$ denotes the set of parents of $V_i$ in $G$. Since our data are time-history dependent, we assume that the joint probability distribution of $\mV$ at each time point according to $G$ can factorize as $p(\mV) = \prod_{i=1}^m p(V_i\mid \mathrm{PA}^{i}), $ 
where $m$ is the number of vertices in $G$. Here $p(V_i\mid \mathrm{PA}^{i})$ can be regarded as ``causal mechanism."  For non-stationary time series data, the causal mechanism $p(V_i\mid \mathrm{PA}^{i})$ can change over time, and the changes may be due to the involved functional models or the causal strengths.   

 Throughout this section, we use the example of traction-separation law to illustrate the proposed causal discovery method without loss of generality. Therefore, $\mV$ is the set consisting of displacement jump $\vec{U}$, porosity $\phi$, coordination number $CN$, fabric tensor $\tensor{A}_{f}$,  and  traction $\vec{T}$.   \cite{Zhangkun2019} developed a constraint-based causal discovery algorithm for non-stationary time series data to identify changing causal modules and recover the causal structure. In this paper, we extend the algorithm in \cite{Zhangkun2019} such that the proposed causal discovery algorithm not only handles non-stationary time-history data but also incorporates certain physical constraints. For example, in constructing the traction-separation law, we have the prior knowledge that the dynamic changes in $\vec{U}$ can cause changes in other variables, not vice versa. Therefore, if there exists a directed edge between the displacement jump $\vec{U}$ and any other variable $V_i$, then $\vec{U} \rightarrow{V_{i}}$. 

Denote $\mV_{-\vec{U}}$  to include all other variables in $\mV$ excluding $\vec{U}$ (e.g., porosity, fabric tensor). Since the causal mechanism can change over time,  we assume that the changes can be explained by certain time-varying confounders, which can be written as functions of time. 
As we have the prior knowledge that $\vec{U}$ itself is a time-dependent variable and could affect all other variables,  we regard $\vec{U}$ as such a confounder and assume  that the causal relation for each $V_i\in \mV_{-\vec{U}}$ can be represented by the following structural equation model:
\begin{eqnarray}
V_{i} &=& g_{i}(\mathrm{PA}^{i}, \theta_{i}(\vec{U}),\epsilon_{i}), 
\label{eq:eq1}
\end{eqnarray}
where $\mathrm{PA}^{i}$ includes $\vec{U}$ if the changes in $\vec{U}$ can affect the changes in $V_i$, $\theta_{i}(\vec{U})$ denotes a function of $\vec{U}$ that influences $V_i$ as effective parameters,  $\epsilon_{i}$ is a noise term that is independent of $\vec{U}$ and $\mathrm{PA}^{i}$. The $\epsilon_{i}$'s are  assumed to be independent. As we treat $\vec{U}$ as a random variable, there is a joint distribution over $\mV \cup \{ \theta_i(\vec{U})\}_{i: V_i\in \mV_{-\vec{U}}}$.
Denote $G^{aug}$ to be the graph by adding $\{ \theta_i(\vec{U})\}_{i: V_i\in \mV_{-\vec{U}}}$ to $G$, and for each $i$, adding an arrow from $\theta_i(\vec{U})$ to $V_i$.  Note that $G$ is the induced subgraph of $G^{aug}$ over $\mV$. Denote the joint distribution of $G^{aug}$ to be $p^{aug}$.

In order to apply any conditional independence test on the variable set $\mV$ for recovering   causal structure, we set the following assumptions  \citep{spirtes2000causation}. 

\noindent {\bf Assumption 1.} (Causal Markov condition) $G^{aug}$ and the joint distribution $p^{aug}$ on $\mV \cup \{ \theta_i(\vec{U})\}_{i: V_i\in \mV_{-\vec{U}}}$ satisfy the causal Markov condition if and only if a vertex of $G^{aug}$ is probabilistically independent of all its non-descendants in $G^{aug}$ given the set of all its parents. 

\noindent {\bf Assumption 2.} (Faithfulness) $G^{aug}$ and the joint distribution $p^{aug}$  satisfy the faithfulness condition if and only if no conditional independence holds unless entailed by the causal Markov condition.

\noindent {\bf Assumption 3.} (Causal sufficiency) The common causes of all variables in $\mV \cup \{ \theta_i(\vec{U})\}_{i: V_i\in \mV_{-\vec{U}}}$  are measured. 

\subsection{Recovery of the causal skeleton}
In this section, we propose a constraint-based method building upon the PC algorithm \citep{spirtes2000causation} to first identify the skeleton of $G$, defined as the obtained undirected graph if we ignore the directions of edges in a DAG $G$. We prove that given   Assumptions 1-3, we can apply conditional independence tests to $\mV$ to recover the skeleton of $G$. Algorithm 1 describes the proposed method, which is supported by Theorem \ref{th1}. The proof is provided in Appendix \ref{sec:proof} following \cite{Zhangkun2019}.

 \begin{algorithm}\label{algo:algo1}
	\begin{algorithmic}[1]
		\State  Object: To obtain the undirected skeleton of $G$  
		\State  Build a complete undirected graph $U_{G}$ with variables $\mV$ 
		\For{each node $V_{i} \in \mV_{-\vec{U}}$}
		\If{$V_{i}$ and $\vec{U}$ are independent given a subset of $\{V_{k} | V_{k}\in\mV_{-\vec{U}}, k \neq i\}$} 
		\State Remove the edge between $V_i$ and $\vec{U}$
		\EndIf
		\EndFor
		\For{every $V_i, V_j \in \mV_{-\vec{U}}$ }
		\If{ $V_{i}$ and  $V_{j}$ are independent given a subset of $\{V_{k} \mid  V_{k}\in\mV_{-\vec{U}}, k \neq i,k \neq j\}\cup{\vec{U}} $}
		\State Remove the edge between $V_{i}$ and $V_{j}$ 
		\EndIf
		\EndFor
		\State \Return $U_{G}$
		\caption{Obtain the undirected skeleton of $G$ }		
	\end{algorithmic}
\end{algorithm}

\begin{theorem}\label{th1}
	Given Assumptions 1-3, for every $V_i, V_j\in \mV_{-\vec{U}}$, $V_i$ and $V_j$ are not adjacent in $G$ if and only if they are independent conditional on some subset of $\{V_k\mid V_k\in \mV_{-\vec{U}}, k\neq i, k\neq j \}\cup \{\vec{U}\}$. 
\end{theorem}
 
In lines 3-7 of Algorithm 1, we determine whether the changes in $\vec{U}$ cause changes in $V_i$. If not, $\vec{U}$ is not in the parent set of $V_i$ and there is no edge between $\vec{U}$ and $V_i$ in $G$. The lines 8-12 of Algorithm 1 aims to identify the causal skeleton between variables in $\mV$ except $\vec{U}$. 
Since how other variables change with $\vec{U}$ and the relations between these variables are usually unknown and potentially very complex, we use a nonparametric conditional independence test, kernel-based condition independence (KCI) test developed by \cite{zhang2012kernel},  to determine the dependence between variables throughout this paper.  This nonparametric approach can not only capture the linear/nonlinear correlations between variables by testing for zero Hilbert-Schmidt norm of the partial cross-covariance operator, but also handle multidimensional data that are common in mechanics  problems. 
 
\subsection{Determination of causal directions}
After obtaining the skeleton $U_G$, we need to determine the causal directions of edges. \cite{meek1995causal} provided a set of orientation rules to determine the directions of undirected edges in a graph based on conditional independence tests.  However, the Meek rule \citep{meek1995causal} is only applicable to edges that satisfy its conditions.   In this section, we first introduce the Meek rule \citep{meek1995causal}, then propose an algorithm to orient the edges that are not covered by the Meek rule after incorporating known physical knowledge. 

Denote $\leftrightarrow$ to be an undirected edge. The Meek rule has the following principles:
\begin{enumerate}
\item For all triples $V_{i} \leftrightarrow{V_{j}}\leftrightarrow{V_{k}}$, if $V_i$ and $V_k$ are marginally independent but conditionally dependent given $V_j$, then $V_{i} \rightarrow V_{j} \leftarrow V_{k}$;
\item If  $V_{i} \rightarrow{V_{j}}\leftrightarrow{V_{k}}$ and there is no edge between $V_i$ and $V_k$, then orient $V_j \rightarrow{V_{k}}$; 
\item  If  $V_{i} \rightarrow{V_{j}}\leftrightarrow{V_{k}}$ and there is an edge between $V_i$ and $V_k$, then orient $V_i \rightarrow{V_{k}}$; 
\item  If $V_{i} \rightarrow V_{j} \leftarrow V_{k}$, $V_{i} \leftrightarrow V_{k} \leftrightarrow V_{k}$, and $V_k\leftrightarrow V_j$, then $V_k\rightarrow V_j$. 
\end{enumerate}

Now we describe our algorithm on how to determine the edge directions in the obtained skeleton $U_G$. Firstly, for any node $V_i$ adjacent to $\vU$, we orient $\vU\rightarrow V_i$ due to the prior physical knowledge that only $\vU$ affects other variables, not vice versa. Then we apply the Meek rule to the obtained graph after orienting the edges from $\vU$ to its neighbours. For instance, suppose $\vU \rightarrow V_i \leftrightarrow V_j$, if $V_j$ and $\vU$ are independent given a set of variables including $V_i$, then we orient $ V_i \rightarrow V_j$; if $V_j$ and $\vU$ are independent given a set of variables excluding $V_i$, then we have $V_j \rightarrow V_i$. 

Next, we discuss how to determine the edge direction between two adjacent variables if they are both adjacent to $\vU$, i.e., $V_i\leftrightarrow V_j$,  $\vU \rightarrow V_i$, and  $\vU \rightarrow V_j$, since such a scenario is not covered by the Meek rule. The modularity property of causal systems \citep{pearl2000causality} demonstrated that if there are no confounders for {\it cause} and {\it effect}, then $p(\mathrm{\it cause})$ and $p(\mathrm{\it effect}\mid \mathrm{\it cause})$ are either fixed or change independently. Based on this principle, since both $V_i$ and $V_j$ change with $\vU$, we can test the conditional independence between $p(V_i \mid \theta_i(\vU))$ and $p(V_j\mid V_i,  \theta_j(\vU))$, as well as between $p(V_j \mid \theta_j(\vU))$ and $p(V_i\mid V_j, \theta_i(\vU))$ to determine the direction between $V_i$ and $V_j$. That says, if $p(V_i \mid \theta_i(\vU))$ and $p(V_j\mid V_i,  \theta_j(\vU))$ are conditionally independent but $p(V_j \mid \theta_j(\vU))$ and $p(V_i\mid V_j, \theta_i(\vU))$ are not, then $V_i\rightarrow V_j$.  
 \cite{Zhangkun2019} developed a kernel embedding of non-stationary conditional distributions and extended the Hilbert Schmidt Independence Criterion (HSIC, \citet{gretton2008kernel}) to measure the dependence between distributions, based on which the causal directions can be determined. For example,  if we have two random variables $V_1$ and $V_2$, we can compute the dependence between $p(V_1)$ and $p(V_2\mid V_1)$ using the
  normalized HSIC, denoted by $\hat{\Delta}_{V_1\rightarrow V_2}$. By the same token, we can compute the dependence between $p(V_2)$ and $p(V_1\mid V_2)$ using the
  normalized HSIC, denoted by $\hat{\Delta}_{V_2\rightarrow V_1}$. If $\hat{\Delta}_{V_1\rightarrow V_2}<\hat{\Delta}_{V_2\rightarrow V_1}$, we orient $V_1\rightarrow V_2$; otherwise $V_2\rightarrow V_1$. After orienting all possible edges, we can get the Markov equivalent class of the DAG $G$. Algorithm 2 summarizes the proposed method on how to determine causal directions.

\begin{algorithm}[h!]
	\begin{algorithmic}[1]
		\State  Object: To orient the directions in the causal skeleton $U_{G}$  
		\State  Input: The undirected skeleton output from Algorithm 1
		\For{any node $V_{i}$ adjacent to $\vU$ }
		\State Orient $\vU \rightarrow{V_{i}}$
		\EndFor

		\For{all other undirected edges}
		\State	 Apply the Meek rule

		\EndFor

		\State Find all nodes that are adjacent to $\vU$ and  have undirected edges with other nodes, denoted by ${\cal S}$
		\If {$\cal S$ is empty }
		\State \Return $G$
		\Else \Repeat
		
		\For{each node {$V\in {\cal S}$}}
		\State Consider the set $\cal Z$ of nodes that either are directed parents of $V$ or have  undirected edges to $V$
		\State Calculate the normalized HSIC using the node $V$ as the {\it effect} and the set of nodes $\cal Z$ as the {\it cause}
		\EndFor
		\State Pick the node $V$ with the smallest normalized HSIC
		\State Orient all edge directions from the nodes in $\cal Z$ to node $V$
		\State Remove the node $V$ from ${\cal S}$
		\Until $\cal S$ is empty
		\EndIf
		\State \Return $G$
		\caption{Obtain the Markov equivalence class of the DAG $G$}
			\end{algorithmic}
\end{algorithm}

\remark{In the numerical examples, we setup a threshold inclusion probability ($20\%$) below which the causality relation is not included in the hierarchical neural network models. This treatment allows us to ensure that the causalities with sufficient likelihoods are included but the less prominent relationship is omitted to improve the efficiency and simplicity of the resultant model. This threshold can be viewed as a hyperparameter. A highly threshold may yield a DAG with less vertices and therefore reduce the total number of required supervised training at the expense of being less precise on the causality relations among the data.}

\section{Supervised learning for path-dependent material laws} 
\label{sec:neuralnetworktraining}

\begin{figure}[h!]
	\centering
	\includegraphics[width=0.8\textwidth]{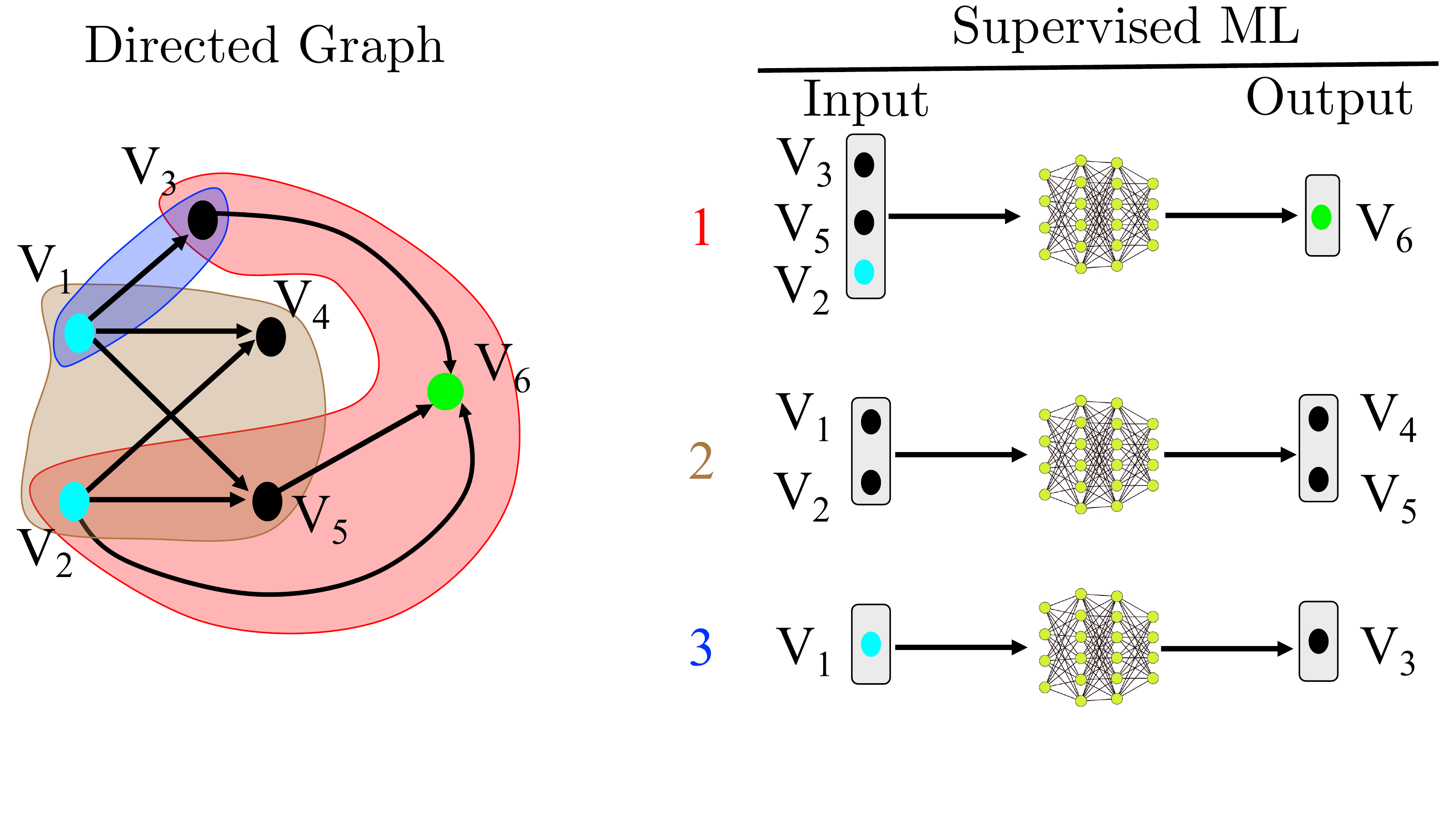}
	\caption{(a) shows a direct graph partitioned into three sub-graphs. (b) indicates how each sub-graph is used in a separate supervised machine learning task to predict the downstream node(s) from the upstream node(s).}
	\label{fig::ml-graph}
\end{figure}

Once causal relations are identified, a directed graph $G = (\mV, \mE)$ can be established where there is an edge $e_{ij} \in\mE$ from the node $V_i \in \mV$ to $V_j \in \mV$ if $V_i$ is a direct cause of $V_j$.  
Denote the leaf node to be the vertex that is not the cause to any other vertices, the root node to be the vertex that is not the target of any other vertices. 
Figure \ref{fig::ml-graph} demonstrates a directed graph indicating an information flow how leaf node(s) is related to root node(s) via some intermediate nodes, e.g., in Figure \ref{fig::ml-graph} $\{ V_1, V_2 \}$, $\{ V_6 \}$, and $\{ V_3, V_4, V_5 \}$ are sets of root, leaf, and intermediate nodes.

Along with the similar idea introduced in \citet{wang2019meta}, we aim to discover all sub-graphs that sequentially pass the information from the root to leaf, see Algorithm 3. 
Each of these subgraphs will contain leaf and root but  without any intermediate nodes. 
As such, supervised learning can help us train neural network that predict the root of each
subgraph with the corresponding leaf (or leaves) as input(s). 
To identify all sub-graphs, we traverse the graph backward from leaf to root nodes. The leaf with its immediate predecessors formed a directed tree which will be added into the list of potential sub-graphs. Then, we remove edges of the founded tree in the graph $G$. In the next step, we select a new leaf node from the updated $G$ and do the process just described until there is not any edges in $G$. For the case shown in Figure \ref{fig::ml-graph} we have the following potential sub-graphs:
\begin{align}
    &
   G_{a} = (\{ V_2, V_3, V_5, V_6 \}, \{e_{26}, e_{36}, e_{56}  \}),
    \\&
    G_{b} = (\{ V_1, V_4, V_5 \}, \{e_{14}, e_{15}  \}),
    \\&
    G_{c} = (\{ V_2, V_4, V_5 \}, \{e_{24}, e_{25}  \}),
    \\&
    G_{d} = (\{ V_1, V_3 \}, \{e_{13}  \}).
\end{align}
Those sub-graphs (directed trees) that share common upstream nodes will be merged into a bigger sub-graph. For the graph in Figure \ref{fig::ml-graph} final sub-graphs are:
\begin{align}
    &
    G_{1} = G_{a} = (\{ V_2, V_3, V_5, V_6 \}, \{e_{26}, e_{36}, e_{56}  \}),
    \\&
    G_{2} = G_{b} \cup G_{c} = (\{ V_1, V_2, V_4, V_5 \}, \{e_{14}, e_{15}, e_{24}, e_{25}  \}),
    \\&
    G_{3} = G_{d} = (\{ V_1, V_3 \}, \{e_{13}  \}).
\end{align}
For each sub-graph, we have a separate supervised machine learning (ML) task. In each ML task,  inputs and outputs features are upstream and downstream nodes in each directed sub-graph, as shown in Figure \ref{fig::ml-graph}(b).

\begin{algorithm}[ht!]
	\begin{algorithmic}[1]
		\State \textbf{Input:} Directed graph $G=(\mV, \mE)$ \Comment{result of causal graph}
		\State $\mathcal{S} \gets \emptyset$ \Comment{a set of 2-tuples (input nodes, output nodes) for ML tasks}
		\While{$\mE \ne \emptyset$}
		\State $V_l \gets$ get a leaf node of $G$ \Comment{if $G$ has multiple leaf nodes returns a random leaf}
		\State $\bar{\mV} \gets  \{  V_l \} $
		\State $\hat{\mV} \gets \{V_i | e_{il} \in \mE \}$
		\State $\mathcal{S} \gets \mathcal{S} \cup \{ (\hat{\mV}, \bar{\mV}) \}$ \Comment{$ \hat{\mV}$ is input node(s), and $\bar{\mV}$ is the output node for a potential ML task}
		\State $\mE \gets \mE \setminus \{ e_{il} | V_i \in \hat{\mV} \}$
		\EndWhile
		\State Modification: Elements $i$ and $j$ of $\mathcal{S}$ known as $(\hat{\mV}_i, \bar{\mV}_i)$ and $(\hat{\mV}_j, \bar{\mV}_j)$, respectively, are merged into one element as $(\hat{\mV}_i, \bar{\mV}_j \cup \bar{\mV}_j)$ if $\hat{\mV}_i \equiv \hat{\mV}_j $.
		\State \Return $\mathcal{S}$
		\caption{Obtain all supervised learning input-output pairs}
	\end{algorithmic}
\end{algorithm}

In the training step, we use the same architecture for all ML tasks consisting of five layers GRU-Dropout-GRU-Dropout-Dense. Each GRU layer has 32 neurons, and the linear activation function is used for the Dense layer. Gated Recurrent Unit (GRU) is one type of Recurrent Neural Networks (RNN) to model history dependence \citep{chung2014empirical}. We use the GRU  to strengthen the robustness of the ML black-box in dealing with any possible path-dependence \citep{wang2019cooperative}. 
The dropout rate in the GRU is controlled to quantify uncertainty in the model prediction, which will be detailed in Section \ref{sec:uncertainty}. All the input and output feature columns for each supervised learning task are normalized to zero mean and unity standard deviation. The loss function is defined as the mean squared root error between output prediction and ground truth. This loss is minimized by the Adam optimizer inside Keras library with Tensorflow 2.0 as its backend. Optimization process is done by the mini-batch stochastic gradient descent algorithm with a batch size 256 during 1000 epochs. The described neural network architecture and hyperparameters are chosen to be as close as possible to the ones used in \cite{wang2019meta}. 

Note that, the tuning of the hyperparameters (e.g. number of neurons, number of layers, type of activation) may have a significant effect on the performance of the neural network models. The best combination of hyperparameters can be estimated via 
a variety of approaches such as the greedy search, random search \citep{bergstra2012random}, random forest \citep{probst2019hyperparameters}, Bayesian optimization \citep{klein2017fast}, meta-gradient iteration or deep reinforcement learning \citep{wang2020non, Fuchs:2021aa}. In this work, we adopt the random search approach in \citet{bergstra2012random} to fine-tune the hyperparameters (cf. Sec. \ref{sec:TSlaw})
A rigorous hyperparameter study that compares different hyperparameter tuning for neural networks that generates constitutive laws may provide further insights on the optimal setup of the hyperparameters but is out of the scope of this study.

For the blind prediction, after training, we start from the root (e.g., $\vU$ in the traction-separation law) and sequentially predict intermediate nodes via their corresponding sub-graph trained neural networks (NN) until reaching the leaf node. For example in the case shown in Figure \ref{fig::ml-graph}: NN 3 predicts  $V_3$ from input $V_1$; $V_4$ and $V_5$ are predicted by NN 2 from inputs $V_1$ and $V_2$; and finally NN 1 is used to predict the target variable $V_6$ from input $V_2$ and the obtained intermediate nodes $V_3$ and $V_5$.

\section{Uncertainty propagation in causal graph with dropout layers}  \label{sec:uncertainty}
As described in Section \ref{sec:neuralnetworktraining}, we use the deep learning method, GRU in the training step and prediction to handle path-dependent predictions. However, the GRU itself is not designed to capture prediction uncertainty, which is of crucial importance in learning material law. In machine learning and statistics, Bayesian methods as probabilistic models provide us a natural way to quantify the model uncertainty through computing 
 the posterior distribution of unknown parameters \citep{xie2020bayesian}. However, these  methods often suffer from a prohibitively high computational cost.  In this paper, we show that the dropout technique \citep{srivastava2014dropout} used in the GRU can quantify uncertainty in prediction as a Bayesian approximation. 

  Dropout, a regularization method that randomly masks or ignores neurons during training, 
  has been widely used in many deep learning models to avoid over-fitting and improve prediction \citep{hinton2012improving, li2016improved, boluki2020learnable}.  \cite{gal2016dropout} firstly prove the link between dropout and a well known probabilistic model, the Gaussian process \citep{rasmussen2003gaussian},  and show  that the use of dropout in the feed forward neural networks  can be interpreted as a Bayesian approximation of Gaussian processes. In the context of  RNNs,  \cite{gal2016theoretically} treated RNNs as probabilistic models by assuming network weights as random variables with a Gaussian mixture prior (with one component fixed at zero with a small variance). Such a technique is similar to the spike-and-slab prior in Bayesian statistics for variable selection \citep{ishwaran2005spike}. Then \cite{gal2016theoretically} show that optimizing the objective in the variational inference \citep{blei2017variational} for  approximating the posterior distribution over the weights  is equivalent to conducting dropout in the respective RNNs, and demonstrate the implementation in one commonly-used RNN model, the long short-term memory \citep{hochreiter1997long}. In this section, we propose to extend the technique developed in \cite{gal2016theoretically} in the context of GRUs for  uncertainty quantification (UQ) in the prediction. 
  
Given training inputs $\bX$ and the corresponding output $\bY$, suppose that we aim to predict an output $\by^*$ for a new input $\bx^*$.  
From the Bayesian point of view, the prediction uncertainty can be characterized by the posterior predictive distribution of  $\by^*$  as follows: 
\begin{equation}
p\left(\by^{*} \mid \bx^{*}, \bX, \bY\right)=\int p\left(\by^{*} \mid \bx^{*}, \bomega\right) p(\bomega \mid  \bX, \bY) \mathrm{d} \bomega,
\end{equation} 
where $\bomega$ includes all unknown model parameters,  $p(\bomega \mid  \bX, \bY)$ is the posterior distribution of $\bomega$. In the GRU, all   unknown weights can be viewed as $\bomega$. As the posterior distribution  $p(\bomega \mid  \bX, \bY)$  is generally intractable, the variational inference method approximates it by proposing a variational  distribution $q(\bomega)$ and then finding the optimal parameters in the variational  distribution through minimizing the Kullback-Leibler (KL) divergence between the approximating distribution and the full posterior distribution: 
\begin{equation}\label{eq:kL}
\mathrm{KL}(q(\boldsymbol{\omega}) \| p(\boldsymbol{\omega} \mid \bX, \bY)) \propto  - \int q(\boldsymbol{\omega}) \log p\left(\bY \mid \bX, \bomega \right) \mathrm{d} \bomega+\mathrm{KL}(q(\boldsymbol{\omega}) \| p(\boldsymbol{\omega})), 
\end{equation} 
where $p(\boldsymbol{\omega})$ is the prior distribution of $\bomega$.

 Given an input sequence $\bX=[\bx_1, \dots, \bx_T]$ of length $T$, the hidden state $\bh_t$ at time step $t$ in the GRU neural network can be generated as follows:
\begin{equation}\label{eq:GRU}
\begin{aligned} 
\bz_{t} &=\sigma\left(\bW_{\bz} \bx_{t}+\bU_{\bz} \bh_{t-1}+\bb_{\bz}\right), \\ 
\br_{t} &=\sigma\left(\bW_{\br} \bx_{t}+\bU_{\br} \bh_{t-1}+\bb_{r}\right), \\
\bi_{t} & = \tanh\left(\bW_{\bh} \bx_{t}+\bU_{\bh}\left(\br_{t} \odot \bh_{t-1}\right)+\bb_{\bh}\right), \\ \bh_{t} &=\bz_{t} \odot \bh_{t-1}+\left(1-\bz_{t}\right) \odot \bi_{t},
\end{aligned}  
\end{equation}   
where $\sigma$ denotes the sigmoid function and $\odot$ denotes the element-wise product. Also, we assume that the model output at time step $t$ can be written as $f_{\bY}(\bh_t) = \bh_t\bW_{\bY}+\bb_{\bY}$. Then the unknown parameters in the GRU are $\boldsymbol{\omega}=\left\{\bW_{\bY}, \bW_{\bz}, \bU_{\bz}, \bW_{\br}, \bU_{\br}, \bW_{\bh}, \bU_{\bh}, \bb_{\bY}, \bb_{\bz},\bb_{\br},\bb_{\bh} \right\}$. We write 
$\bh_{t}=f^\bomega_{\bh}\left(\bx_{t}, \bh_{t-1}\right)$ and $f^{\bomega}_{\bY}$ for the output in order to make the dependence on $\bomega$ clear. 

Then the right hand of \eqref{eq:kL} can be written as follows: 

\begin{eqnarray}\label{eq:KL2}
  - \int q(\boldsymbol{\omega}) \log p\left(\bY \mid  f_{\bY}^{\bomega}\left(\bx_1, \dots, \bx_T, f_{\bh}^{\bomega}\left(\bx_T, f_{\bh}^{\bomega}(\dots f_{\bh}^{\bomega}(\bx_1, \bh_0)\dots)         \right)\right)\right)
\mathrm{d} \bomega+\mathrm{KL}(q(\boldsymbol{\omega}) \| p(\boldsymbol{\omega})), 
\end{eqnarray} 
which can be approximated by Monte Carlo integration with the generated samples  $\hat{\bomega}^b\sim q(\bomega)$ and plug in the sampled $\hat{\bomega}^b$'s to \eqref{eq:KL2}.

Following \cite{gal2016theoretically}, we use a mixture of Gaussian distributions as the  variational distribution for every weight matrix row $\bomega_k$: 
\begin{equation}
q(\bomega)=\prod_{k=1}^Kq(\bomega_k), \  \  q\left(\bomega_{k}\right) = \pi  \mathcal{N}\left(\bomega_{k} ; \mathbf{0}, \tau^{2} I\right)+(1-\pi) \mathcal{N}\left({\bomega}_{k} ; {\bm m}_{k}, \tau^{2} I\right),
\end{equation} 
where $\pi$ is the dropout probability, ${\bm m}_{k}$ is the  variational parameter (row vector), and $\tau^2$ is a small variance. We optimize over ${\bm m}_{k}$ by minimizing the KL divergence in \eqref{eq:KL2}. Sampling each row of $\hat{\bomega}^b$  is equivalent to randomly mask rows in each weight matrix, i.e., conducting dropout. Then the predictive posterior distribution can be approximated by 
\begin{equation}\label{eq:pred}
p\left(\by^{*} \mid \bx^{*}, \bX, \bY\right)=\int p\left(\by^{*} \mid \bx^{*}, \bomega\right) p(\bomega \mid  \bX, \bY) \mathrm{d} \bomega \approx \frac{1}{B}\sum_{b=1}^B  p(\by^{*} \mid \bx^{*}, \hat{\bomega}^b),
\end{equation} 
where $\hat{\bomega}^b\sim q(\bomega)$ and $B$ is the total number of generated samples.

To implement the dropout in the GRU, we  re-parametrize \eqref{eq:GRU} as follows: 
\begin{equation}
\left(\begin{array}{c}\bz_t \\ \br_t  \\ \bi_t \end{array}\right)=\left(\begin{array}{c}\sigma \\ \sigma \\ \tanh \end{array}\right)\left(\left(\begin{array}{c}\bx_{t} \circ {\bm m}_{\bx} \\ \bh_{t-1} \circ {\bm m}_{\bh}\end{array}\right) \cdot \bomega\right), 
\end{equation}
where ${\bm m}_{\bx}$ and ${\bm m}_{\bh}$  denote randomly masks repeated at all time steps. 
 
We take the prediction tasks in Fig. \ref{fig::ml-graph} for example. The posterior predictive distribution of $V_3$ given $V_1$ is straightforward by \eqref{eq:pred} since only root and leaf nodes are involved. When involving the intermediate nodes, e.g., the subgroup $G_1$,  the  posterior predictive distribution of $V_6$ given the root nodes ($V_1$ and $V_2)$ can be computed by 
\begin{eqnarray}\label{eq:MC}
p\left(V_6 \mid V_1, V_2\right)=\int p\left(V_6 \mid V_2, V_3, V_5, \bomega\right) p(V_3\mid V_1, \bomega)p(V_5\mid V_1, V_2, \bomega)p(\bomega \mid V_1, V_2) \mathrm{d} V_3\mathrm{d} V_5\mathrm{d} \bomega.
\end{eqnarray}
 Specifically, when we generate $B$ samples from the posterior distribution of $\bomega$, we also generate $B$ samples for the intermediate nodes. Those $B$ samples of $\bomega$ and the corresponding intermediate nodes can be then used in Monte Carlo integration to calculate \eqref{eq:MC}. 
Throughout this paper, we set $B$=200.

\section{Numerical Examples} \label{sec:examples}
In this section, we conduct  two numerical experiments  to test our proposed framework that combines causal
discovery with deep learning to build constitutive laws for granular materials. In Section \ref{sec:TSlaw}, the causal discovery 
is conducted to determine the constitutive relationships for 
an RVE interface composed of spherical grains. The subsequent 
supervised machine learning then leverages the causal relations learned from the causal discovery algorithm to establish a serial of supervised learning that constitutes a forecast engine for traction. The propagation of uncertainty is enabled by the dropout technique that approximates Monte Carlo simulations to determine the confidence intervals for a given dropout rate. In Section \ref{sec:bulkplasticity}, the same exercise is repeated for another data set to generate hypoplasticity surrogate 
model for a discrete element assembly where new topological 
measures are computed and incorporated into the proposed framework
to (1) discover new physical mechanisms and (2) determine the benefit 
of the new discovery on the accuracy, robustness, and consistency of the forward predictions on unseen events. 

\subsection{Numerical Example 1: Machine Learning traction-separation law}
\label{sec:TSlaw}

Traction-separation laws are known as one of the main ingredients of cohesive fracture models used for brittle materials \citep{pandolfi2000three, park2011cohesive}. Generally, a traction-separation law constitutes a relation between traction and displacement jump fields over the fracture surface. There exist many hand-crafted models developed by experts for different applications \citep{park2011cohesive}, while no unified framework had been developed until recently by \citet{wang2019meta} who suggest an approach based on reinforcement learning. Also, in some applications such as granular materials more descriptors, e.g., porosity or fabric tensor, should be considered in these constitutive laws \citep{sun2013multiscale} to derive more predictive models. Lack of robustness in adding more descriptors is another weakness of classical models. 

Our first test for the data-driven causal discovery model with dropout UQ is on the traction-separation law data publicly available in the repository Mendeley data
(cf. \citet{wang2019discrete}). This dataset has also been used in \citet{wang2019meta}
where the traction-separation law is determined from reinforcement learning. Our major point of departure is three-fold. Firstly, we develop a causal discovery algorithm to identify causal relations among history-dependent physical quantities in RVE simulations. Secondly, 
we decouple the causal discovery from the training of the neural network such that we now first discover causal relations, then utilize the discovered relationships to generate quantitative predictions using the method detailed in section \ref{sec:neuralnetworktraining}. Thirdly, we introduce the Bayesian approximation using the dropout technique to propagate the uncertainty in the causal graph, building upon the theoretical framework established in \citet{gal2016theoretically}.

The database includes 100 DEM experiments. In each DEM experiment, the time history of 
all the variables included in the DAG are recorded. Each experiment is conducted by a different ratio of normal to tangential loading rate and loading-unloading cycles on the same representative volume element of granular materials. As such, the total number of time-history data points in these experiments may vary from 51 to 111. 

The interested reader is referred to the appendix in \citet{wang2019meta} for more information. In our study, feature space consists of displacement jump vector, traction vector, coordination number, symmetric part of fabric tensor, and porosity. We use half of the experiments for causal discovery and training artificial neural networks, and the rest is used for test and validation. In the causal discovery step, as different experimental setups may lead to different causal relations among variables, we apply the  proposed causal discovery algorithm (Algorithms 1 and 2 in Section \ref{sec:discovery}) to each of the training experiment, and then report the final causal graph by calculating the inclusion probabilities of directed edges appearing in all training experiments. The inclusion probability of one edge is defined as the proportion of causal graphs containing this edge.   
The directed edges with inclusion probabilities being larger than a pre-defined threshold (20\% in our paper) are kept in the final causal graph. When both edge directions between two variables appear with positive inclusion probabilities (e.g., $V_i\rightarrow V_j$ and $V_j \rightarrow V_i$ both exist), we keep the edge direction that has a higher inclusion probability. 
The goal of the resultant model is to predict the same granular assembly responds to a different cyclic loading path unseen in the training. As such, the focus of this model is to generate a surrogate for one representative element volume.

\begin{figure}[h!]
    \centering
    \includegraphics[width=0.3\textwidth]{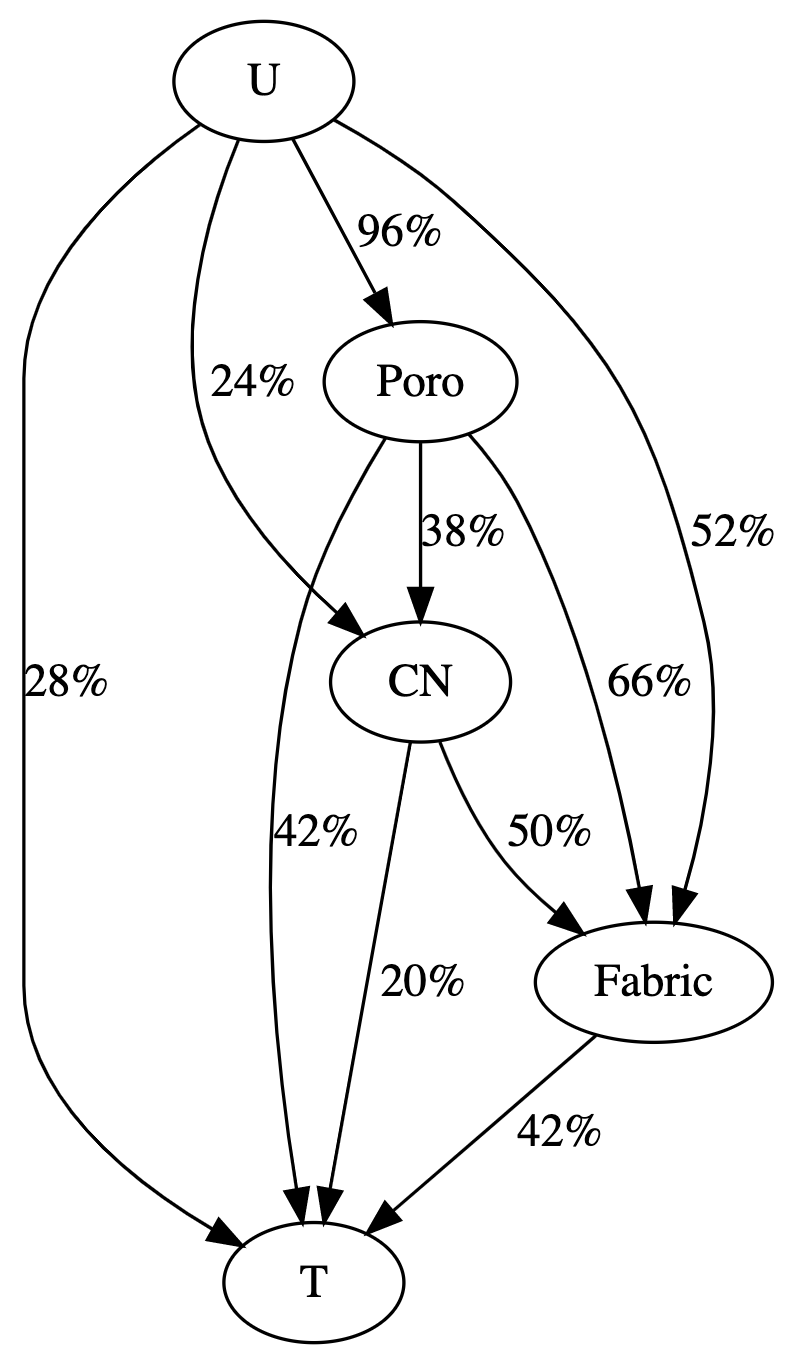}
    \caption{Final causal graph for the traction-separation law deduced from time-history of displacement, traction, porosity, coordination number, and fabric tensor. The number on each edge   represents the edge inclusion probabilities 
     among all possible causal relations from the training data sets.}
    \label{fig::trac-sep-final-graph}
\end{figure}

Fig. \ref{fig::trac-sep-final-graph} plots the final causal graph on the training data sets with edge inclusion probabilities. The strong confidence ($96\%$) in the edge starting from the displacement jump vector to porosity is consistent with the common field knowledge, i.e., the immediate consequence of displacement jump is the volume change.
 
  The displacement jump vector, as the only control variable, affects all the intermediate physical quantities and traction vector. This observation may seem to be trivial, but it is not always the case which will be shown in the next example. The causal effects of fabric and coordination number on traction is aligned with the modern Critical State Theory \citep{li2012anisotropic} which is obtained without expert interpretation by the causal algorithm.
 Note that fabric encodes microstructural information in more detail such as directional dependence due to its tensorial nature, rather than porosity which smears out information into one scalar quantity. Therefore, it is reasonable to see that fabric has a considerable contribution in describing material behavior with a complex arrangement of force chains at the microstructural level.

\remark To make sure the suggested neural network architecture in Sec. \ref{sec:neuralnetworktraining} works satisfactorily in this problem, we trained several neural networks with different hyper-parameters for each sub-graph learning task in Fig. \ref{fig::trac-sep-final-graph}. We used the random search approach \citep{bergstra2012random} implemented in Keras Tunner package \citep{o2019keras} for this study. The number of GRU layers is kept fixed and equal to two, and in the training stage, the dropout rate is set to zero. The number of epochs is also set to 200. The parameters under this study are as follows: the number of units in GRU layers are sampled from the set $\{ 8, 16, 32, 64 \}$, the Adam learning rate is sampled from the set $\{ 0.01, 0.001, 0.0001 \}$, the batch size for the SGD algorithm is sampled from the set $\{32, 64, 128, 256, 512\}$. Based on these hyperparameter ranges, each subgraph learning task has 240 different configurations; however, in the random search algorithm, we set the number of trials to 100 for each subgraph training task to reduce overall computational time. For this hyperparameter tuning task, we choose 50 data sets as the training set and another 50 data sets as the validation set. Our metric for selecting the best configuration is the minimum validation loss. We found that the learning rate 0.001 and batch size 32 are common among all the best configurations of subgraphs. In Table \ref{tab:hyptun} we study the effect of number of units in GRU layers when learning rate and batch size have their optimal values.

\begin{table}[h!]
\centering
 \begin{tabular}{l | l | l | l | l } 
subgraph & mean & standard deviation & number of configurations & suggested NN \\
\hline
Porosity  & 1.6e-6 & 7.36e-7 & 10 & 3.7e-6\\
CN  & 1.04e-3 & 2.3e-5 & 5 & 1.07e-3\\
Fabric  & 2.2e-3 & 3.98e-4 & 7 & 2.4e-3\\
Traction & 5.1e-5 & 1.01e-5 & 8 & 9.1e-5
\end{tabular}
 \caption{This table reports the mean and standard deviation of the validation loss among different configurations which are different based on their number of units utilized for each GRU layer when the optimal learning rate 0.001 and batch size 32 is chosen. Notice that we randomly conduct 100 trials for each subgraph with different hyperparameters. The last column shows the validation loss when the neural network has the same architecture suggested in Sec. \ref{sec:neuralnetworktraining}. Based on the standard deviation values, we observe that the number of units in GRU layers has a marginal effect on the performance. The suggested fixed neural network architecture in Sec. \ref{sec:neuralnetworktraining} for all sub-graphs has almost the same performance as the best optimal configurations.}
   \label{tab:hyptun}
\end{table}

\begin{figure}[h!]
    \centering
    \includegraphics[width=0.35\textwidth]{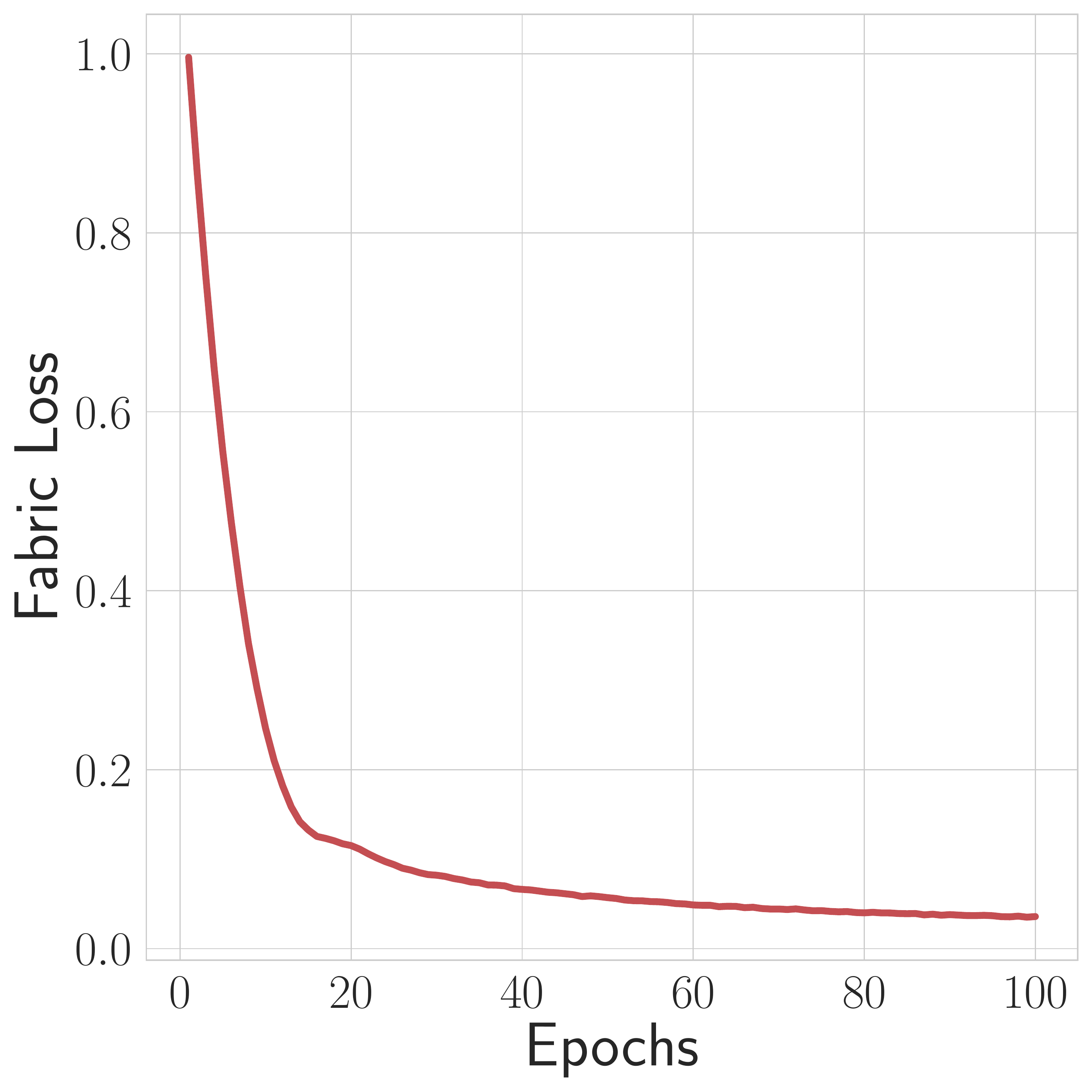}
    \includegraphics[width=0.35\textwidth]{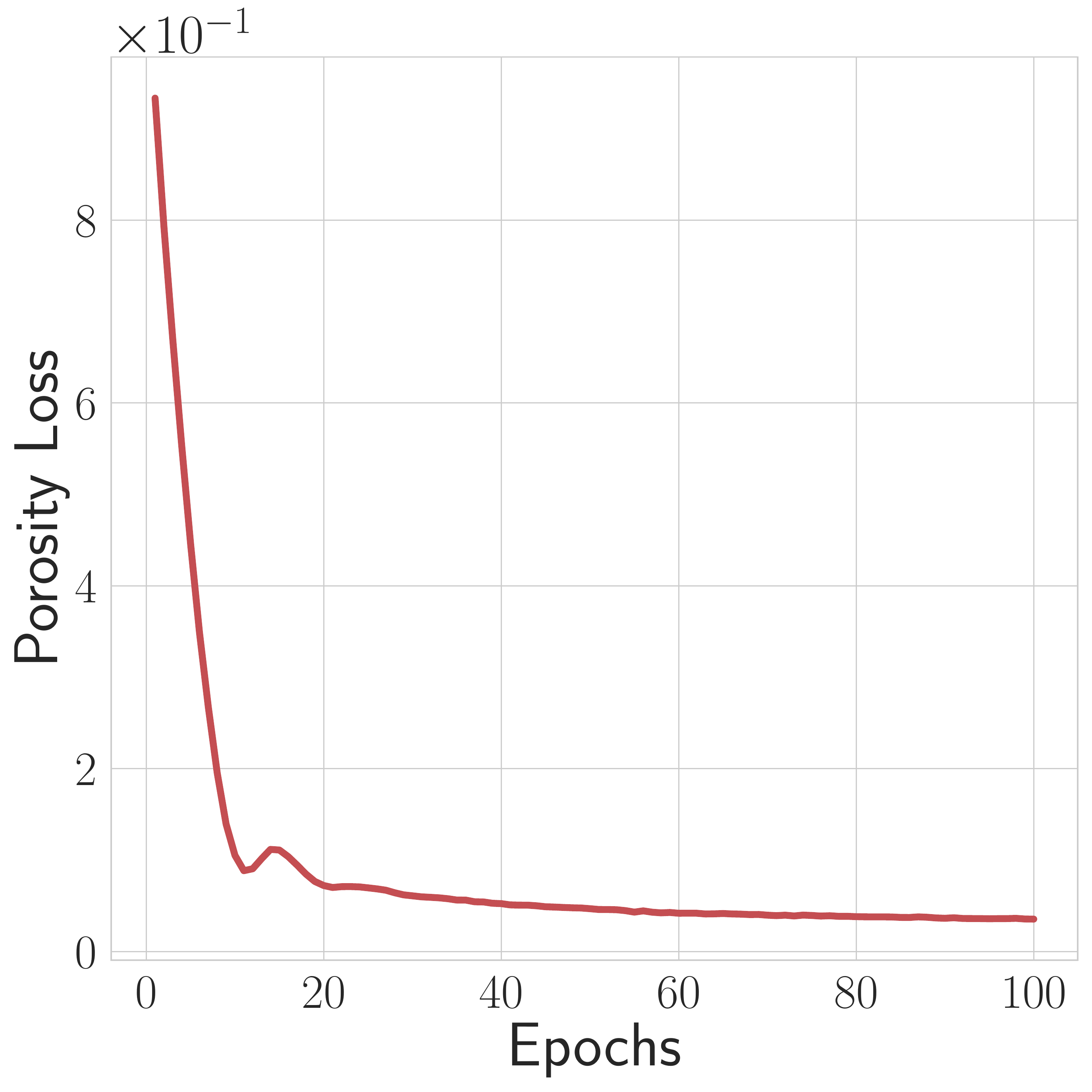}
    \includegraphics[width=0.35\textwidth]{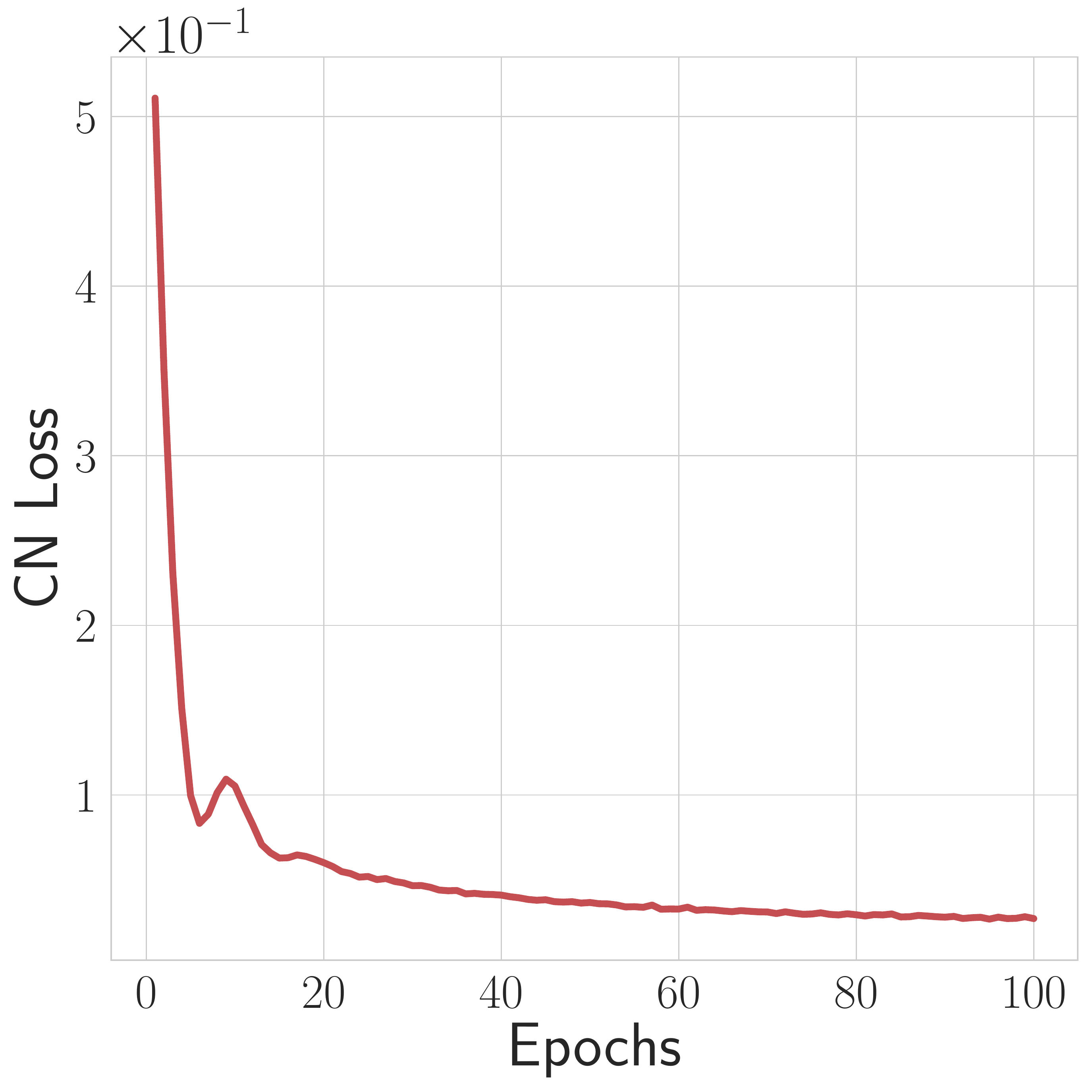}
    \includegraphics[width=0.35\textwidth]{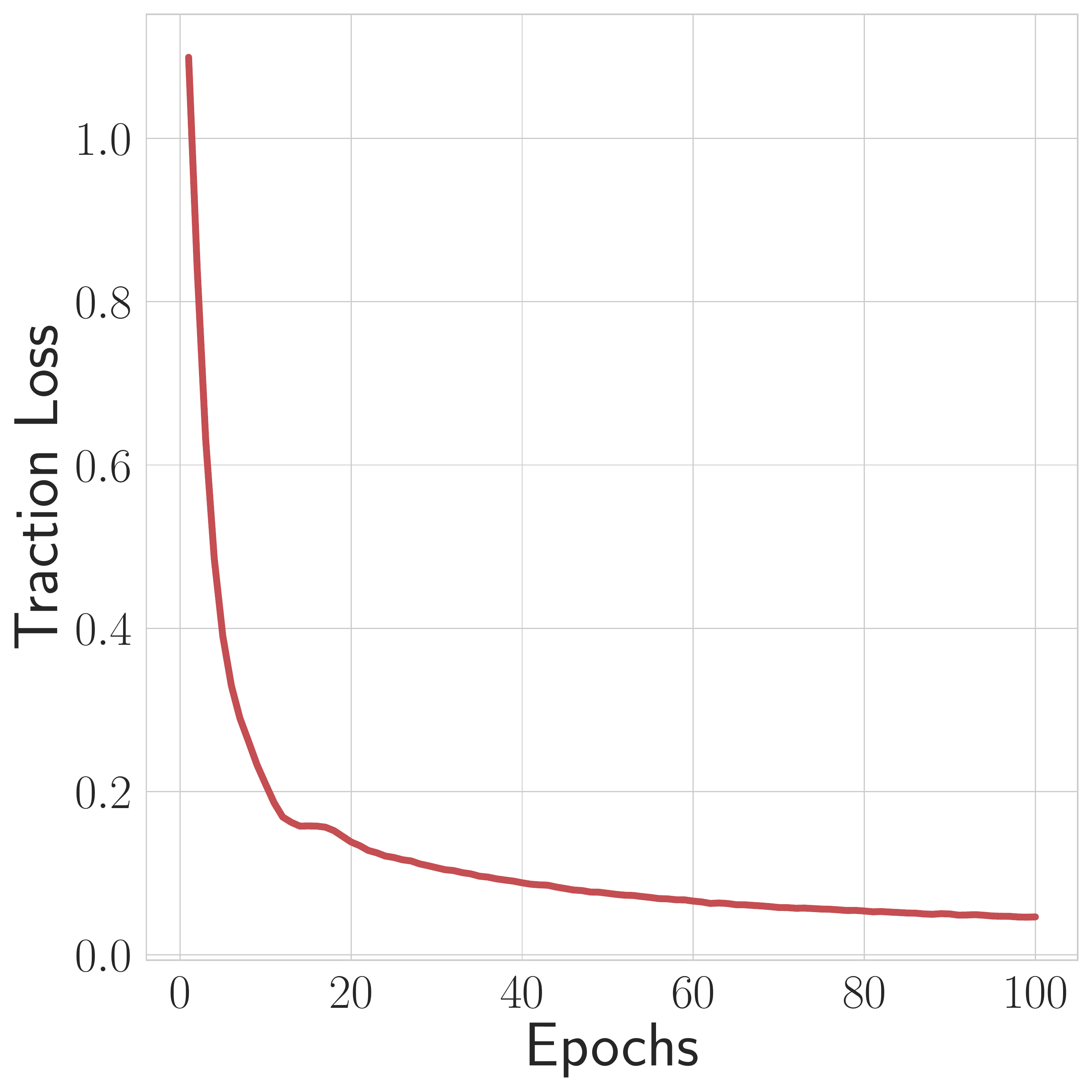}
    \caption{Training loss convergence behavior for four supervised learning tasks deduced from the causal graph. Top left: fabric is predicted based on displacement, porosity and coordination number; Top right: porosity is predicted based on displacement; Bottom left: coordination number is predicted based on displacement and porosity; Bottom right: traction is predicted based on displacement, coordination number, fabric, and porosity.}
    \label{fig::trac-sep-loss}
\end{figure}

Applying Algorithm 3 to Fig. \ref{fig::trac-sep-final-graph}, we need to perform four supervised learning tasks: 1) predict Poro from the input $\vU$; 2) predict CN from the input $\vU$ and the intermediate node Poro; 3) predict fabric from the  input $\vU$ and the intermediate nodes Poro and CN; and 4) predict the target variable $\vec{T}$ from the  input $\vU$ and the obtained  intermediate nodes Poro, CN, and fabric. We then use the GRU to train  each sub-graph with the 
  dropout rate  being 0.2 for both training and feed-forward predictions.  Fig. \ref{fig::trac-sep-loss}   confirms that the neural network architecture proposed in section \ref{sec:neuralnetworktraining} yields satisfactory  performance  for all supervised tasks. To examine the generalization performance of the trained neural networks, we study the empirical cumulative distribution
functions (eCDFs) for training and test data sets following \citet{wang2019meta}.

\begin{figure}[h!]
    \centering
    \includegraphics[width=0.4\textwidth]{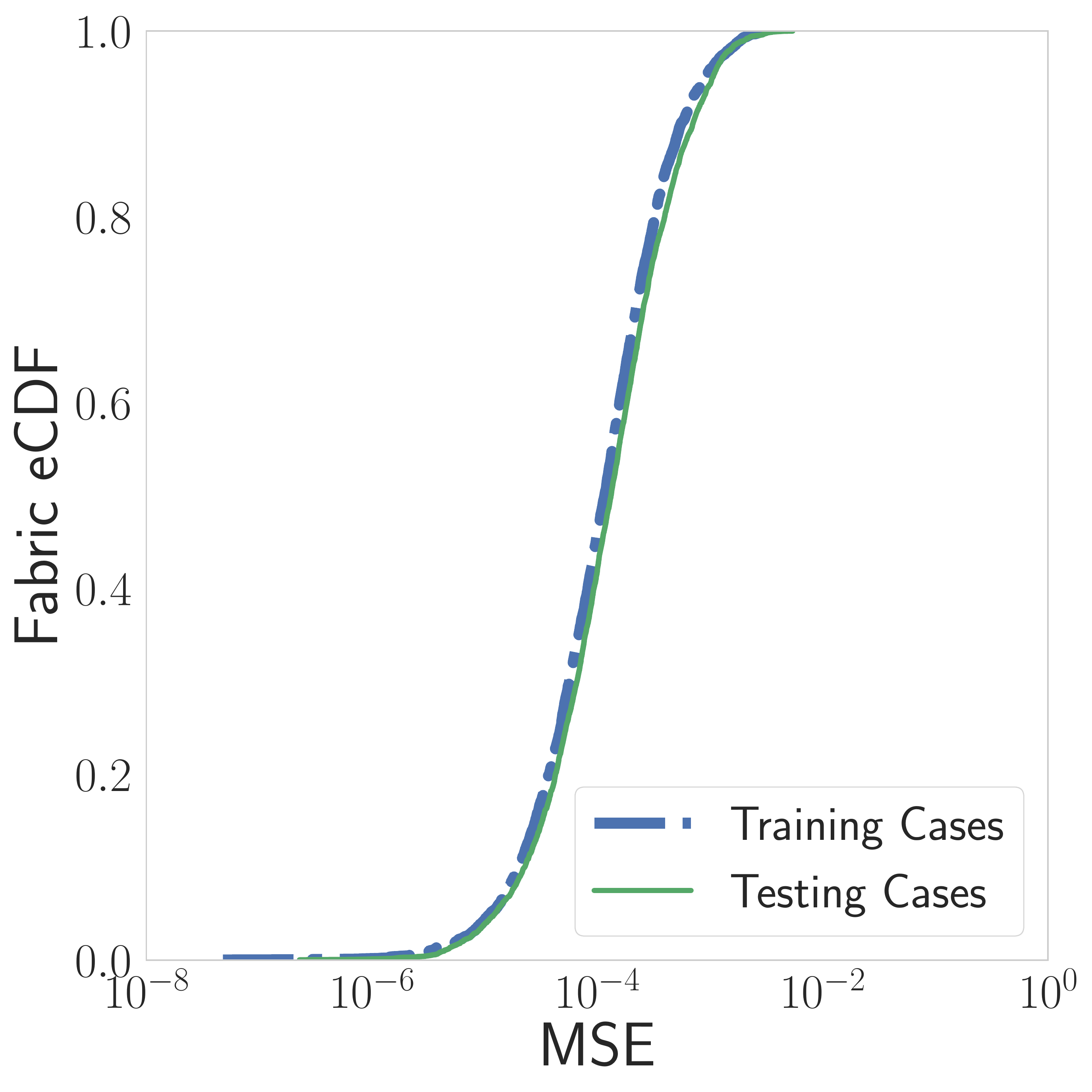}
    \includegraphics[width=0.4\textwidth]{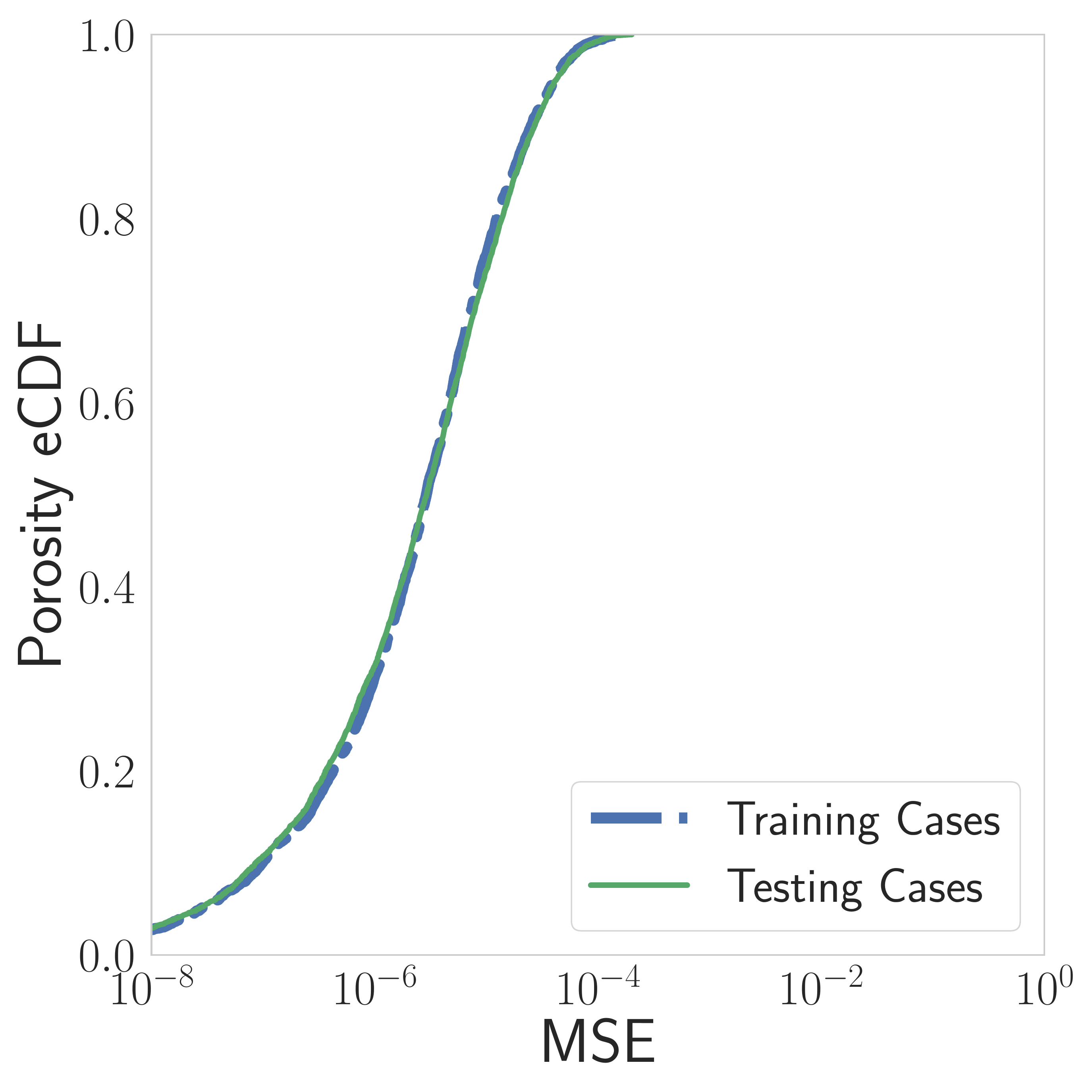}
    \includegraphics[width=0.4\textwidth]{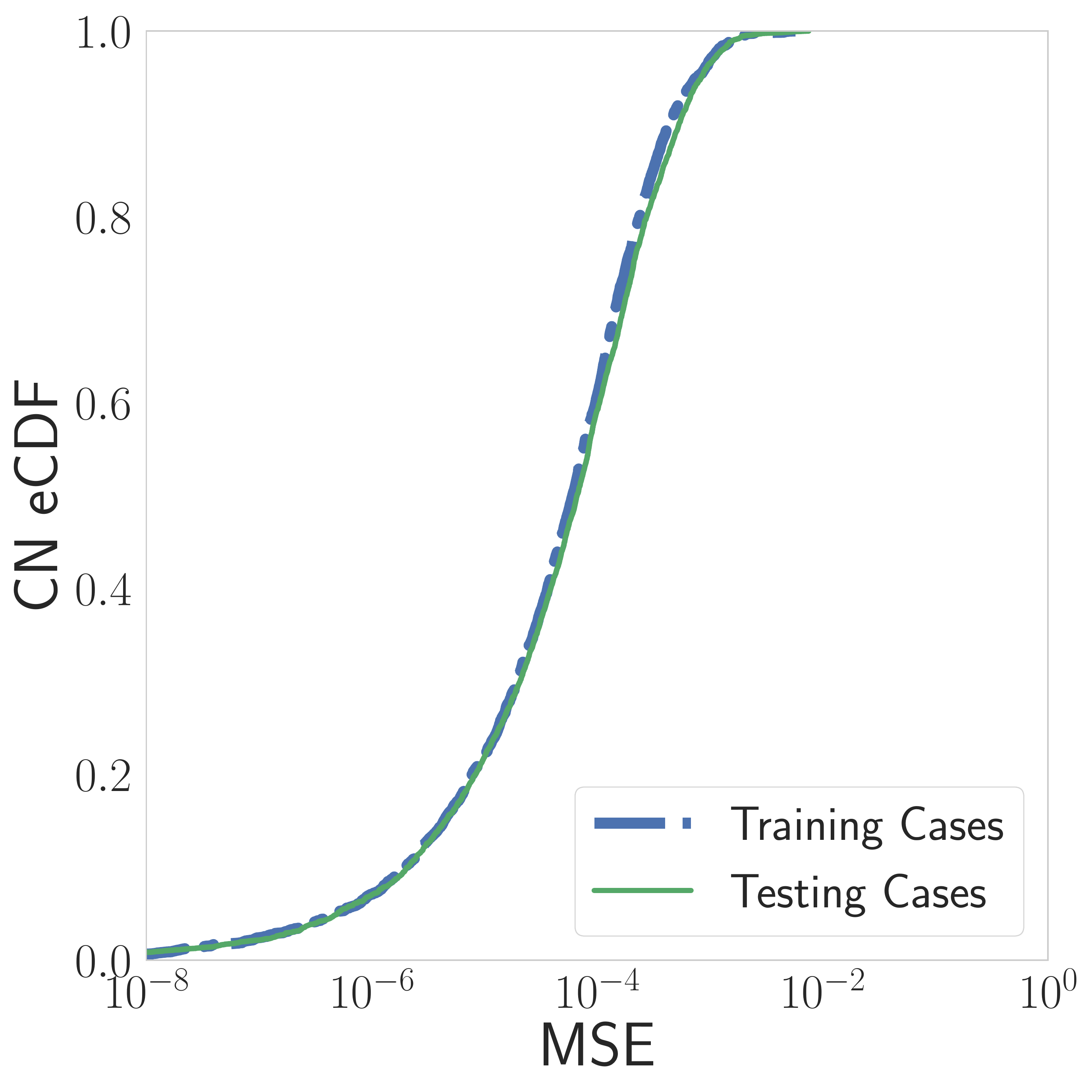}
    \includegraphics[width=0.4\textwidth]{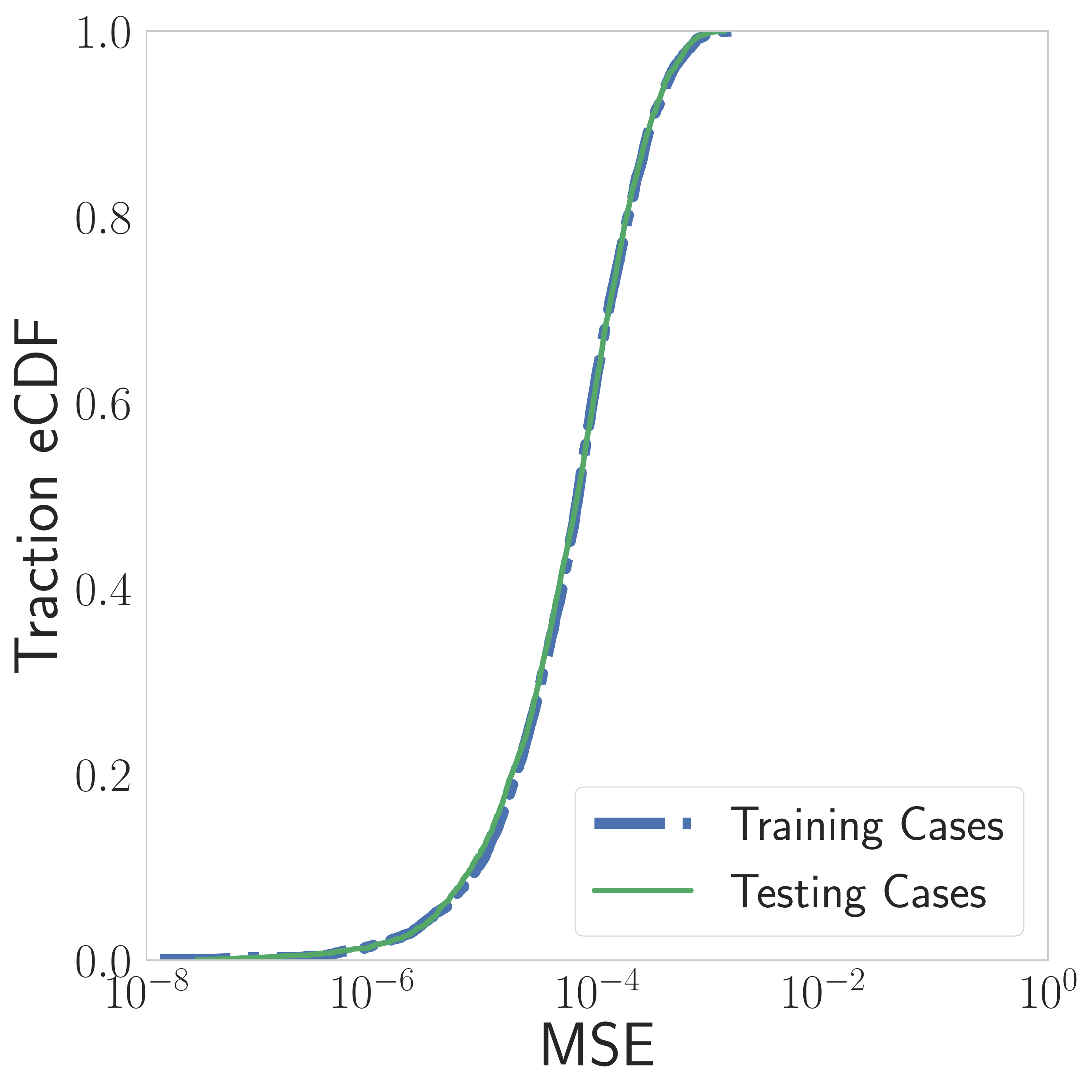}
    \caption{Empirical Cumulative Distribution Function (eCDF) for prediction on training data sets and mean value of predictions on test data sets. We use 50 data sets for training and 50 other data sets for testing purposes. For each test case, we perform 200 feed-forward predictions with the drop-out rate 0.2.}
    \label{fig::trac-sep-ecdf}
\end{figure}

We define the point-wise scaled mean squared error (MSE) between a set of ground-truth values with size $N$ and its corresponding approximation set as:
\begin{equation}
e_i = \frac{1}{N} \sum_{i=1}^{N} (\mathcal{S}( y^{\mathrm{true}}_i) - \mathcal{S}( y^{\mathrm{appx}}_i) ),
\end{equation}
where $\mathcal{S}$ is a scaling function. In this paper, the scaling function linearly transforms a set of values into a new set where all values are in the range $[0,1]$. We perform 
$200$ feed-forward predictions to obtain the distribution of each feature output at a specified load-step. For eCDF calculation only, we use the average of these $200$ predictions to approximate the feature output. In this way, the discrete eCDF of a target output feature, such as porosity, at data point $i$ is defined as $F_N(e_i) = \frac{1}{M} \sum_{j=1}^{M} \mathbb{1}(e_i \ge e_j)$ where $e_i$ is the point-wise scaled MSE between the feature ground-truth value and its predictions' average, $M$ is the total number of instances (i.e., the total number of data points across 50 training data sets) used for eCDF calculations, and $\mathbb{1}(\cdot)$ is the indicator function. 
Fig. \ref{fig::trac-sep-ecdf} plots eCDFs for all feature outputs in training and testing modes. In these plots, the eCDFs for test and training cases are almost the same, indicating  no  under-fitting or over-fitting issue exists. Note that the use of dropout in the GRU is not only for uncertainty quantification in prediction, but also to 
   improve model generalization performance. 

\begin{figure}[h!]
    \centering
    \includegraphics[width=0.3\textwidth]{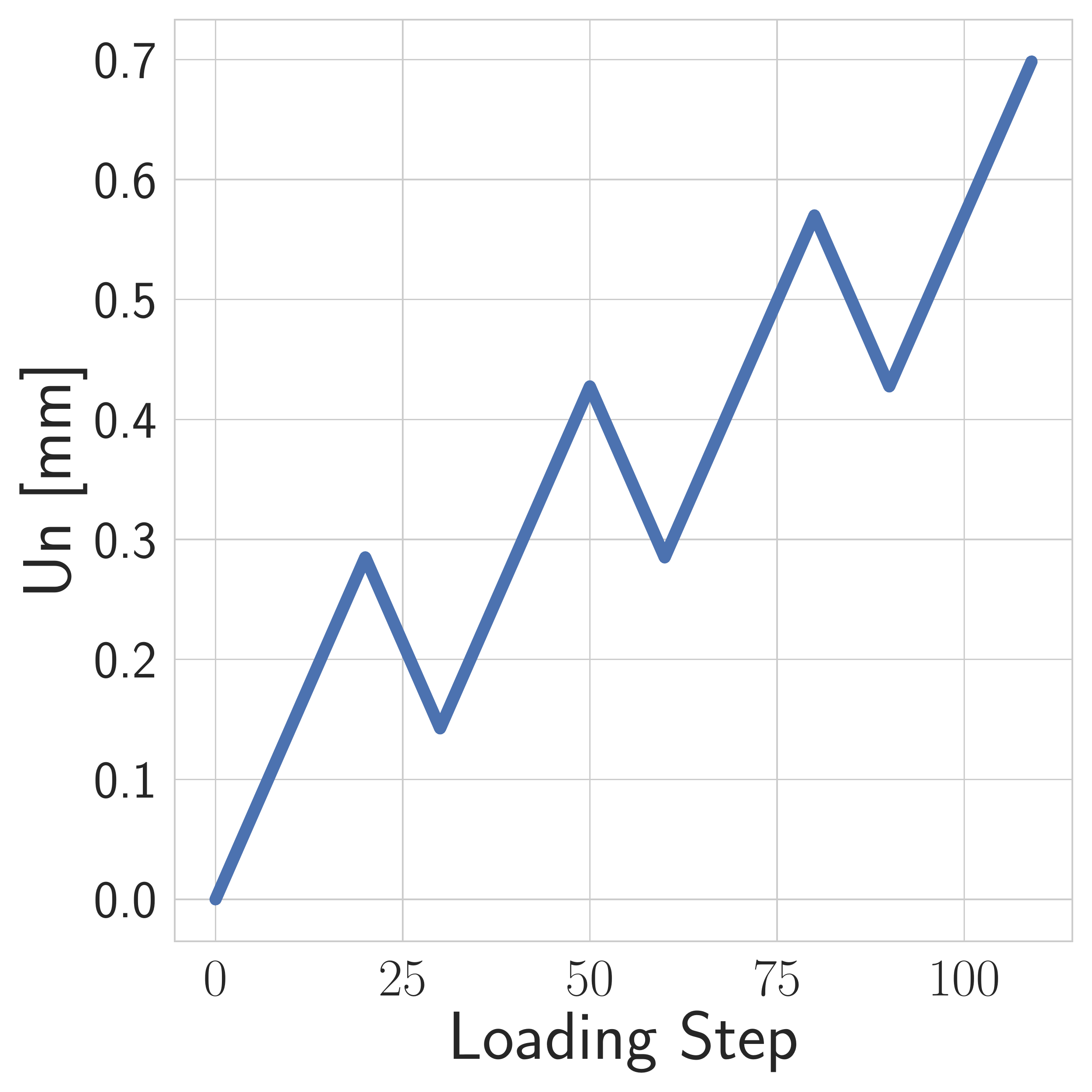}
    \includegraphics[width=0.3\textwidth]{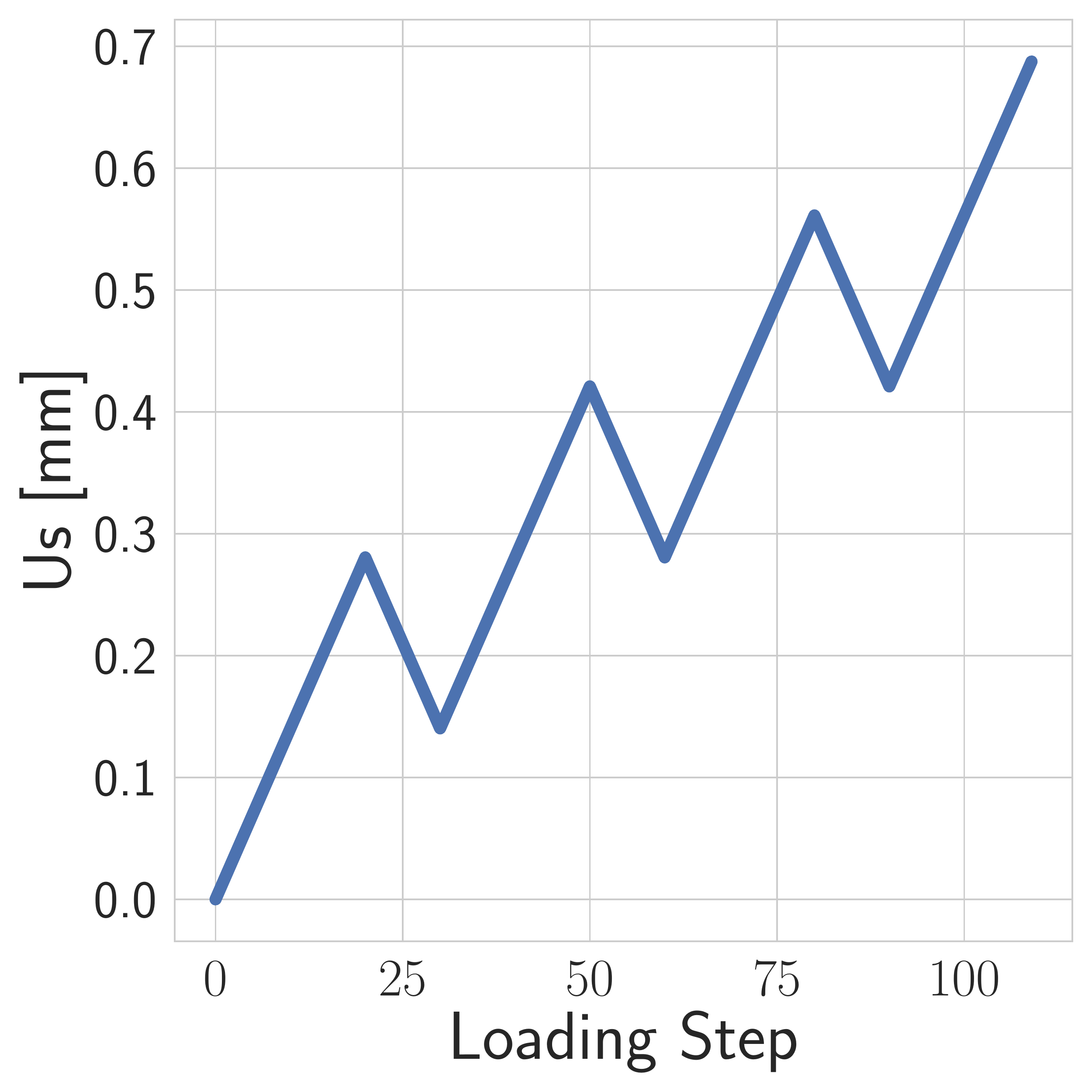}
    \caption{Applied normal and shear displacements in one of the experimental cases. Increment in normal jump indicates more compression.}
    \label{fig::trac-sep-un-us}
\end{figure}

In the following, we present prediction results for one of the test cases where its applied normal and shear displacements are plotted in Fig. \ref{fig::trac-sep-un-us}. Normal and shear displacement jumps experience cyclic loading-unloading path and are kept equal in magnitude.

\begin{figure}[h!]
    \centering
    \includegraphics[width=0.5\textwidth]{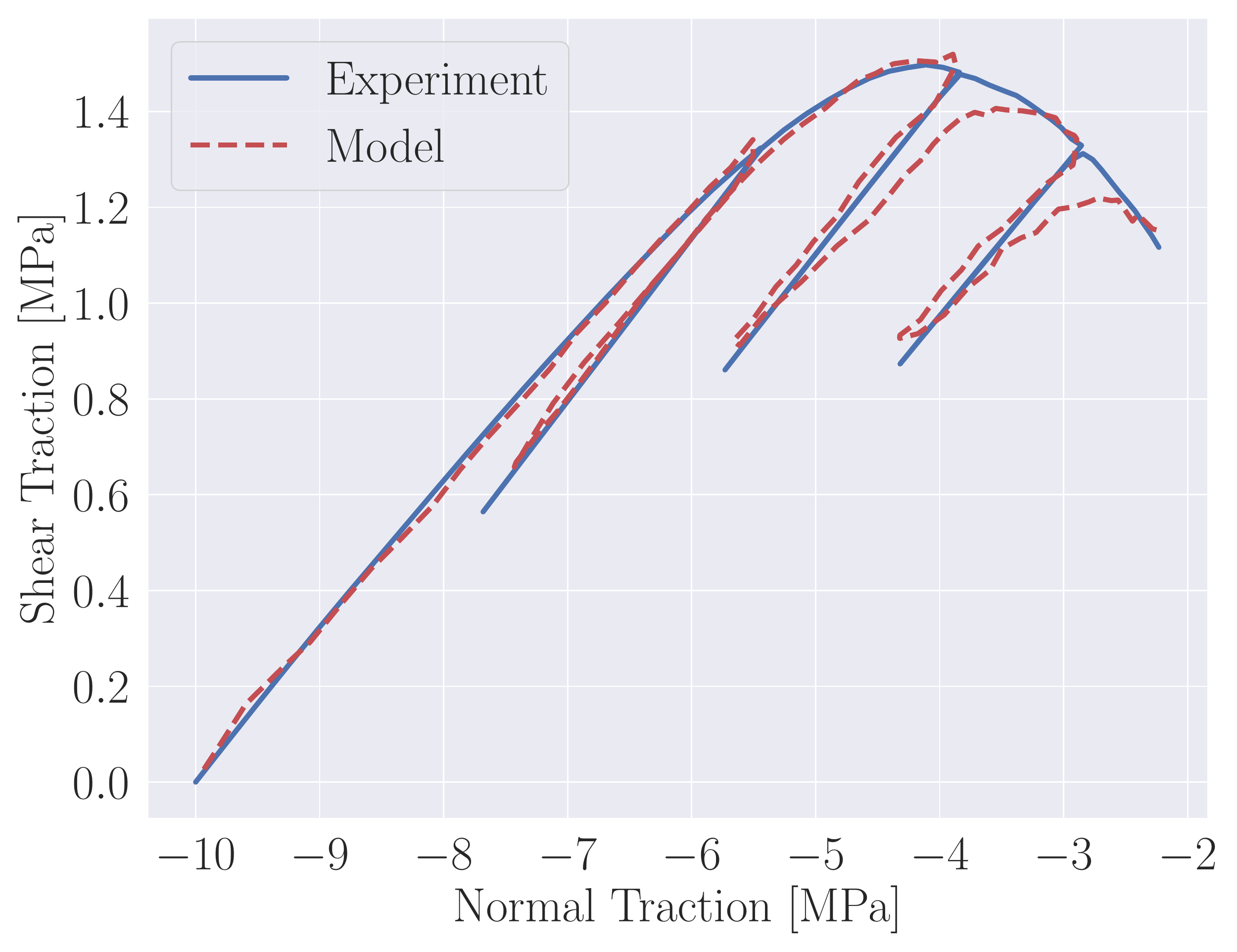}
    \caption{Comparison of normal-shear traction between model and experiment in one case.}
    \label{fig::trac-sep-tn-ts}
\end{figure}

\begin{figure}[h!]
    \centering
    \includegraphics[width=0.4\textwidth]{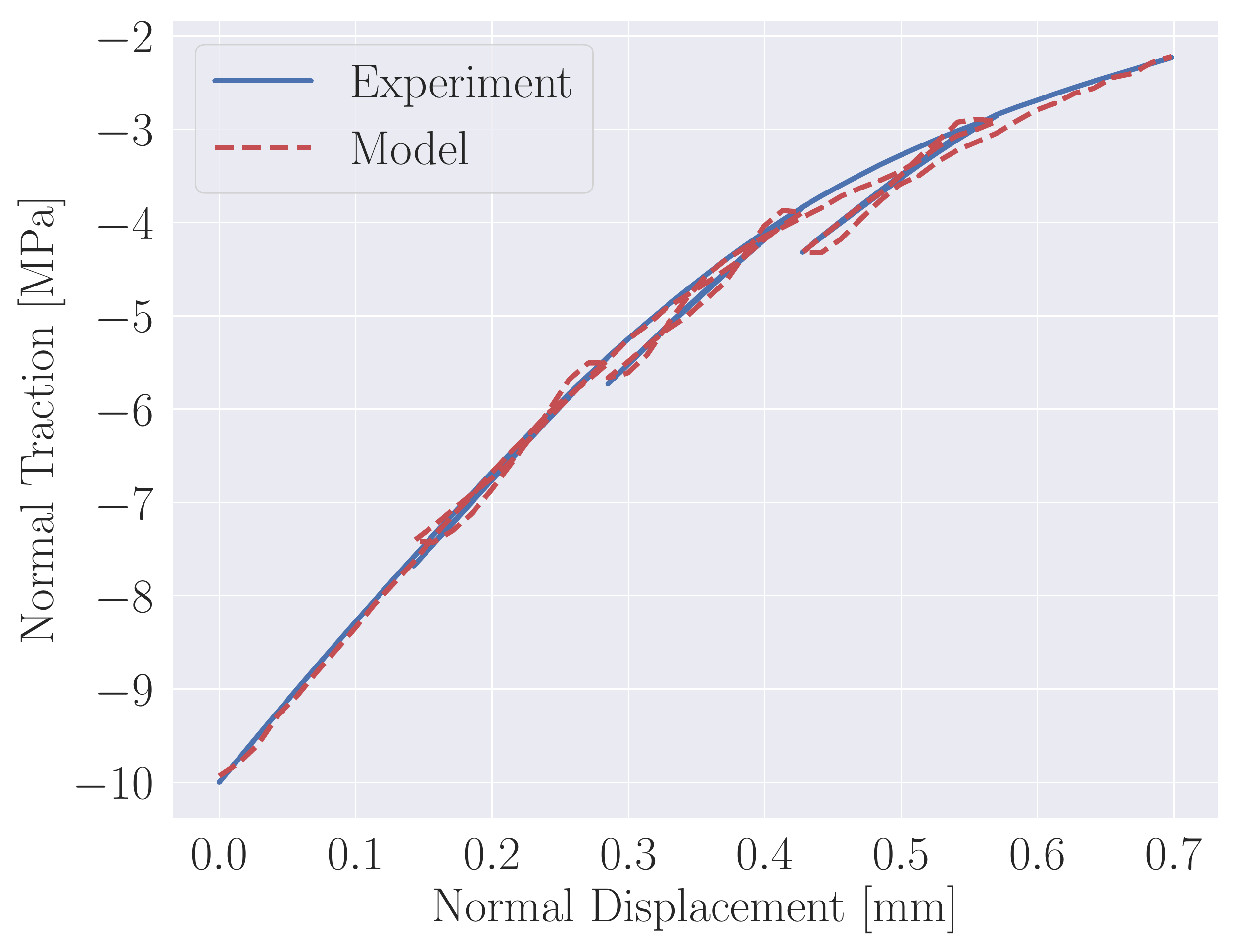}
    \includegraphics[width=0.4\textwidth]{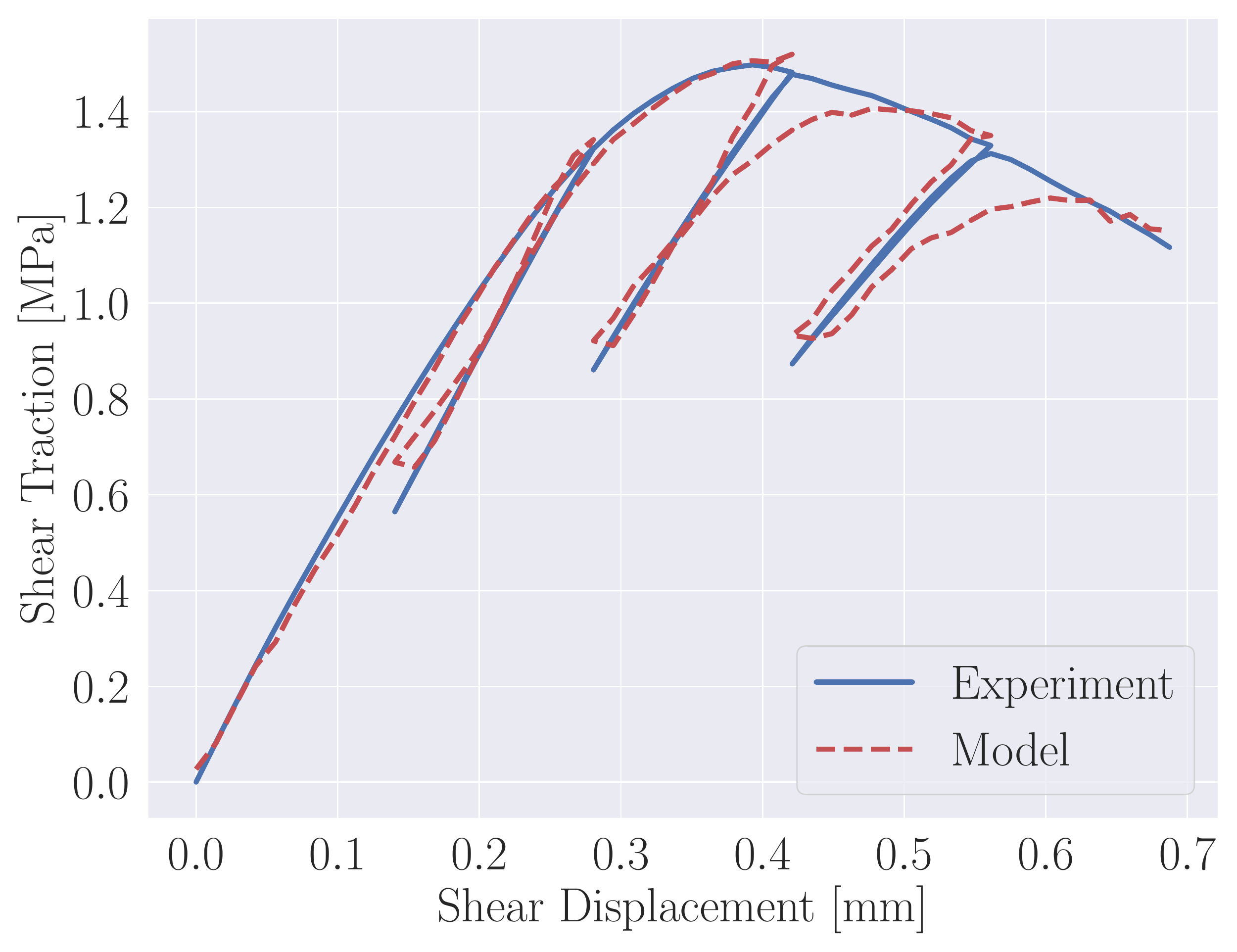}
    \caption{Comparison of traction-displacement between model and experiment in one case.}
    \label{fig::trac-sep-u-t}
\end{figure}

\begin{figure}[h!]
 \centering
 \subfigure[]
{\includegraphics[width=0.33\textwidth]{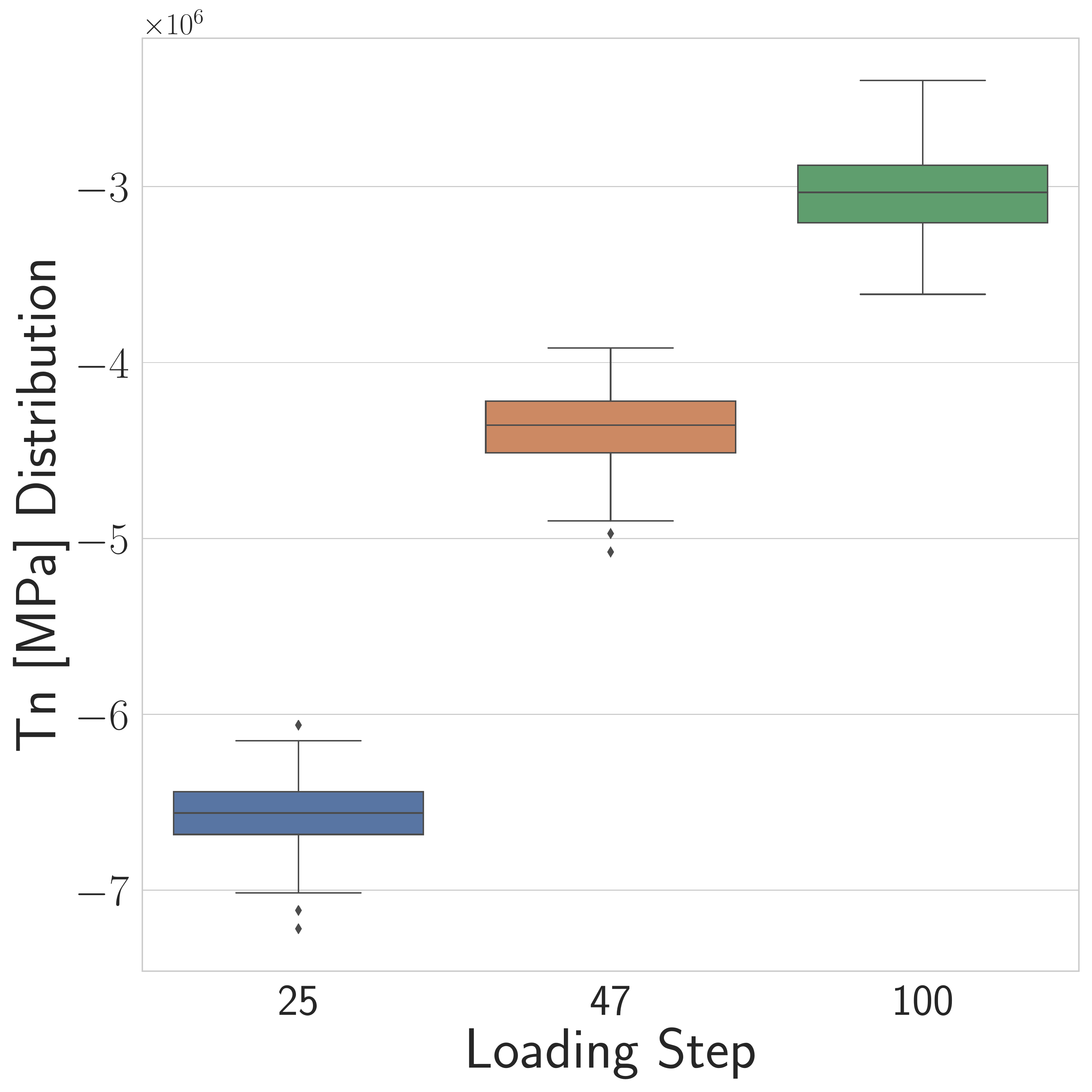}}
\hspace{0.01\textwidth}
 \subfigure[]
{\includegraphics[width=0.33\textwidth]{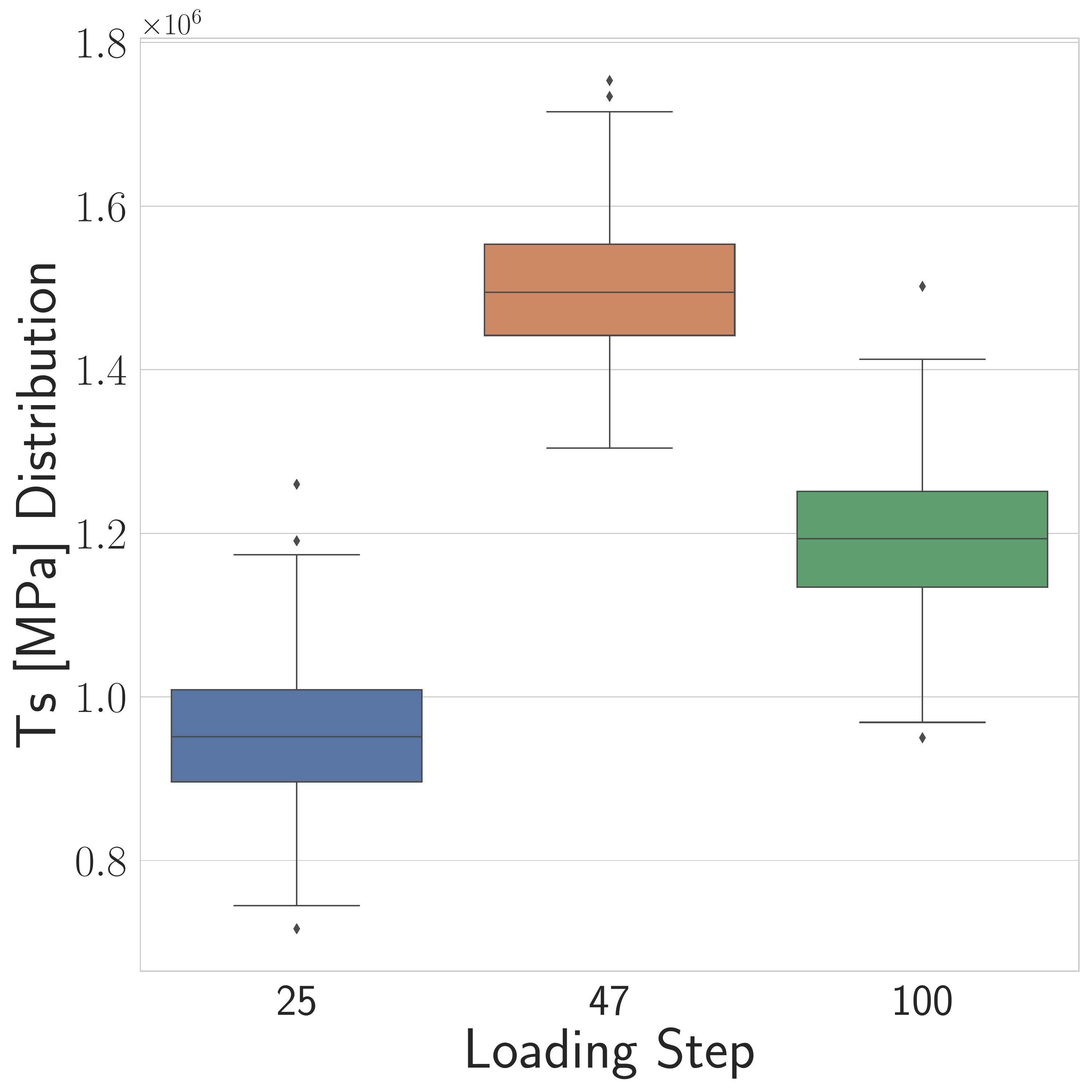}}
\hspace{0.01\textwidth}
    \caption{Box plots of density distributions of normal (a) and shear (b) traction distributions at three load steps.The top line is maximum value and the bottom line is the minimum value. The box composed of three thick lines are separately first quantile, median, third quantile} \label{fig::trac-sep-dist-traction}
\end{figure}

\begin{figure}[h!]
 \centering
 \subfigure[]
{\includegraphics[width=0.3\textwidth]{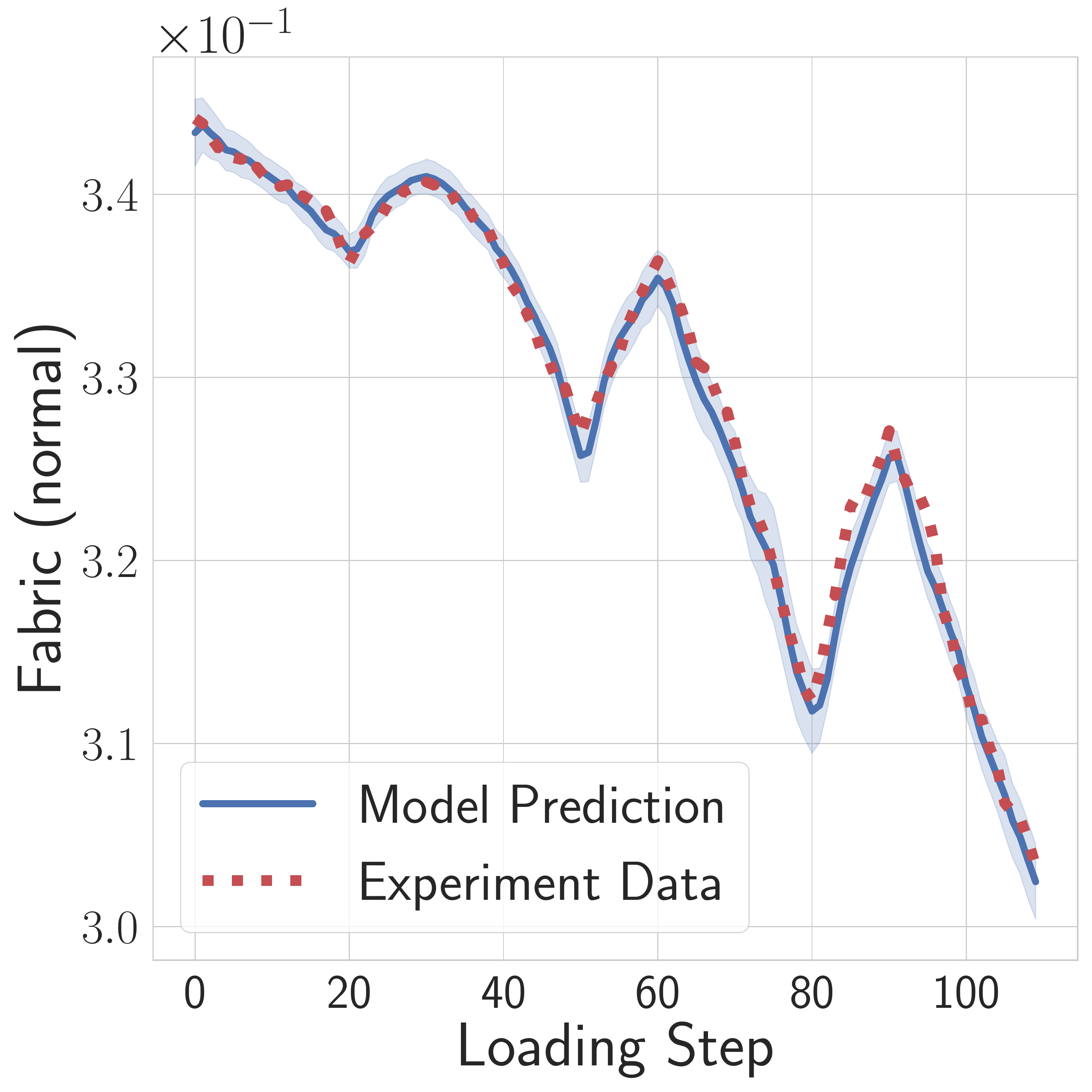}}
\hspace{0.01\textwidth}
 \subfigure[]
{\includegraphics[width=0.3\textwidth]{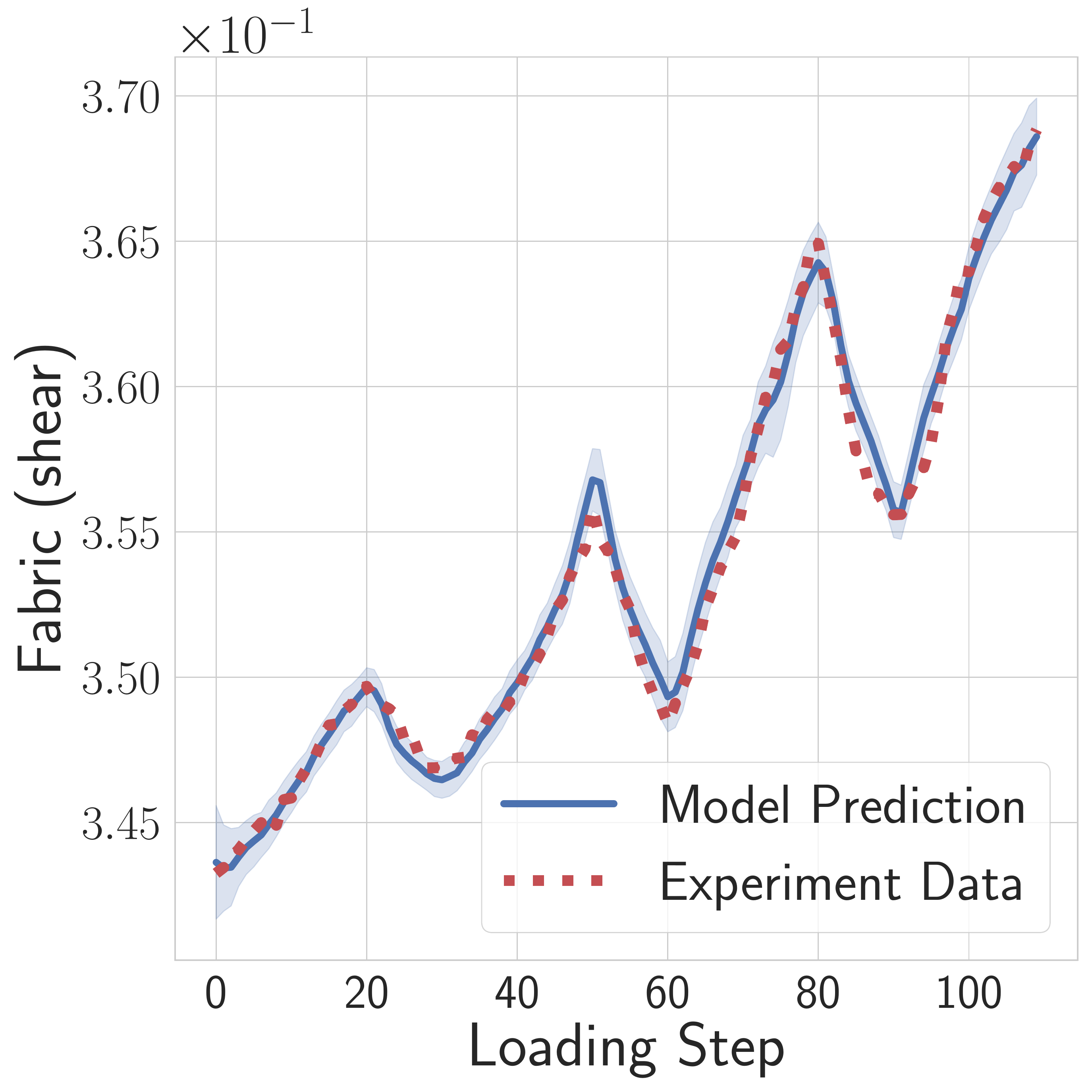}}
\hspace{0.01\textwidth}
 \subfigure[]
{\includegraphics[width=0.3\textwidth]{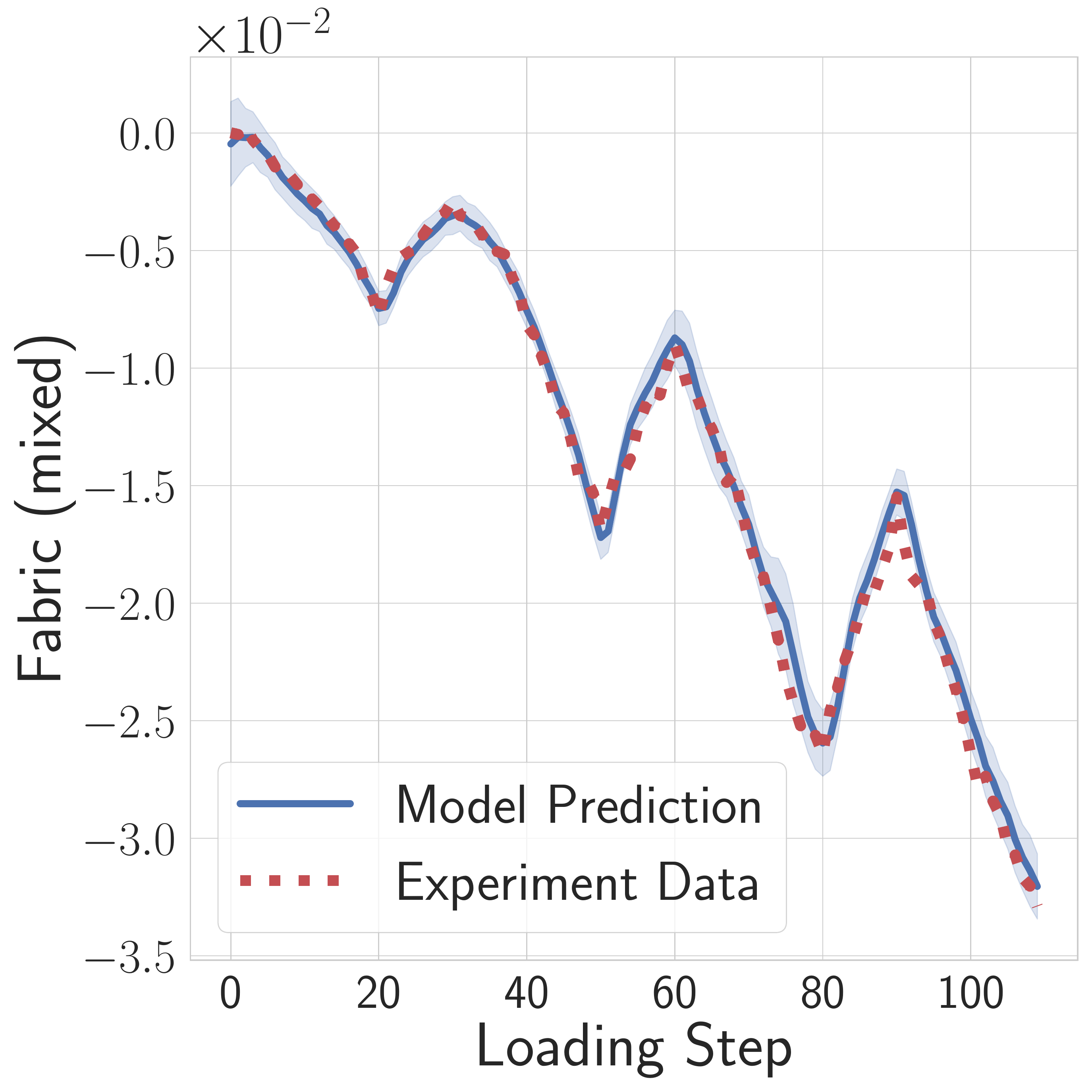}}
    \caption{Model prediction for components of symmetric fabric tensor $\tensor{A}$. Normal (a), shear (b), and mixed (c) components of fabric tensors are $A_{nn}$, $A_{ss}$, and $A_{ns}$, respectively. Shaded area includes predictions within $95\%$ confidence interval.}
    \label{fig::trac-sep-fabric}
\end{figure}

We focus on the average of model predictions in Figs. \ref{fig::trac-sep-tn-ts} and \ref{fig::trac-sep-u-t}. In Fig. \ref{fig::trac-sep-tn-ts}, we see that the initial friction angle is close to $16.7$ degree which is almost half of the inter-particle friction angle. This reduction in the overall friction angle might be due to the induced dilation in the normal displacement. Another reason could be related to initial confining pressure: the higher the confining pressure is, the lower the friction angle is. In each loading-unloading branch, the behavior is almost linear without any energy dissipation, but further loading after a level makes the behavior nonlinear. If we only follow the loading path we observe the strain-softening which is the dominant mechanism of a dense granular assemblage; see Fig. \ref{fig::trac-sep-tn-ts} and \ref{fig::trac-sep-u-t}(b). In other words, the material shows an unstable peak shear strength which is followed by a softening behavior until it reaches the critical state. The sign of changes in normal traction (Fig. \ref{fig::trac-sep-dist-traction}(a)) and shear traction (Fig. \ref{fig::trac-sep-dist-traction}(b)) are in agreement with the fabric normal (Fig. \ref{fig::trac-sep-fabric}(a)) and shear (Fig. \ref{fig::trac-sep-fabric}(b)) components, respectively. This confirms the tendency of fabric tensor to trace the load direction \citep{li2009micro,li2012anisotropic,Wang2016b}. 
Overall the proposed data-driven scheme can replicate  main features of a realistic experiment, and there exists a good agreement between the model and experiment. However, there is an issue corresponding to second loading-unloading cycle where hysteresis is predicted by the model while experiment shows almost zero energy dissipation. This is mainly due to the neural network capacity and design and can be resolved by enriching the neural network architecture with wider neurons or deeper layers or hyper-parameter tuning. Note that one needs to be aware of the over-fitting issue when the model complexity increases by increasing the number of neurons. Generally, a more complex neural network should be trained with more data.

\begin{figure}[h!]
 \centering
 \subfigure[]
{\includegraphics[width=0.33\textwidth]{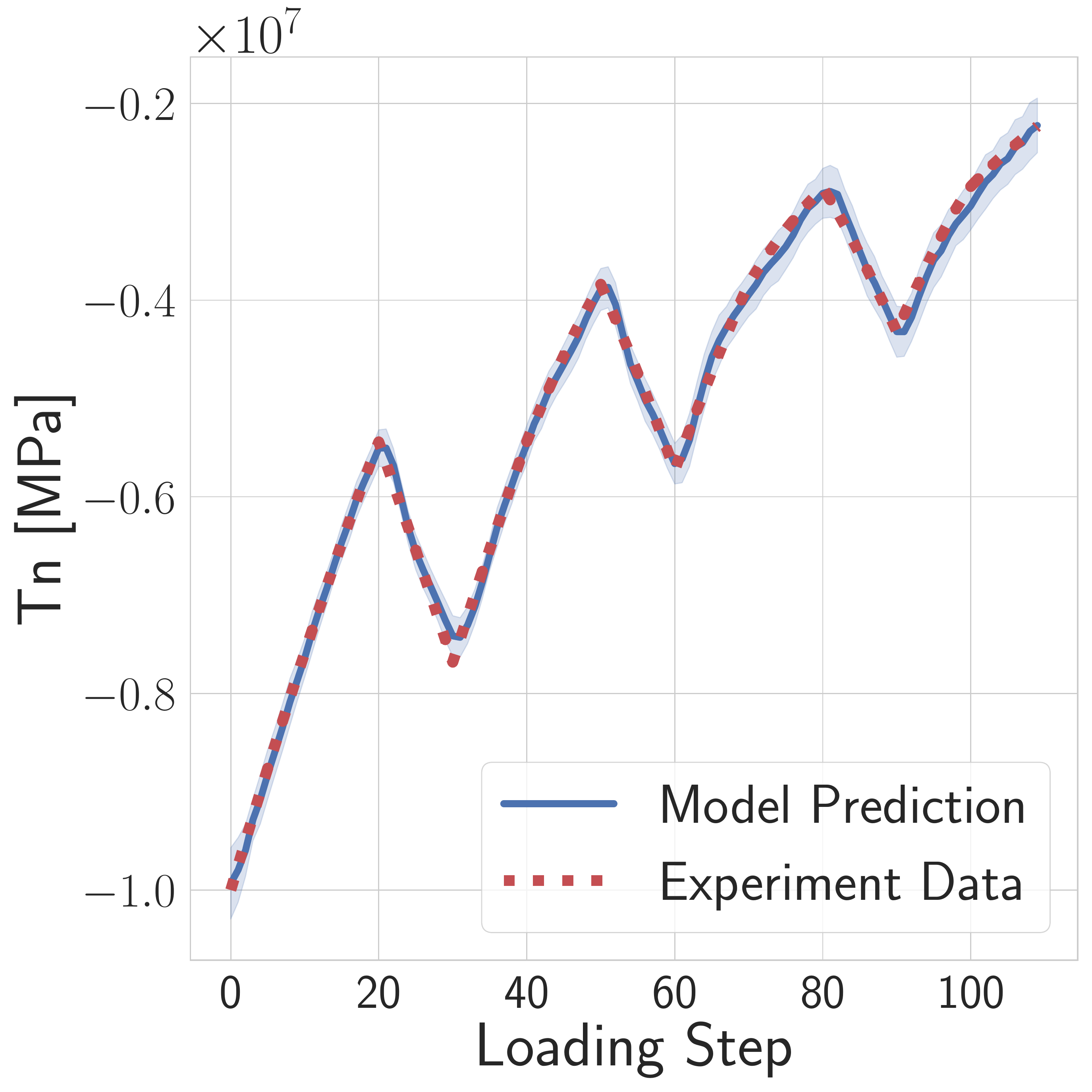}}
\hspace{0.01\textwidth}
 \subfigure[]
{\includegraphics[width=0.33\textwidth]{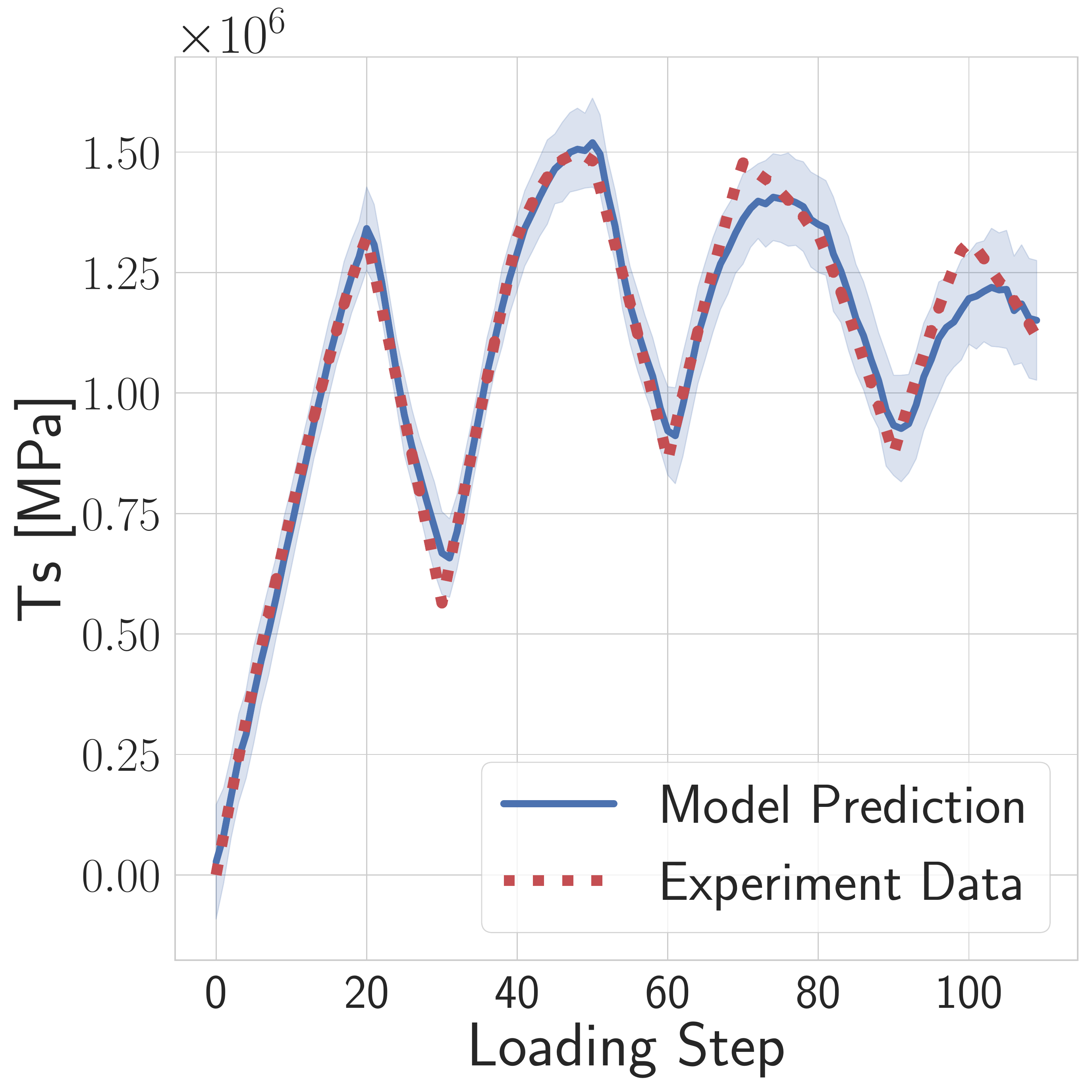}}
\hspace{0.01\textwidth}
    \caption{Model predictions for normal (a) and shear (b) traction values. Shaded area includes predictions within $95\%$ confidence interval.} \label{fig::trac-sep-traction}
\end{figure}

The uncertainty in traction vector prediction is shown in Fig \ref{fig::trac-sep-traction}. Density distributions of traction vector at three loading steps are plotted in Fig \ref{fig::trac-sep-dist-traction}. In this figure, steps $25$, $47$, and $100$ belong to the first unloading, second peak, the last peak conditions, respectively (see Fig. \ref{fig::trac-sep-traction}).
 Fig. \ref{fig::trac-sep-traction} suggests that the model is able to track the path-dependent behavior of experiments with narrow variation bands in most of the loading steps. This figure also suggests that the uncertainty for shear traction is higher than normal traction and increases at peak loads. We know that, from mechanics, the shear mode of deformation is more complex and nonlinear than the normal mode and consequently deserves higher uncertainty, which agrees with these results (also see step 47 and 100 in Fig. \ref{fig::trac-sep-dist-traction} for a quantitative comparison). At peak values, the complexity is more profound due to the cyclic loading or softening, so more uncertainty is expected. 

\begin{figure}[h!]
 \centering
 \subfigure[]
{\includegraphics[width=0.3\textwidth]{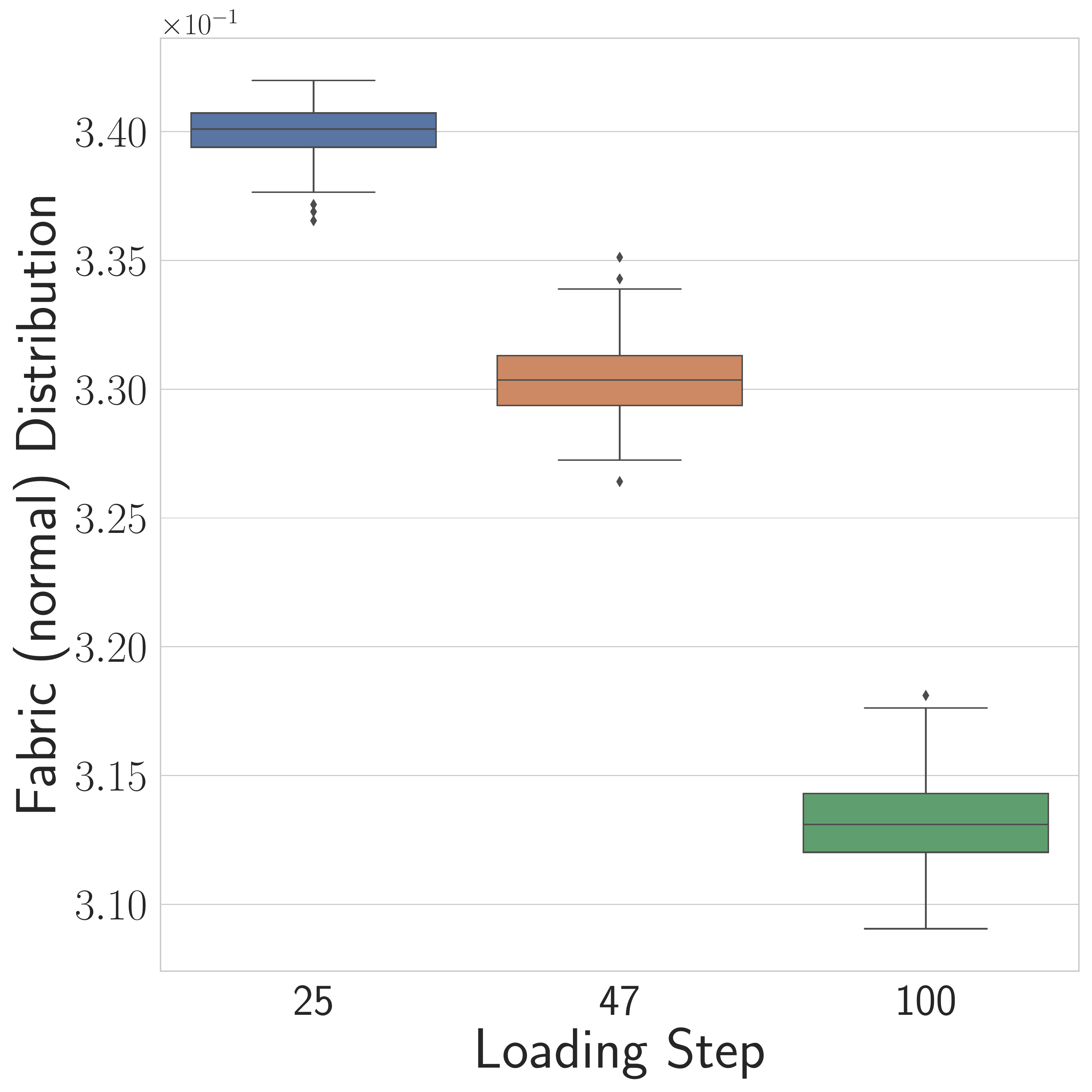}}
\hspace{0.01\textwidth}
 \subfigure[]
{\includegraphics[width=0.3\textwidth]{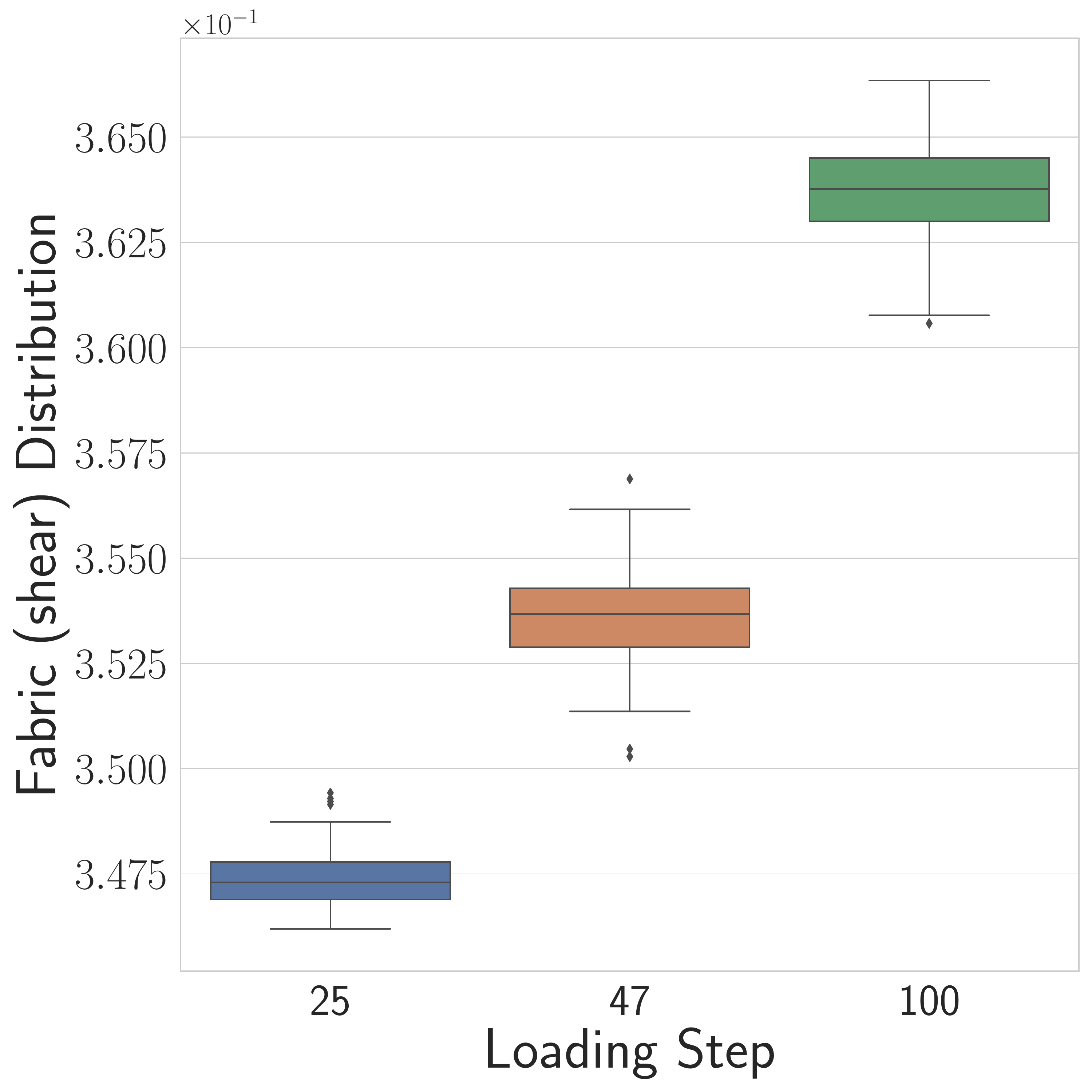}}
\hspace{0.01\textwidth}
 \subfigure[]
{\includegraphics[width=0.3\textwidth]{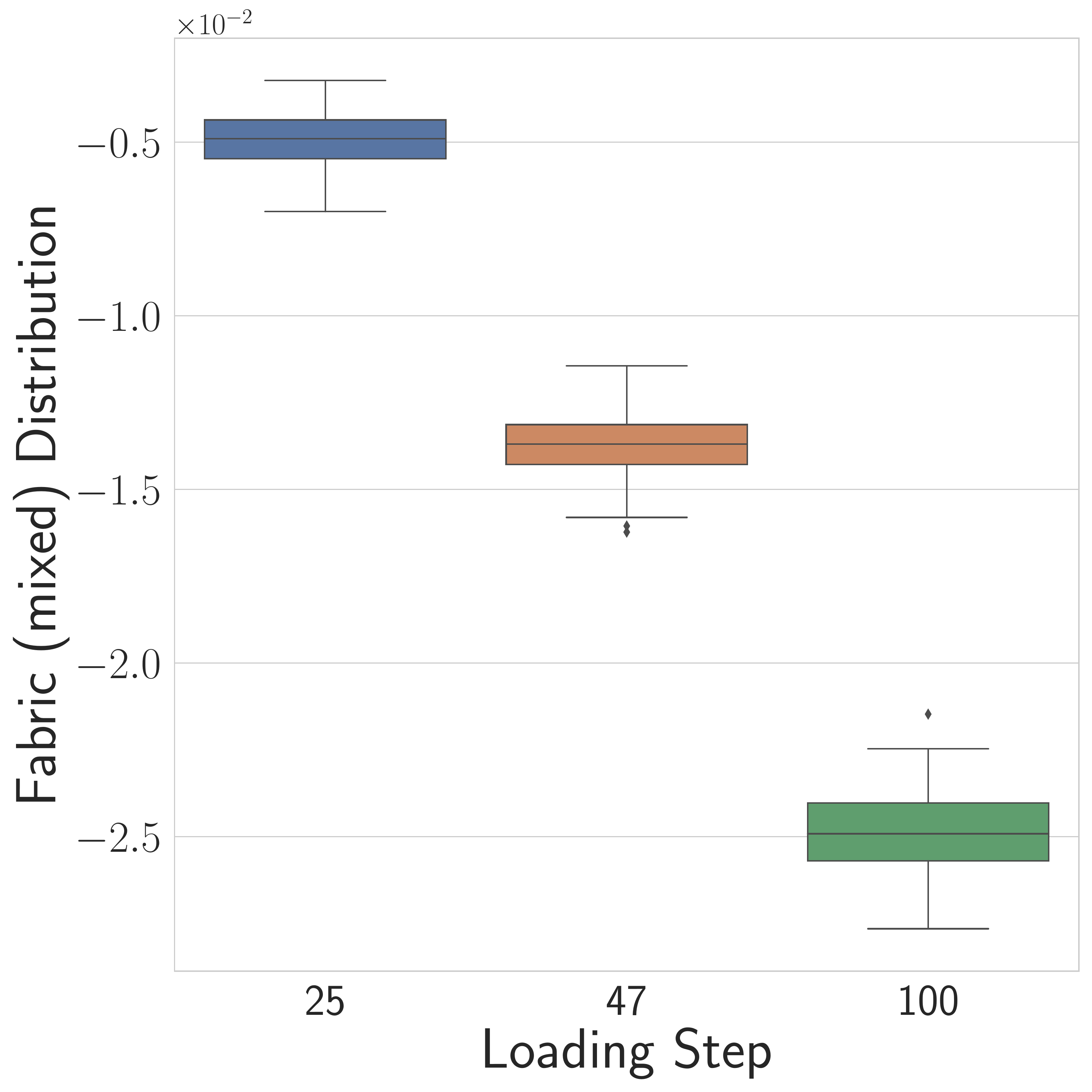}}
    \caption{Box plots of density distributions of fabric's components.}
    \label{fig::trac-sep-dist-fabric}
\end{figure}

Model prediction for fabric tensor is plotted in Fig. \ref{fig::trac-sep-fabric}. Density distributions of fabric tensor at three loading steps are plotted in Fig. \ref{fig::trac-sep-dist-fabric}. The uncertainty in fabric tensor  has narrow variation bands in most of the loading steps. Similar to the traction prediction, the uncertainties in normal,  shear, and mixed modes are higher at peak loads due to the cyclic loading or softening. However, comparing the normal, shear, and mixed modes, we do not observe significant differences in uncertainty at the three loading steps (Fig. \ref{fig::trac-sep-dist-fabric}). 
 Interestingly, we observe that traction predictions have less uncertainty at initial load steps, step 0 to 20, comparing to fabric while fabric is an intermediate node for traction prediction. This means that traction prediction is potentially less dependent on the fabric at initial loading steps and neural network weights are automatically adjusted to make predictions with high confidence as much as possible by an appropriate combination of porosity, displacement, and fabric. We observe an almost linear correlation between normal displacement jump and porosity, so we have not presented porosity prediction results due to its simplicity. Such a correlation is expected since the normal displacement is the boundary condition in this problem, and dilation is explicitly controlled during the experiment.

\subsection{Numerical Example 2: Machine Learning hypoplasticity}
\label{sec:bulkplasticity}
In the second numerical experiment, we attempt to generate a predictive surrogate model for one numerical granular assembly undergoing monotonic true triaxial compression loading. 
For convenient purpose, discrete element simulations
are used as   replacement of physical tests. These discrete element 
simulations are run via the open-source software YADE \citep{vsmilauer2010yade}. 

In total, we conduct 60 true triaxial compression tests with loading path that varying the principle stress $\sigma_{1}, \sigma_{2}$ and $\sigma_{3}$ are performed on the same numerical specimen. Before the shearing phase, the material is subjected to hydrostatic  loading to compress the assembly hydrostatically to reach the initial confining pressure. Following this step, a vertical compression or extension or a change of the applied tractions on the side walls are prescribed to generate different stress paths. To facilitate third-party validation and re-production of the simulation results, the data used for the causal discovery are given access to the public via Mendeley Data \citep{hypopl_data}. 

\subsubsection{Data-driven causal relations of granular matter}
\begin{figure}[h!]
 \centering
\includegraphics[width=0.95\textwidth]{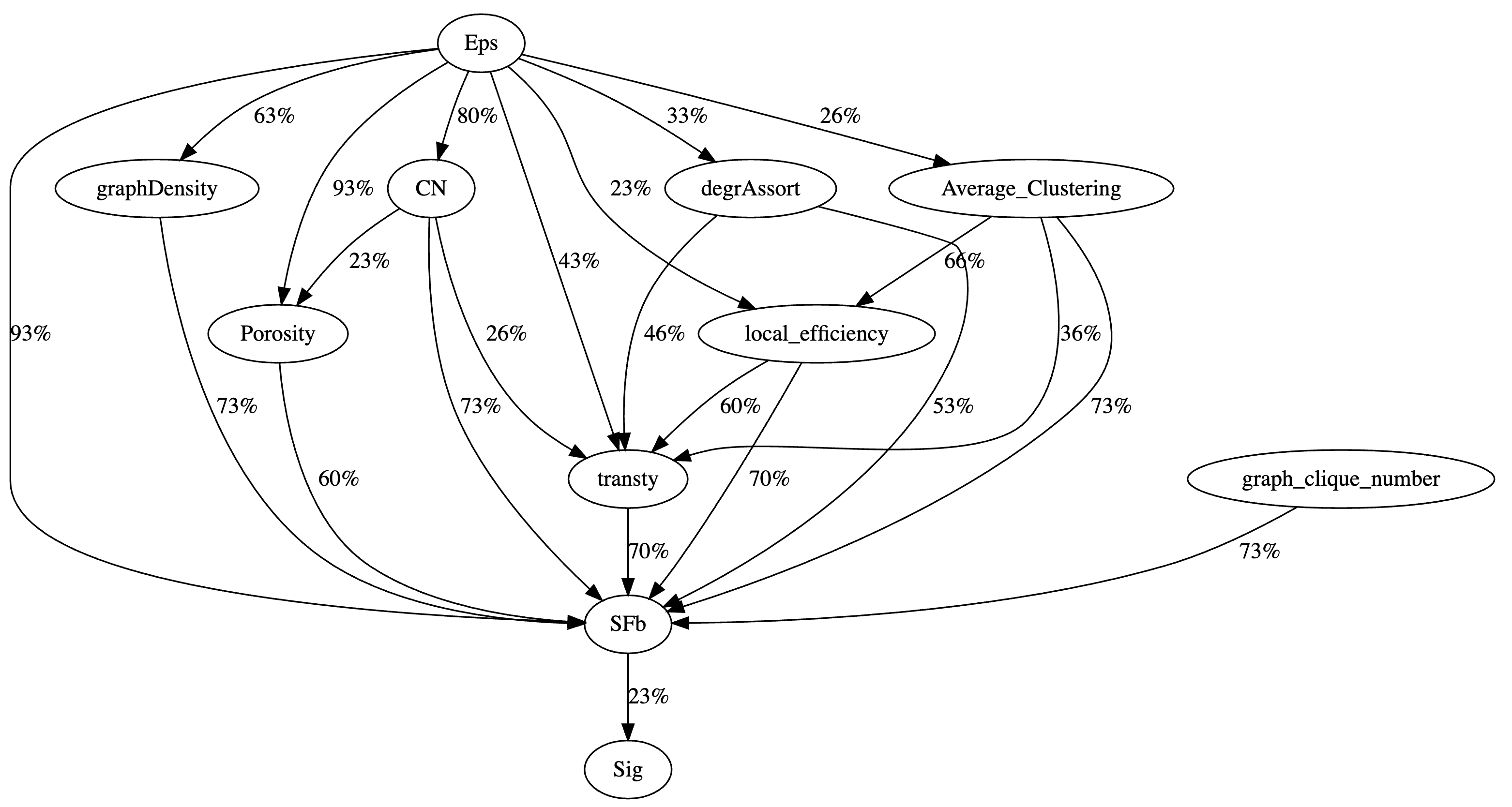}
    \caption{Final Causal graph for the hypoplasticity relations deduced from time-history of strain, stress, and 9 other measures of microstructural and topological properties.  
The number on each edge   represents the edge inclusion probabilities 
among all possible causal relations from the training data sets. 
}
    \label{fig:hypoplasticity}
\end{figure}

Fig. \ref{fig:hypoplasticity} shows the final causal graph of the causal discovery algorithm applied to the true triaxial test data generated from discrete element simulations. The number on each edge represents the edge inclusion probability   in the calibration experimental data sets. 

The causal discovery driven by the small set of calibration data reveals a number of key observations that are worth-noticing. First, 
the causal discovery algorithm does re-discover the conventional wisdoms, such as the fact that 1) the changes of coordination number is due to the expansion of the void space; 2) both the coordination number and the porosity changes may cause changes on the fabric tensors; and 3) the dominate role of the strong fabric tensors on the resultant stress. These observations are consistent with  previous 
findings in a number of discrete element simulation literature \citep{wang2017evolution, sun2013multiscale, kuhn2015stress, shi2018noncoaxiality} and the anisotropic critical state theory \citep{li2012anisotropic, fu2015relationship, zhao2013unique}. 

In addition to the rediscoveries of known knowledge, the causal discovery algorithm also finds a few causal relationships not 
known in the existing literature (to the best knowledge of the authors).
For instance, the causal discovery algorithm is able to 
establish a casual relationship that changes in average clustering coefficient may affect the local efficiency of the particle connectivity, whereas the degree of assortativity coefficient, a measure of the similarity of the connections of the graphs, may affect the graph transitivity. Interestingly, the causal discovery algorithm also finds that changes of the strong fabric tensor may be caused by changes of the strain (93\%), porosity (60\%), coordination number (73\%), graph density (73\%), local efficiency  (70\%)  and graph transitivity (70\%), degree of assortativity (70\%) as well as graph clique number (73\%). 
This discovery indicates that the changes of the strong fabric tensors 
are driven by the changes of the underlying connectivity topology 
and the volume changes of the void space. 

Furthermore, another interesting discovery is that the changes of the stress tensor is only conditionally independently caused by the changes of the strong fabric tensor. This result is consistent with the previous finding 
of 2D granular materials reported in \citet{shi2018noncoaxiality} where it is shown that (1) the principal direction of the strong fabric tensor (but not necessarily other fabric tensors) is coaxial with the homogenized Cauchy stress, and (2) the fabric tensor and stress tensor are related by a scalar coefficient that may vary according to the mean pressure.

\subsubsection{Predictions based on discovered causal relationships}
Here we investigate the accuracy, robustness and the limitations of the machine learning predictions generated based on the deduced causal relations. For comparison purposes, we complete the training of two sets of neural networks -- one employs the newly discovered causal relationships into the predictions, another one employs only the 
strain, fabric tensor and porosity to predict the stress. The latter 
neural network is then used as a controlled experiment for the former one. The supervised learning procedures used to train the two models 
are identical.

\begin{figure}[h!]
 \centering
\includegraphics[width=0.45\textwidth]{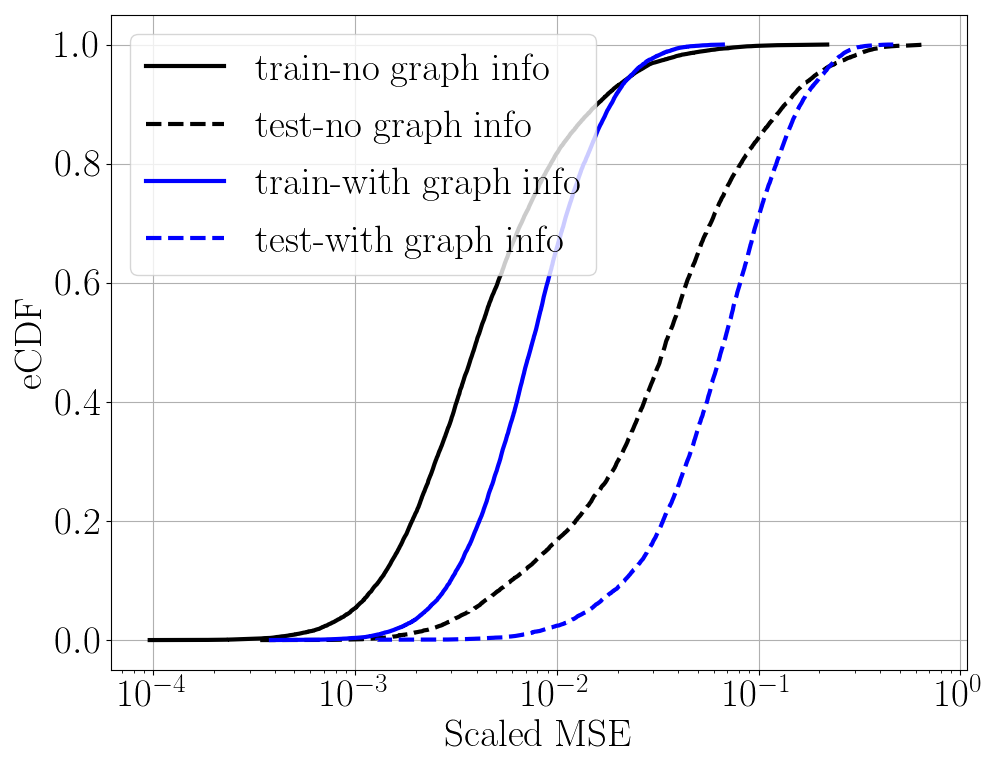}
    \caption{Empirical Cumulative Distribution Function (eCDF) for prediction on training data sets and mean value of predictions on test data sets. There are 60 simulations, 30 used for calibration and 30 for blind forward testing. Blue curves are predictions made from neural networks generated according to the causal graph, black curve is the control experiment counterpart generated from predictions that takes on strain fabric tensor and porosity as inputs to predict Cauchy stress.}
    \label{fig:compareECDF}
\end{figure}

We first do not introduce the usage of dropout layer in the GRU and hence the dropout rate is zero. 
The hyperparameters are obtained from repeated trial-and-errors and they are summarized in Table \ref{tab:hyperparameters}. All the sub-graph predictions, regardless of the number of input variables, are trained by the neural network with the identical architecture listed in Table \ref{tab:hyperparameters}. 
After the predictions, we conduct a cross-validation in which the trained neural networks are tasked to predict both the homogenized Cauchy stress obtained from the calibration and testing simulation data. The results are shown in Fig. \ref{fig:compareECDF}. 
Unlike the traction-separation law examples, the predicted stress-strain curves for the true triaxial test exhibit profound over-fitting regardless of whether the additional graph metrics are used for the predictions. 

\begin{table}[h!]
\centering
 \begin{tabular}{l | l  | l } 
NN setting description & Abbreviation & Values \\
\hline
Neuron type subset & $NeuronType$ & GRU\\
Hidden layers subset & $numHiddenLayers$ & $3$\\
Number of neurons per layer  & $numNeuronsPerLayer$ & 32 \\
Dropout rate subset  & $DropOutRate$ & $0.0$ \\
Optimizer type  subset  & $Optimizer$ &  Adam \\
Activation functions subset  & $Activation$ &  $relu$\\
Batch sizes subset  & $BatchSize$ & 128 \\
Minimum Learning rate & $ReduceLROnPlateau$ & 0.95
\end{tabular}
 \caption{Hyperparameters used to train the neural network}
   \label{tab:hyperparameters}
\end{table}

The roughly 2-order of difference in stress predictions suggests that 
either regularization strategy or more data is needed to 
circumvent the mismatch of accuracy on the calibration and blind prediction data. Notice that expanding the data set is not difficult for 
discrete element simulations, it is certainly  very difficult 
to conduct 60 true triaxial tests physically in a typical laboratory. 
As such, the results indicate the difficulty to create forecast engine 
to predict stress responses for unconventional stress paths 
even when the simulations are free of the issues, noises and errors 
exhibited in physical experiments. 

\begin{figure}[h!]
 \centering
 \subfigure[Calibration No. 23 (TXE)]
 {\includegraphics[width=0.45\textwidth]{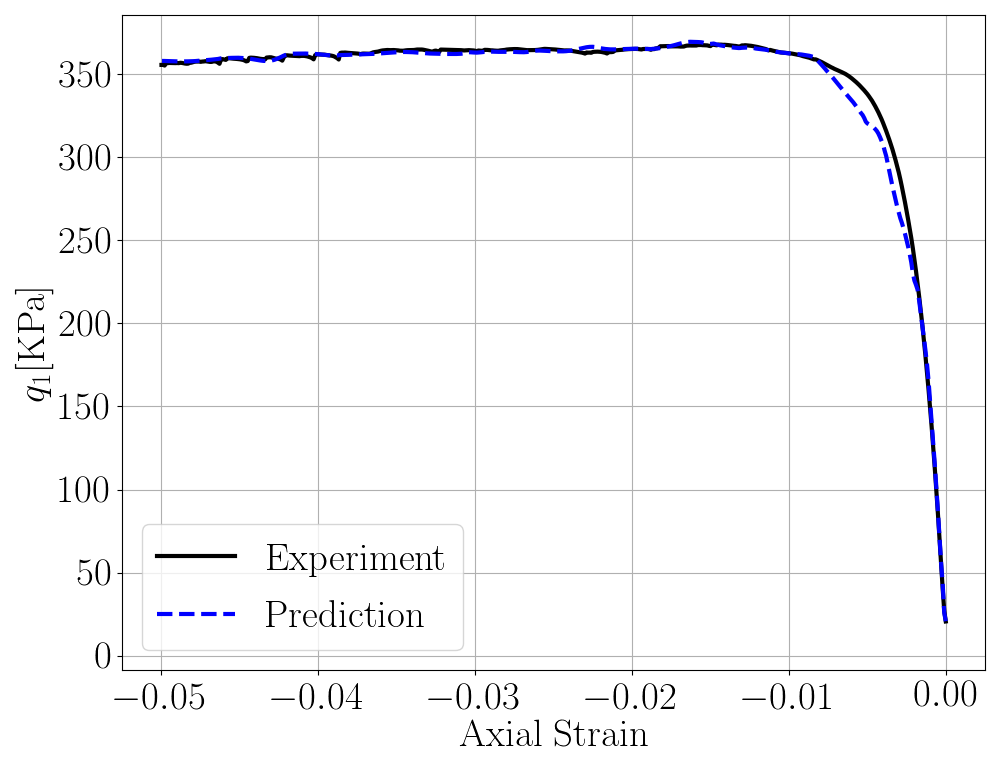}}
\hspace{0.01\textwidth}
 \subfigure[Calibration No. 29 (TXE)]
{\includegraphics[width=0.45\textwidth]{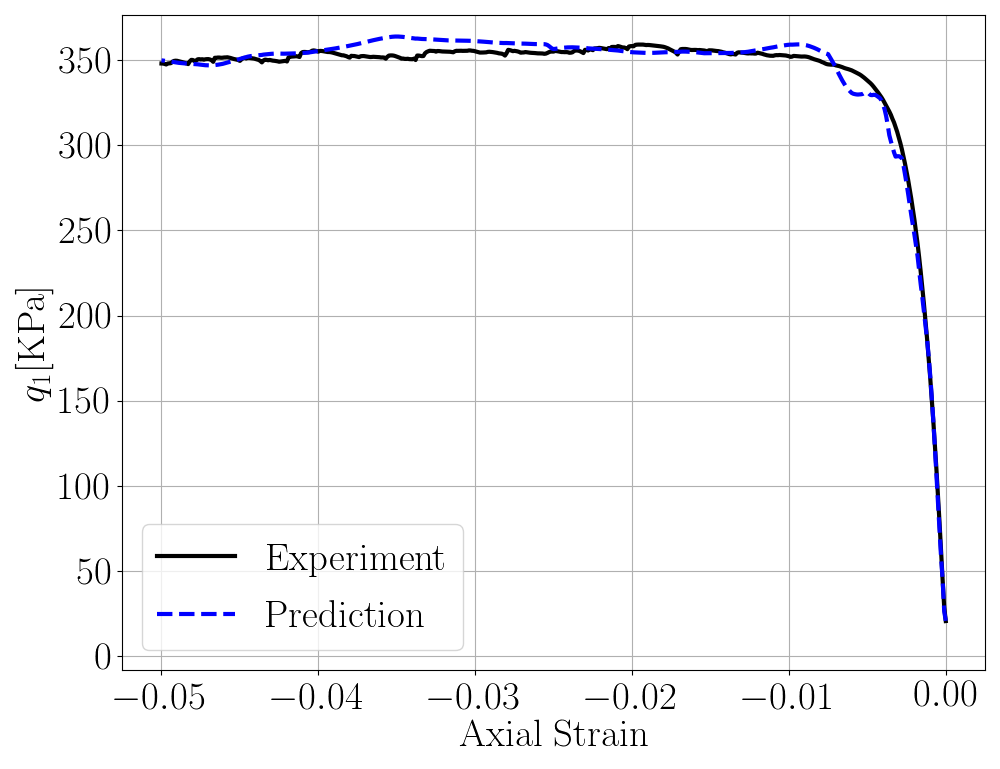}}
\hspace{0.01\textwidth}
 \subfigure[Blind Test  No. 50 (TXC)]
{\includegraphics[width=0.45\textwidth]{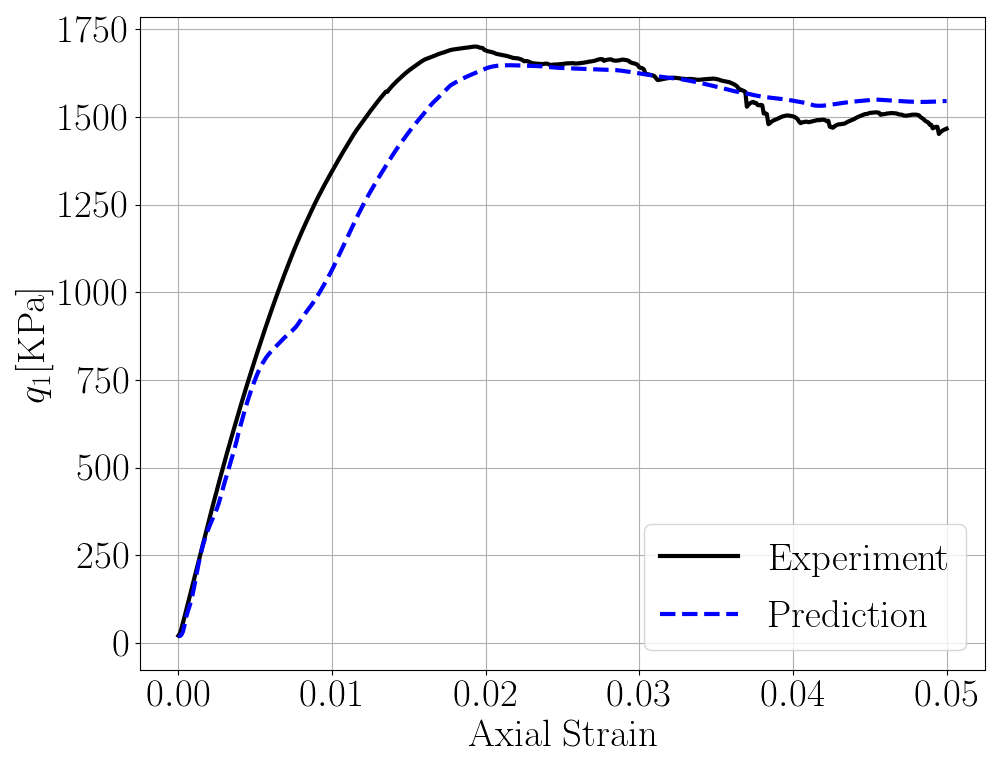}}
\hspace{0.01\textwidth}
 \subfigure[Blind Test  No. 56 (TXC)]
{\includegraphics[width=0.45\textwidth]{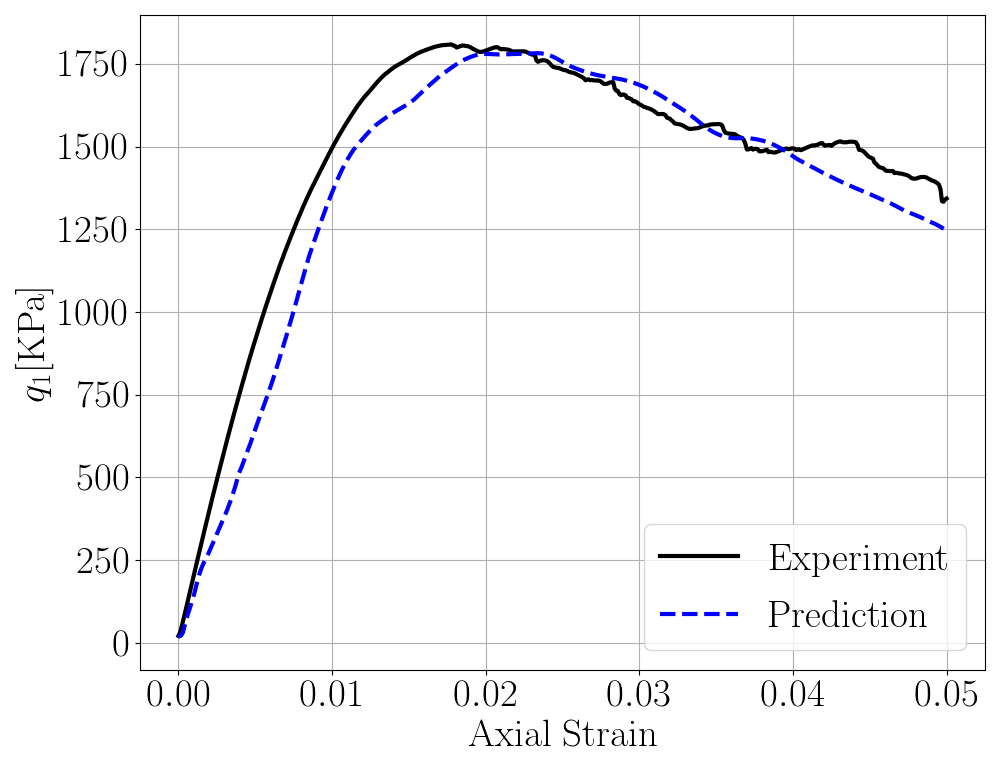}}
    \caption{Difference between the major and minor principal stress vs. axial strain. Compressive strain has positive sign convention.}
    \label{fig:q1}
\end{figure}

\begin{figure}[h!]
 \centering
 \subfigure[Calibration No. 23 (TXE)]
 {\includegraphics[width=0.45\textwidth]{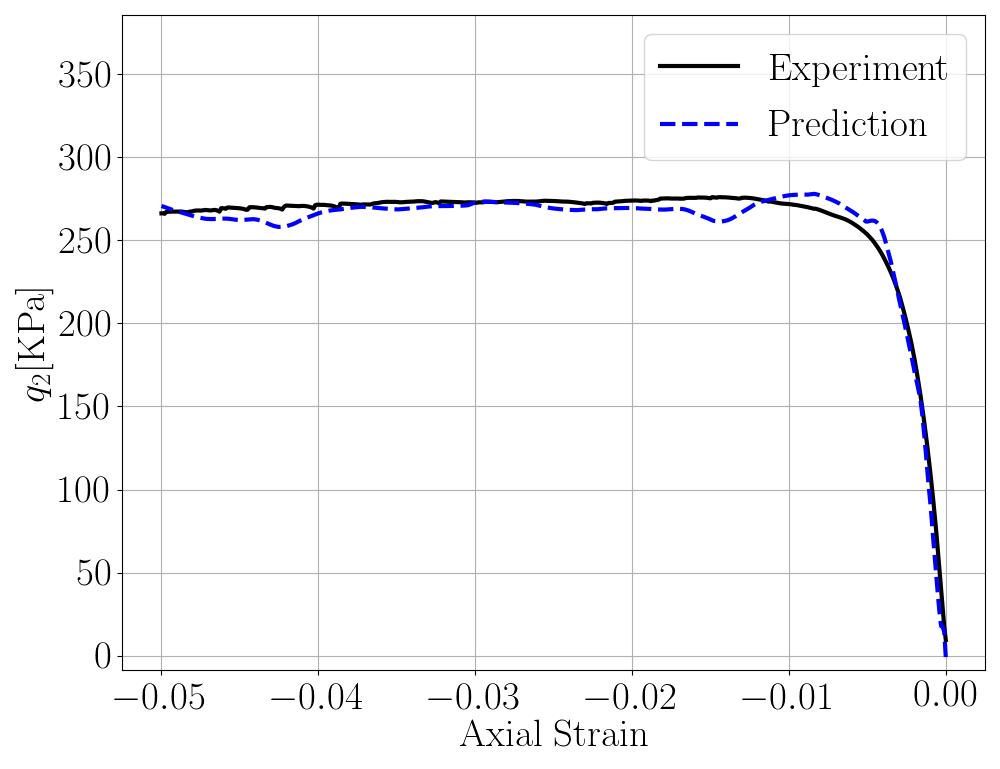}}
\hspace{0.01\textwidth}
 \subfigure[Calibration No. 29 (TXE)]
{\includegraphics[width=0.45\textwidth]{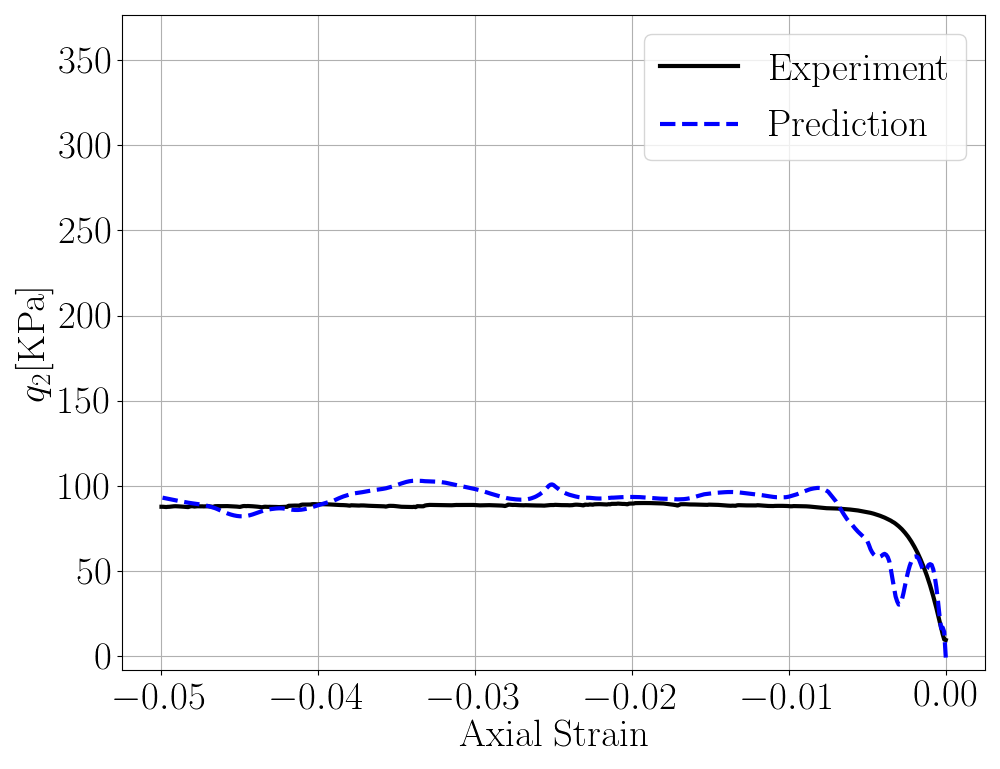}}
\hspace{0.01\textwidth}
 \subfigure[Blind Test  No. 50 (TXC)]
{\includegraphics[width=0.45\textwidth]{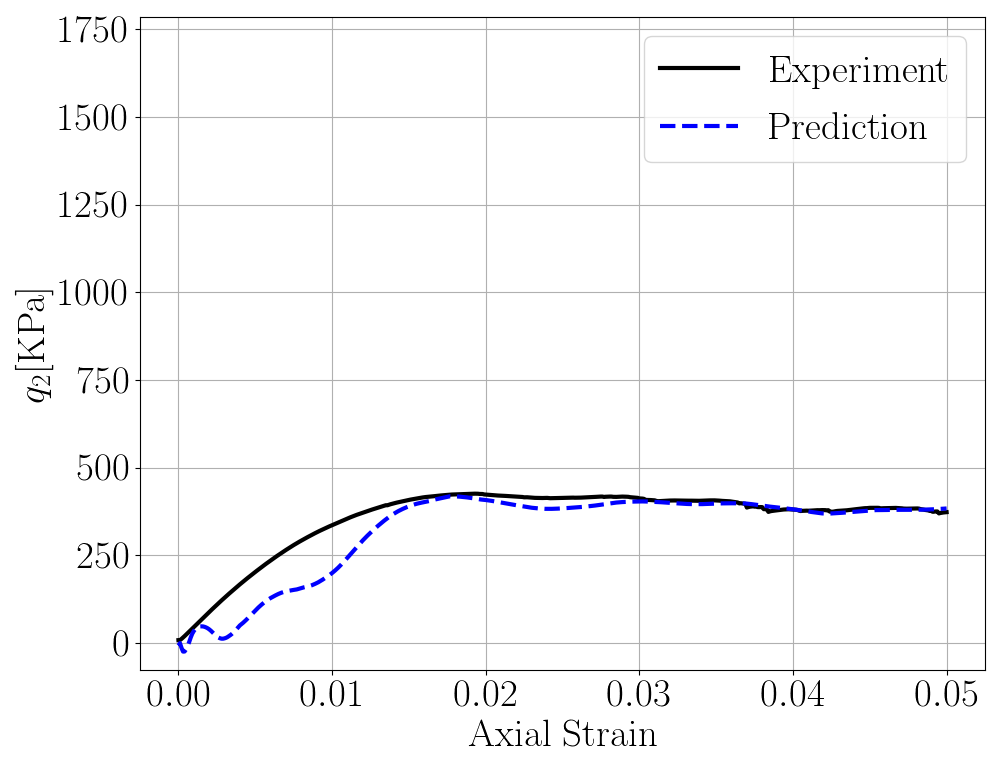}}
\hspace{0.01\textwidth}
 \subfigure[Blind Test No. 56 (TXC)]
{\includegraphics[width=0.45\textwidth]{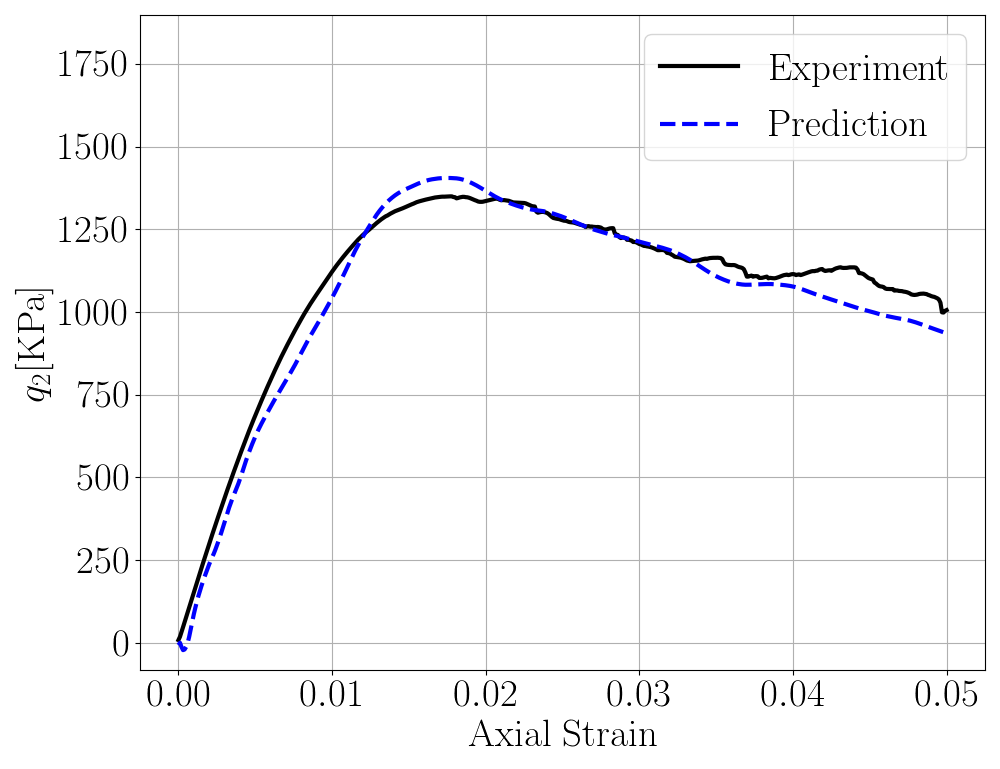}}
    \caption{Difference between the major and immediate principal stress vs. axial strain. Compressive strain has positive sign convention.}
    \label{fig:q2}
\end{figure}

\begin{figure}[h!]
 \centering
 \subfigure[Calibration No. 23 (TXE)]
 {\includegraphics[width=0.45\textwidth]{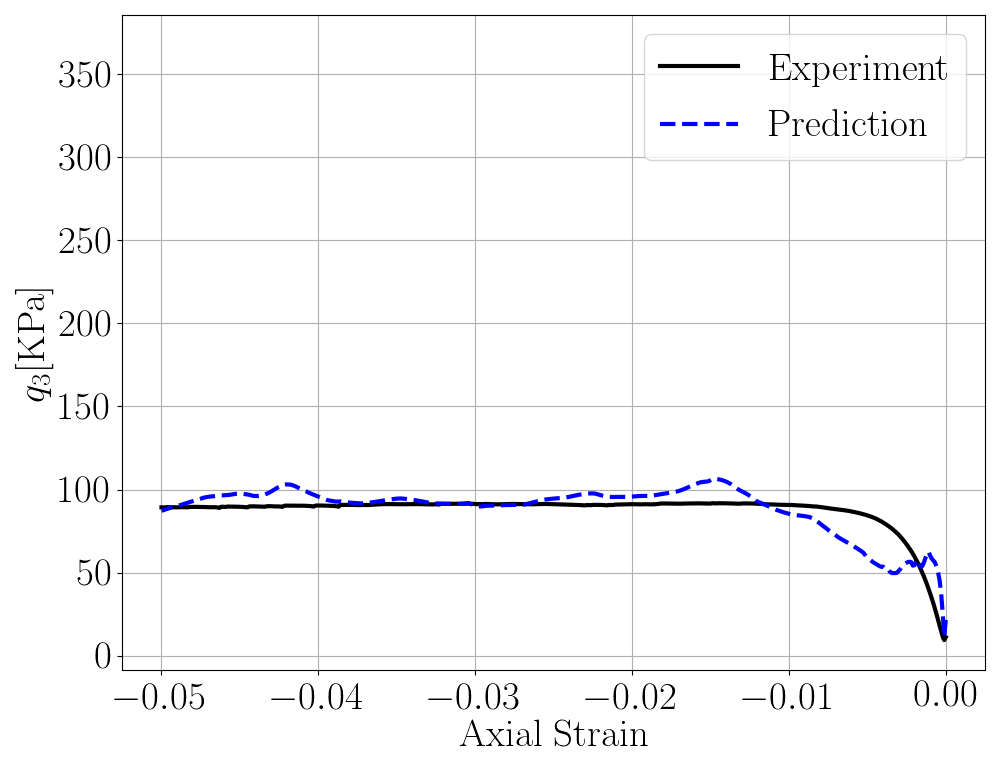}}
\hspace{0.01\textwidth}
 \subfigure[Calibration No. 29 (TXE)]
{\includegraphics[width=0.45\textwidth]{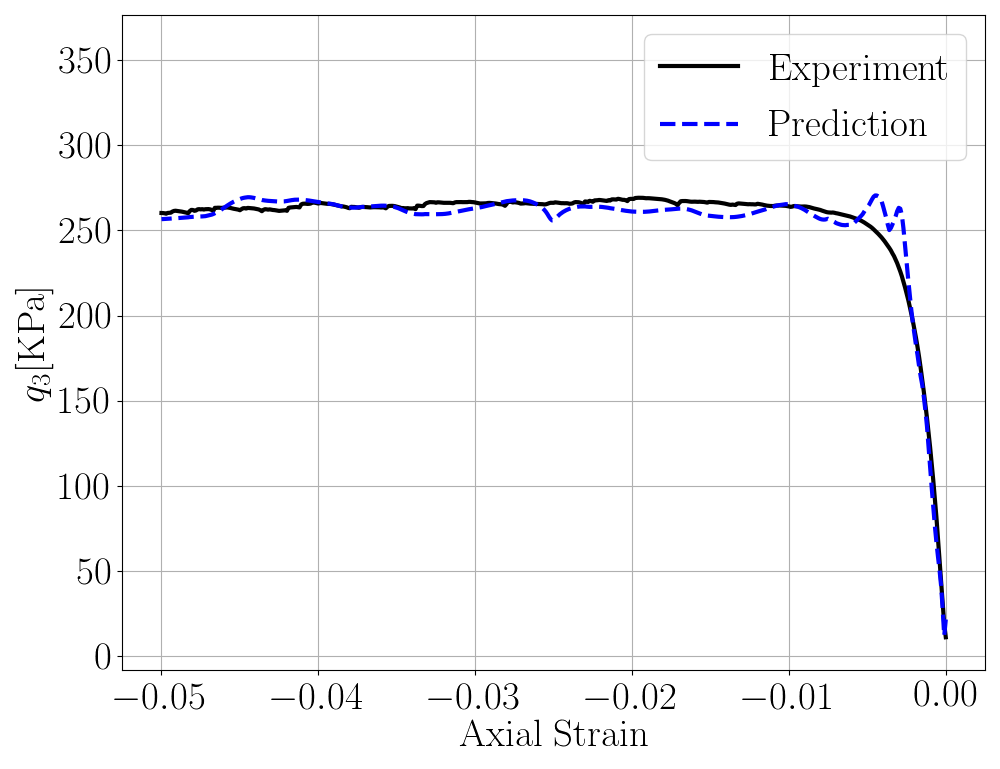}}
\hspace{0.01\textwidth}
 \subfigure[Blind Test  No. 50 (TXC)]
{\includegraphics[width=0.45\textwidth]{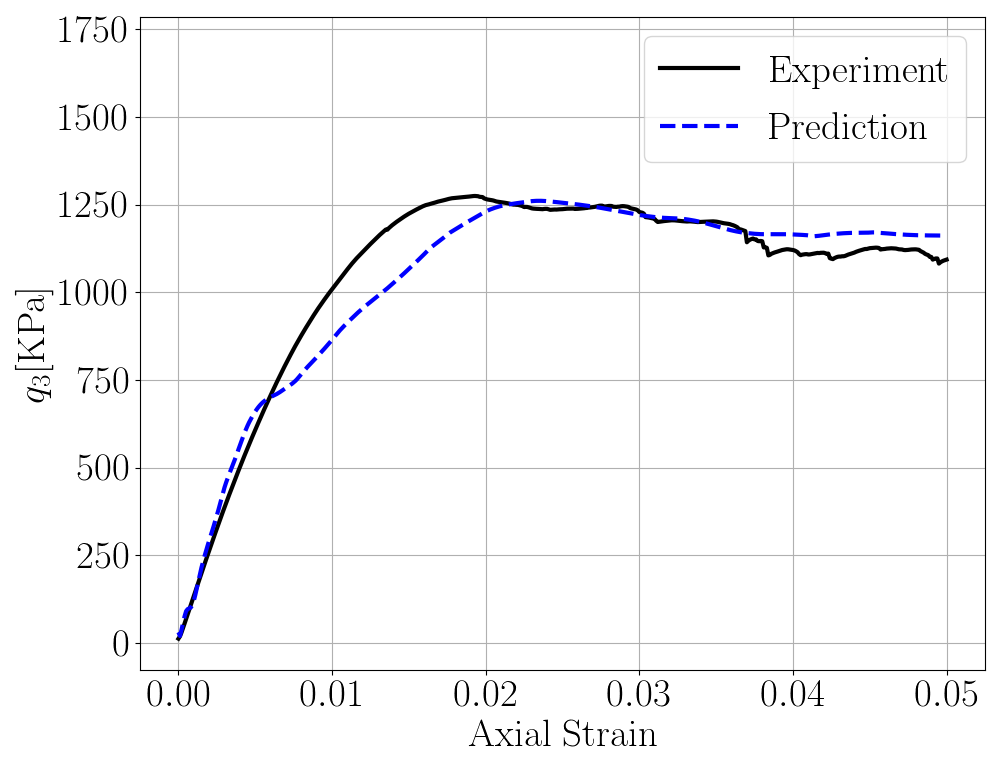}}
\hspace{0.01\textwidth}
 \subfigure[Blind Test  No. 56 (TXC)]
{\includegraphics[width=0.45\textwidth]{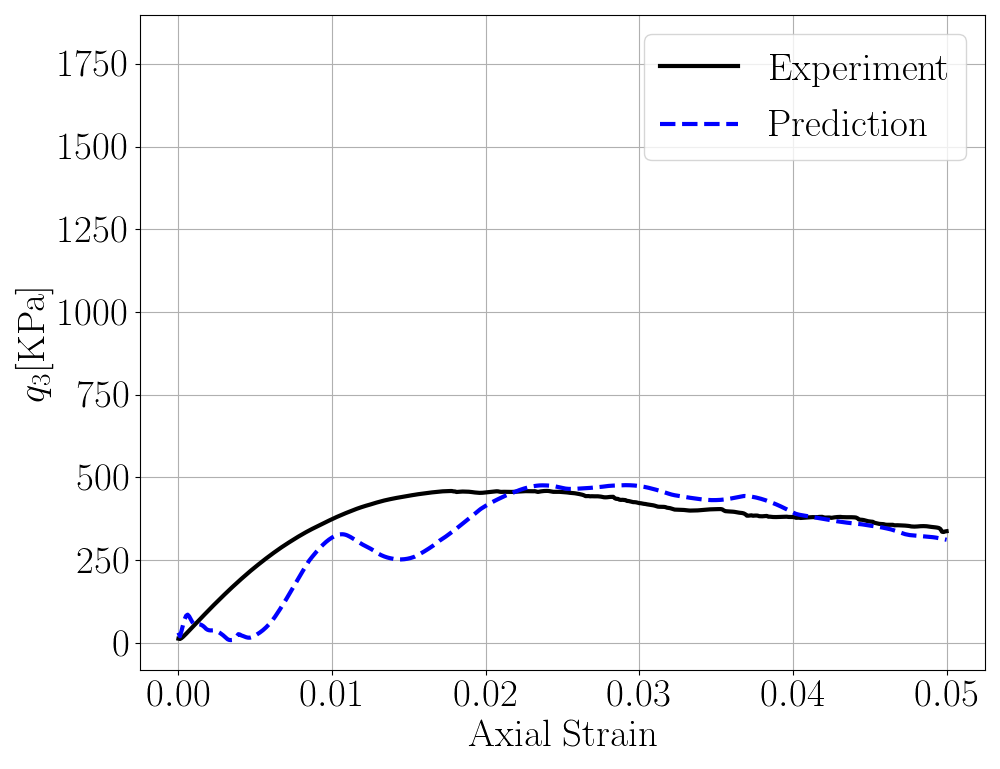}}
    \caption{Difference between the immediate and minor principal stress vs. axial strain. Compressive strain has positive sign convention.}
    \label{fig:q3}
\end{figure}

Interestingly, the predictions from the neural network
with the new graph measures do not help significantly 
on the mean errors of  predictions. However, a closer 
examination of the tail of the eCDF on the two testing curves in Figure \ref{fig:compareECDF} does indicate that the neural network armed with the new knowledge produces a less catastrophic worst-case scenario. 
Figures \ref{fig:q1}, \ref{fig:q2}, and \ref{fig:q3} show the predicted 
principal stress difference against the benchmark data, in which $q_1=\sigma_1 - \sigma_2$, $q_2=\sigma_1 - \sigma_3$, and $q_3=\sigma_2 - \sigma_2$ where $\sigma_1 \ge \sigma_2 \ge \sigma_3$ are principal stresses. In both the calibration and the testing cases, the discrepancy of the principal stress difference are minor. 

For brevity, we do not intend to present all the 30 forward prediction results. Here, we pick three samples, two calibrations  (Test No. 23 and 29) and two blind predictions (Test No. 50 and 56) for close examination.
Figs. \ref{fig:q1}, \ref{fig:q2}, and \ref{fig:q3}  compare the difference of the three principal stress inferred from the recurrent neural network 
and obtained from discrete element simulations. In these figures, 
TXC and TXE denote triaxial compression and extension tests.
For simplicity, this test does not contain cyclic loading, as a result, the prediction task is much simpler. Nevertheless, despite of the relatively small data set, the trained neural networks in the causal graph is capable of predicting important characteristics, such as hardening/softening properly. The predictions also exhibit more fluctuation, which is undesirable. However, this can be presumably suppressed with a different set of activation functions and other regularization strategies. 

\begin{figure}[h!]
 \centering
 \subfigure[Calibration No. 23 (TXE)]
 {\includegraphics[width=0.45\textwidth]{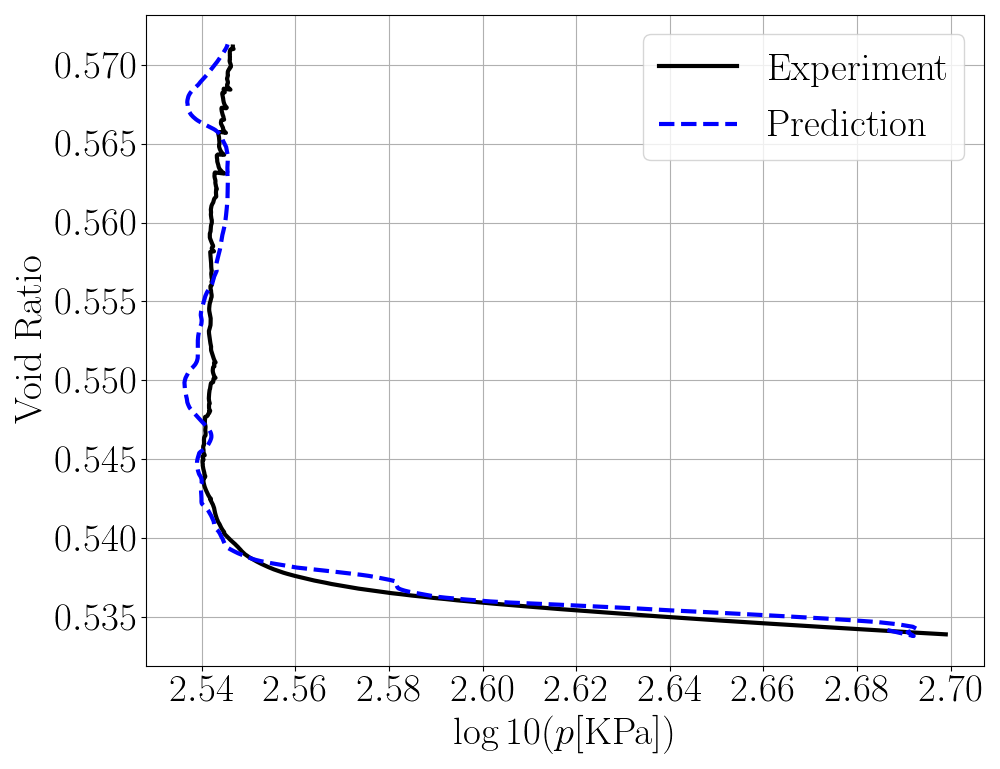}}
\hspace{0.01\textwidth}
 \subfigure[Calibration No. 29 (TXE)]
{\includegraphics[width=0.45\textwidth]{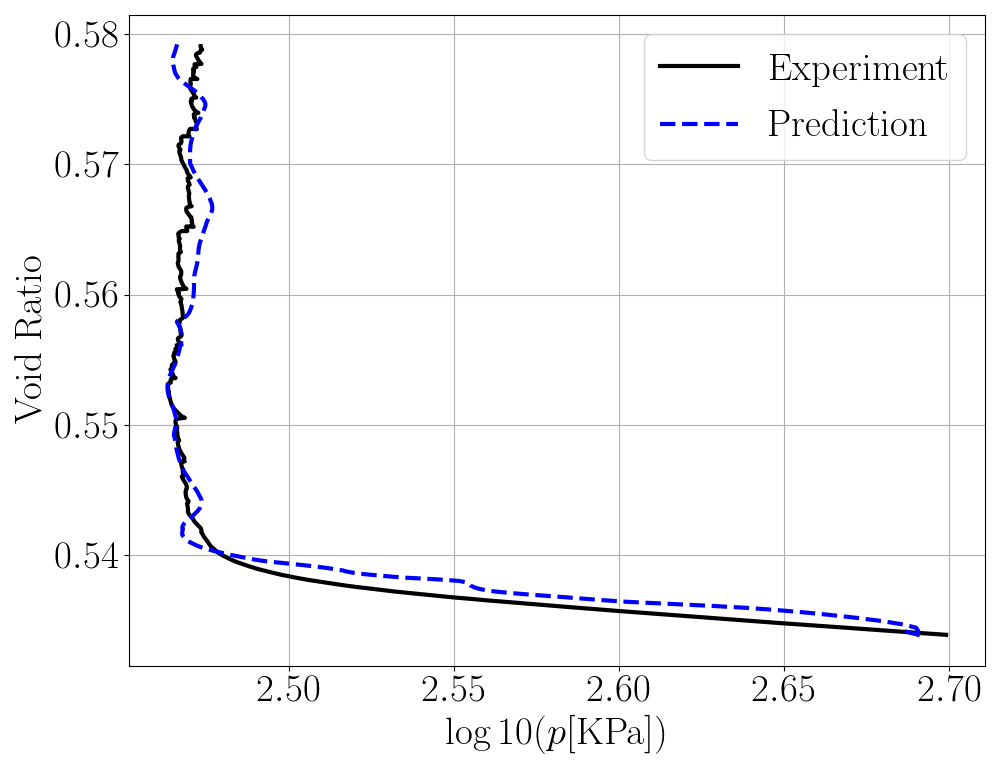}}
\hspace{0.01\textwidth}
 \subfigure[Blind Test  No. 50 (TXC)]
{\includegraphics[width=0.45\textwidth]{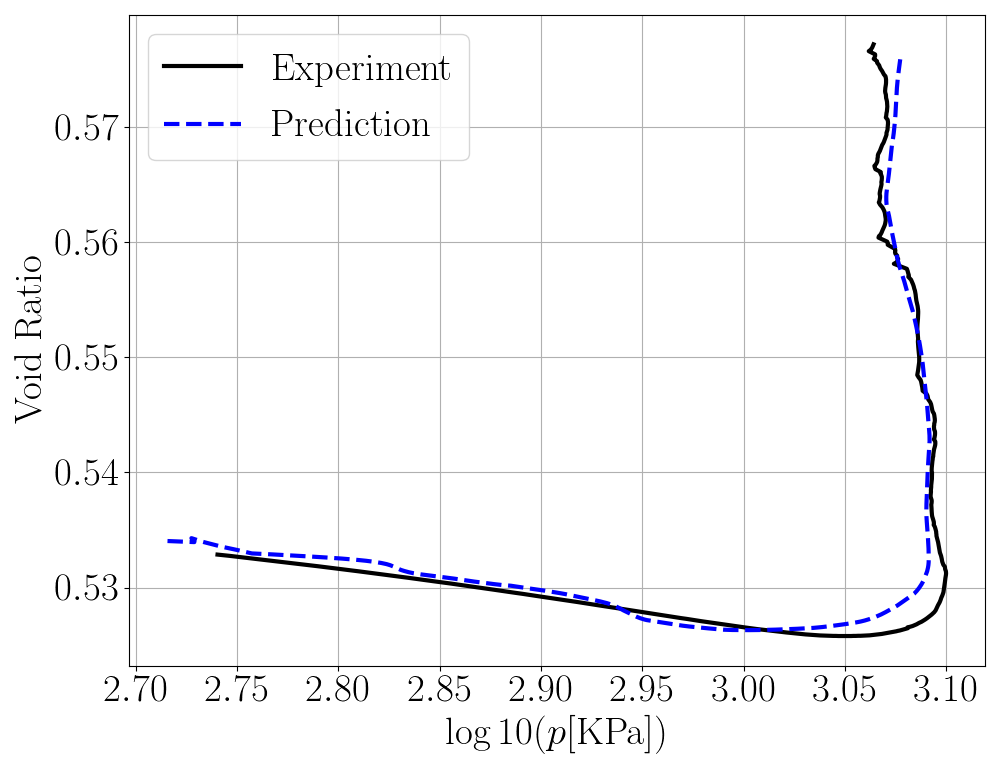}}
\hspace{0.01\textwidth}
 \subfigure[Blind Test  No. 56 (TXC)]
{\includegraphics[width=0.45\textwidth]{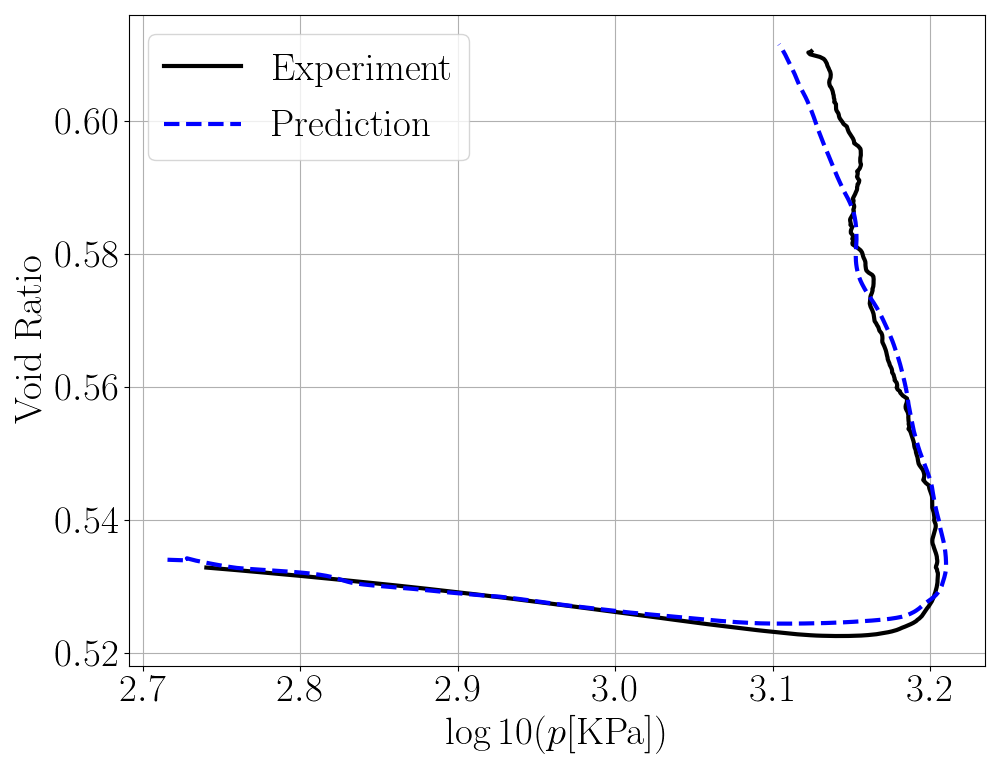}}
    \caption{State path (void ratio vs. logarithm of mean pressure). Compressive strain has positive sign convention.}
    \label{fig:statepath}
\end{figure}

The second important characteristics that warrants attention is the state path in the void ratio vs. logarithm of mean pressure. Here, we consider compressive pressure as positive and the results are shown in Fig. \ref{fig:statepath}. Again, the predictions indicate that the 
trained neural network is able to predict the elastic compression followed by the plastic dilatancy in the triaxial compression (TXC) cases and the elastoplastic expansion in the triaxial extension (TXE) cases. 

\begin{figure}[h!]
 \centering
 \subfigure[Calibration No. 23 (TXE)]
 {\includegraphics[width=0.45\textwidth]{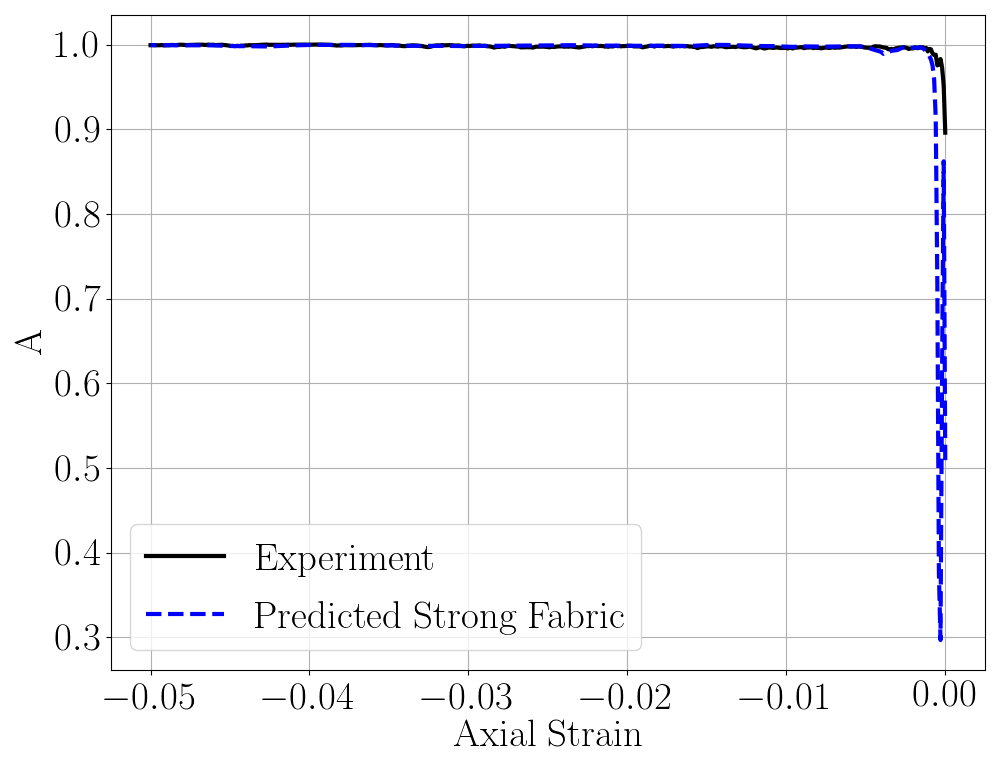}}
\hspace{0.01\textwidth}
 \subfigure[Calibration No. 29 (TXE)]
{\includegraphics[width=0.45\textwidth]{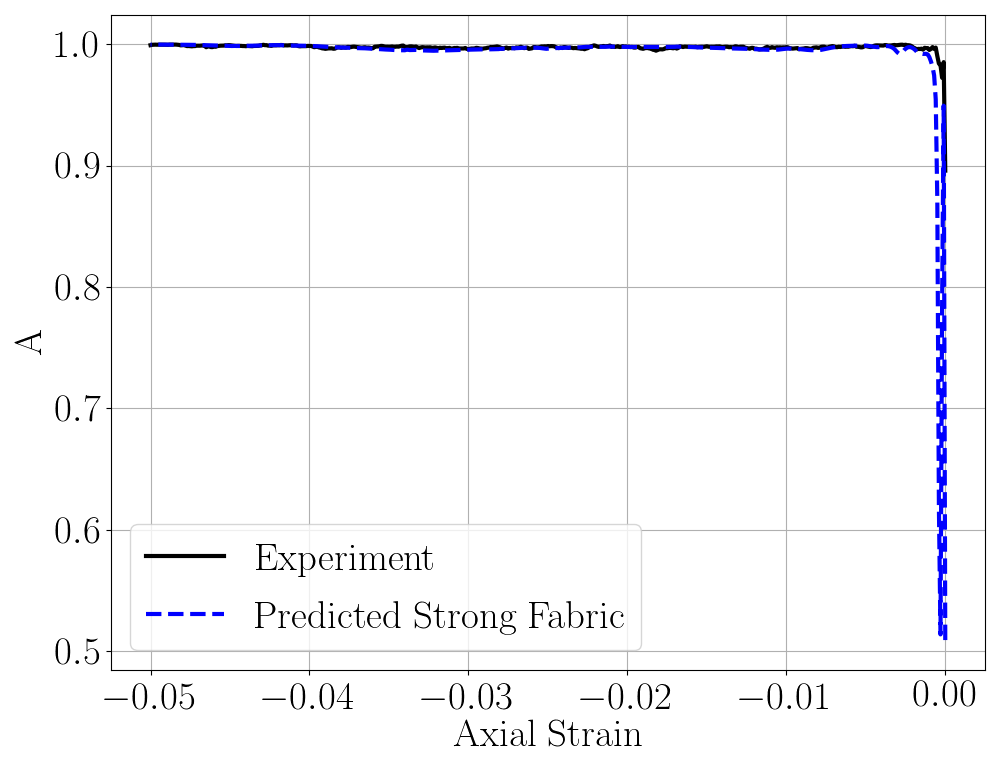}}
\hspace{0.01\textwidth}
 \subfigure[Blind Test  No. 50 (TXC)]
{\includegraphics[width=0.45\textwidth]{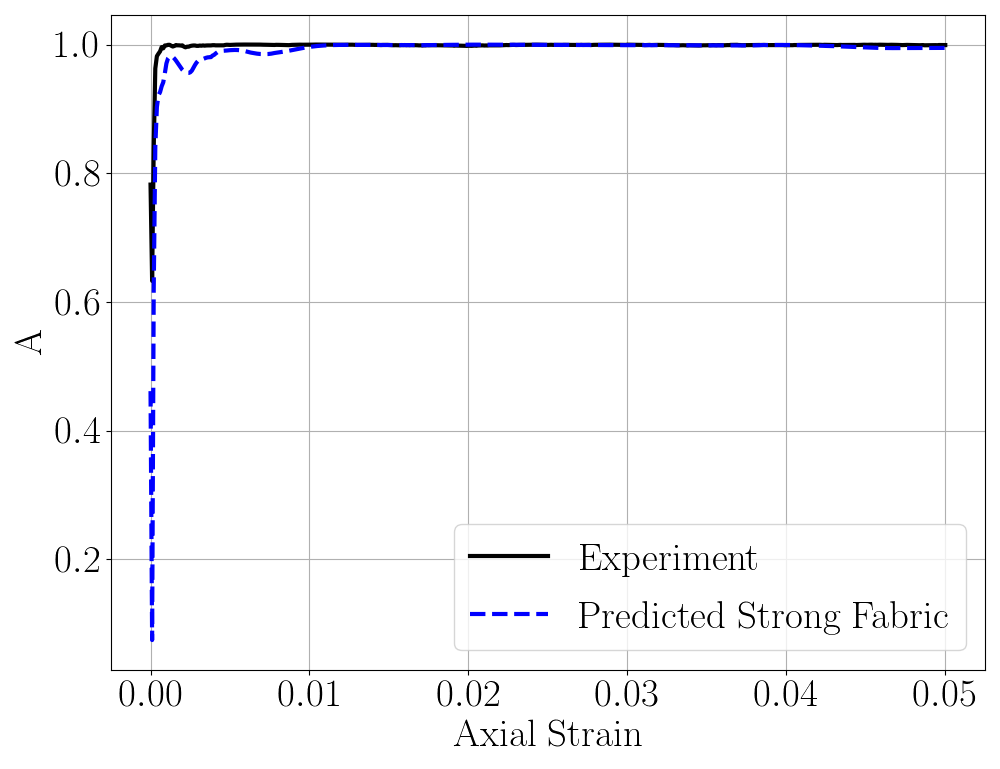}}
\hspace{0.01\textwidth}
 \subfigure[Blind Test  No. 56 (TXC)]
{\includegraphics[width=0.45\textwidth]{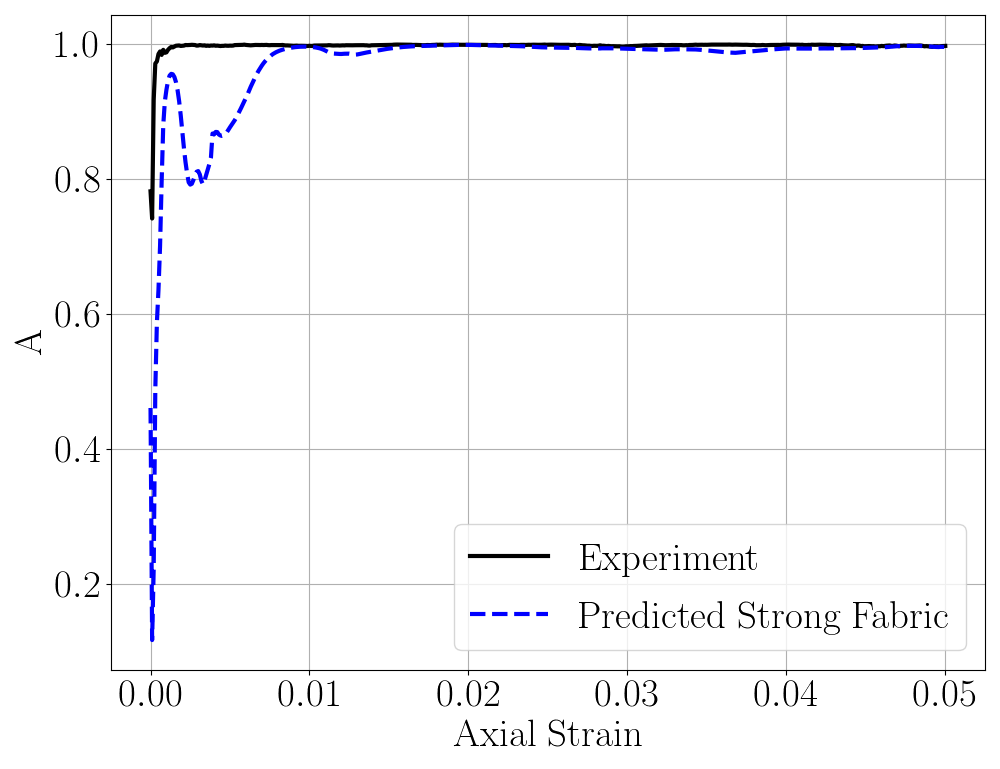}}
    \caption{Normalized fabric anisotropy variable vs. axial strain. Compressive strain has positive sign convention.}
    \label{fig:A}
\end{figure}

Next, we examine the strong fabric tensor and its relationships with  graph measures. Here, the fabric tensor $\tensor{F}$ is computed by the summation of the dyadic product of the branch vectors $\vec{n}$ divided by the
number of grain contacts $n_{c}$, i.e., 
\begin{equation}
\tensor{F} = \frac{1}{n_{c}} \sum_{i=1}^{n_{c}} \vec{n} \otimes \vec{n}.
\end{equation} 
The strong fabric tensor is obtained by considering only a subset of the 
contact of which the contact normal force is larger than a threshold value. In this work, this threshold value is set to be the averaged contact force. For brevity, we only show the normalized fabric anisotropy variable, which measures the alignment between the fabric tensor and 
the normalized deviatoric component of the stress $\tensor{n}^{\text{dev}}$, 
\begin{equation}
A = 
\frac{1}{\sqrt{\tensor{F}: \tensor{F}}} 
\tensor{F}  : \tensor{n}^{\text{dev}}, 
\end{equation}
in Fig. \ref{fig:A}.
Recall that the normalized fabric anisotropy variable $A=1$ is a necessary condition for a material to reach the critical state \citep{fu2011fabric, li2012anisotropic, zhao2013unique}, hence 
the predictions of $A$ may indicate how accurate the neural network in the causal graph predicts the critical state. 
Comparing the predictions of $A$ in the calibration cases and blind tests indicates that the neural network prediction tends to delay the predicted onset of the critical state. This may explain the over-fitting exhibited in Fig. \ref{fig:compareECDF}.

\begin{figure}[h!]
 \centering
 \subfigure[Calibration No. 23 (TXE)]
 {\includegraphics[width=0.45\textwidth]{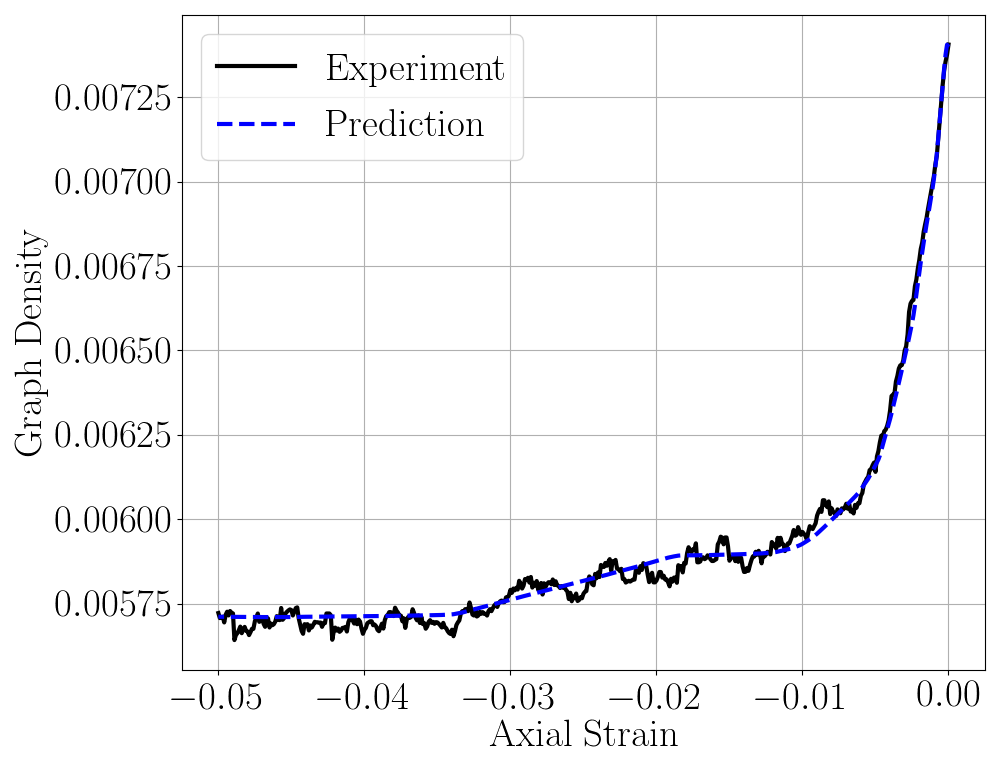}}
\hspace{0.01\textwidth}
 \subfigure[Calibration No. 29 (TXE)]
{\includegraphics[width=0.45\textwidth]{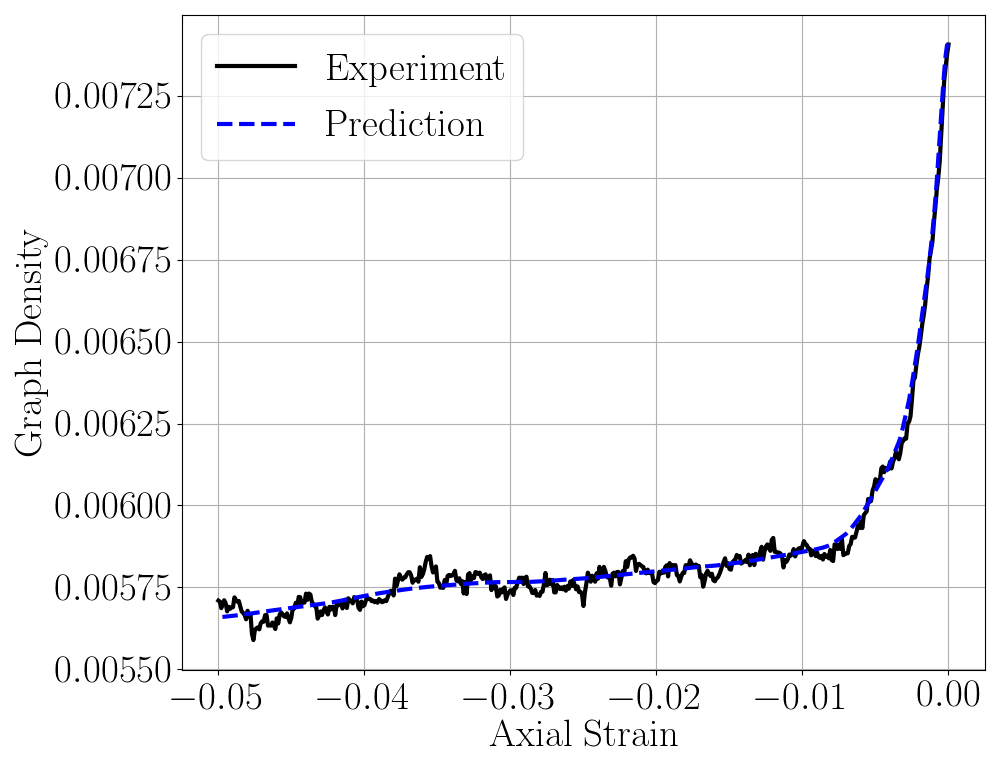}}
\hspace{0.01\textwidth}
 \subfigure[Blind Test  No. 50 (TXC)]
{\includegraphics[width=0.45\textwidth]{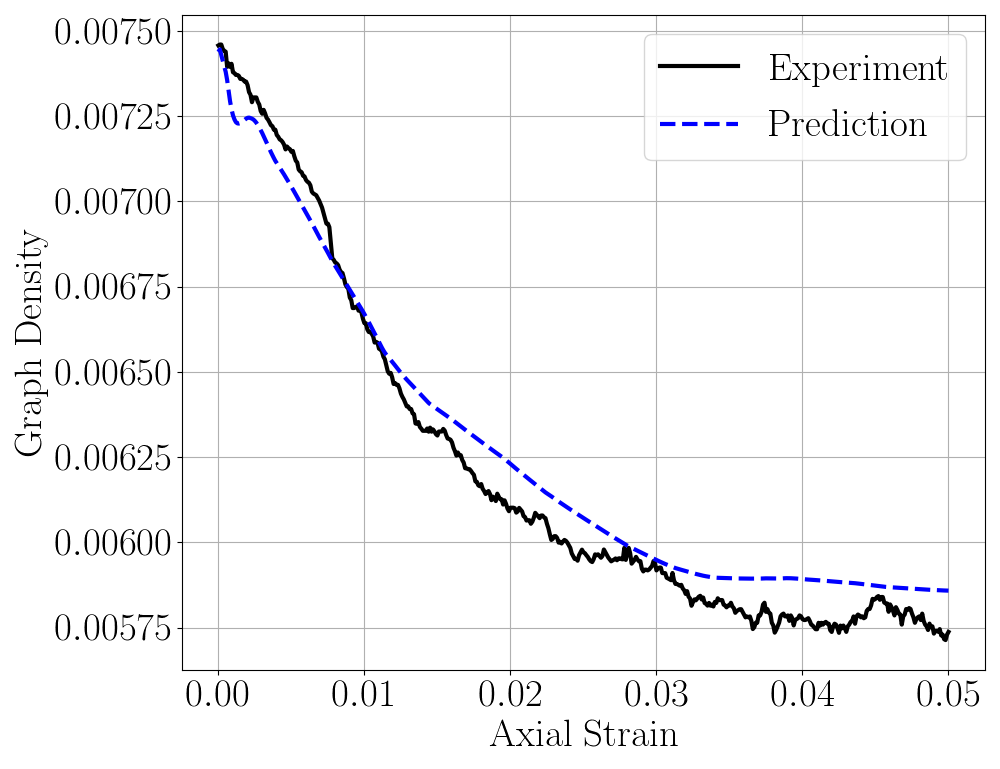}}
\hspace{0.01\textwidth}
 \subfigure[Blind Test  No. 56 (TXC)]
{\includegraphics[width=0.45\textwidth]{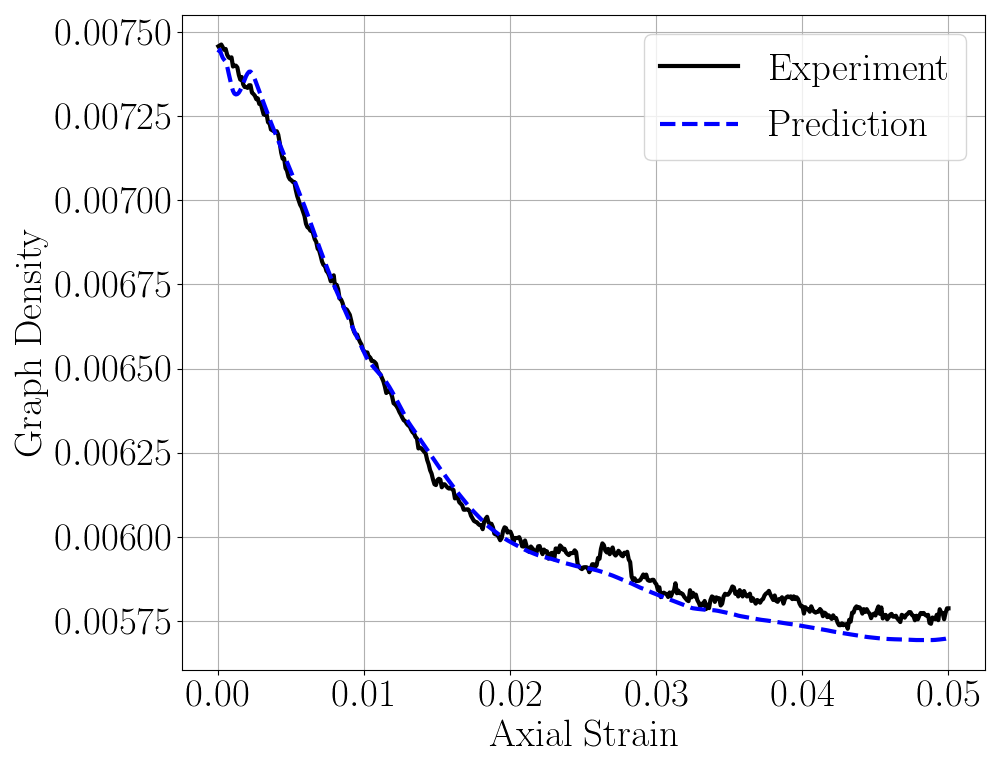}}
    \caption{Graph density vs. axial strain. Compressive strain has positive sign convention.}
    \label{fig:graphdensity}
\end{figure}

Finally, we examine the predictions of the graph measures most likely to be influential to the predictions of the fabric tensors. Fig. \ref{fig:graphdensity} shows that the graph density reduces during the shear phase in both triaxial compression and extension tests.  According to the causal graph,  the deformation is causing the graph density changing which in turn affects the fabric tensors. These causal relationships seem reasonable as the deformation during the shear phase is likely to cause plastic dilatancy and therefore reduces the 
number of contacts, which explains the drop in the graph density and the resultant changes in fabric tensors.

\begin{figure}[h!]
 \centering
 \subfigure[Calibration No. 23 (TXE)]
 {\includegraphics[width=0.45\textwidth]{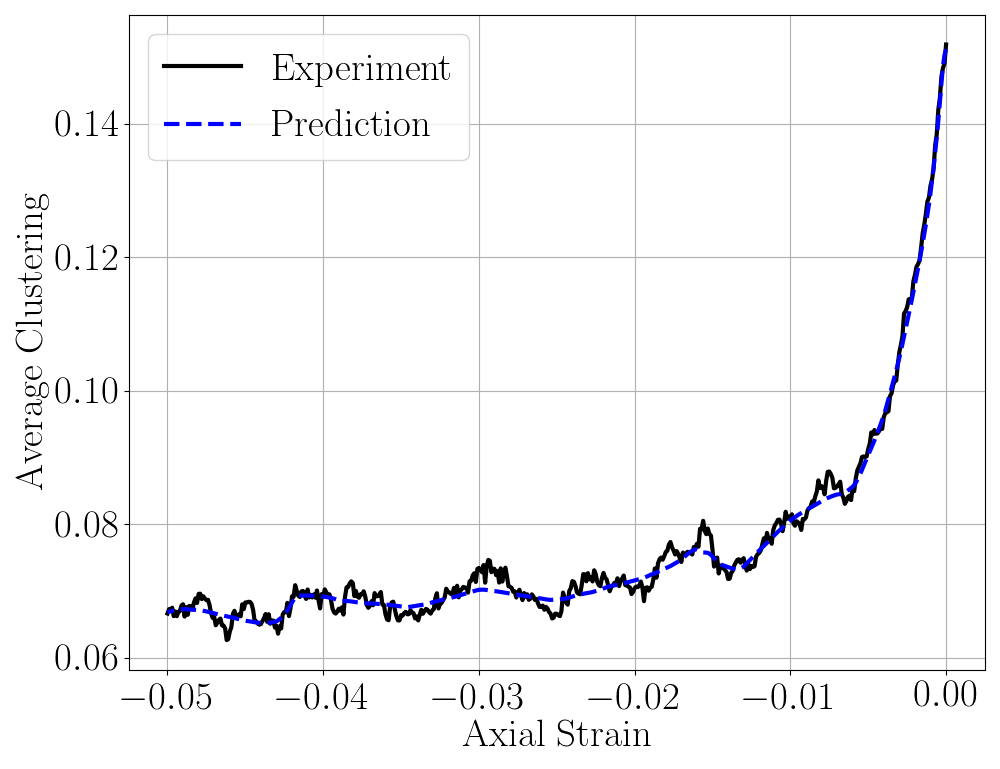}}
\hspace{0.01\textwidth}
 \subfigure[Calibration No. 29 (TXE)]
{\includegraphics[width=0.45\textwidth]{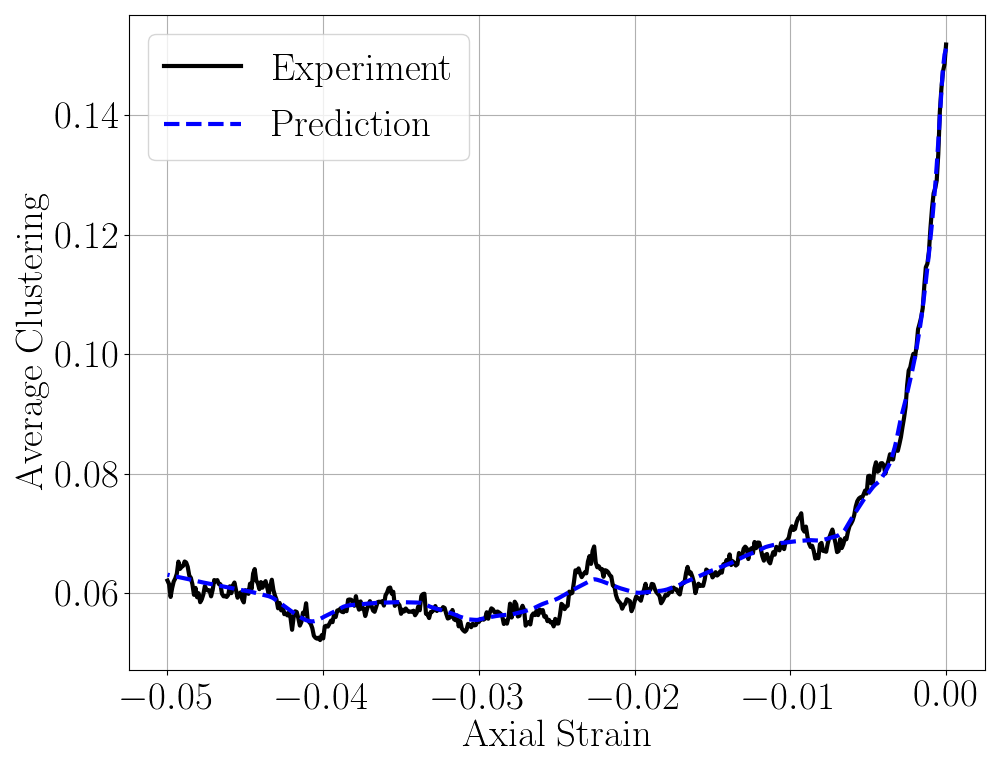}}
\hspace{0.01\textwidth}
 \subfigure[Blind Test  No. 50 (TXC)]
{\includegraphics[width=0.45\textwidth]{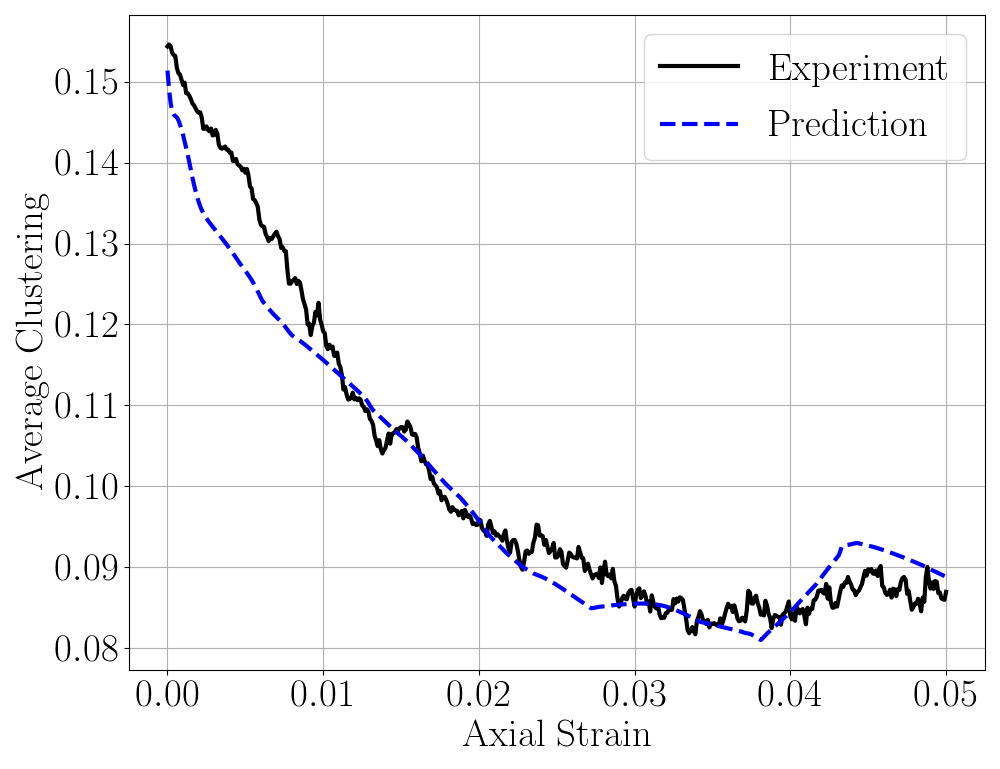}}
\hspace{0.01\textwidth}
 \subfigure[Blind Test  No. 56 (TXC)]
{\includegraphics[width=0.45\textwidth]{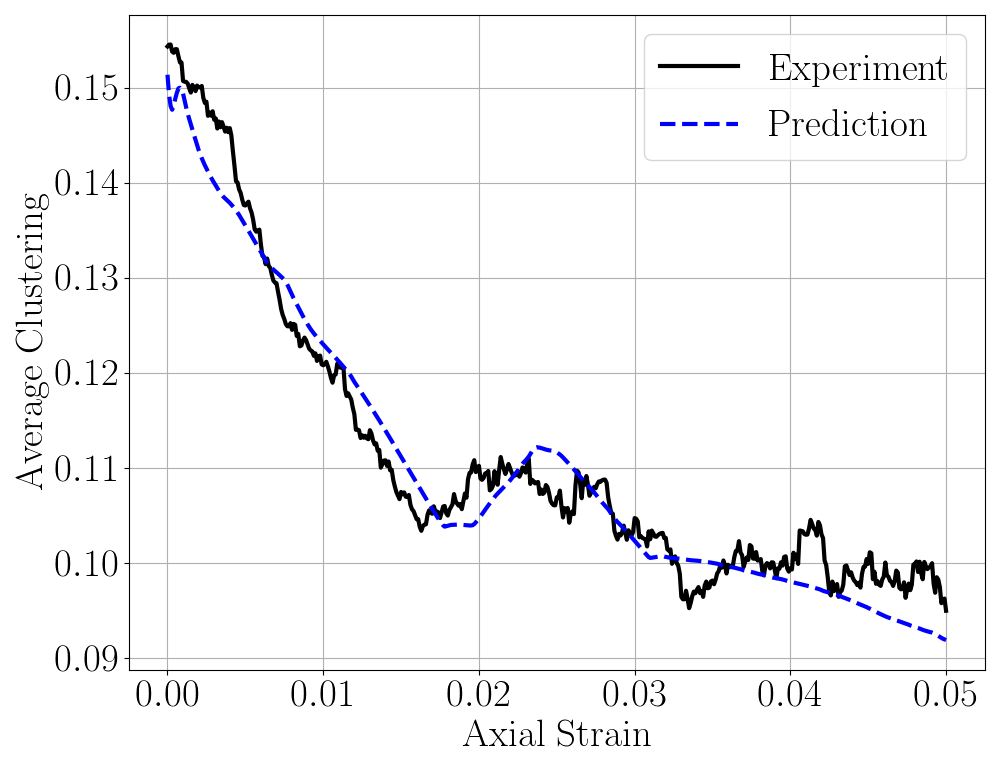}}
    \caption{Average Clustering vs. axial strain. Compressive strain has positive sign convention.}
    \label{fig:averageclustering}
\end{figure}

A similar reasoning can also be used to explain the drops in the average clustering shown in Fig. \ref{fig:averageclustering} where the shear deformation tends to reduce the tendency of the particles to cluster together and that in return leads to the evolution of the fabric tensor. 

These results indicate that, while the causal discovery may reveal potential causal relations not apparent to domain experts, the knowledge from causal relationships does not necessarily lead to more accurate predictions. Factors such as the choices of the supervised machine learning methods and 
the availability of data are also key factors that affect the usefulness 
of the new knowledge for predictions.

\subsubsection{Uncertainty propagation with dropout layer}

As the final  numerical experiment, we activate dropout layers to collect results of stochastic forward passes through the model. This gives us a Monte Carlo estimate of the predictions. 
 Note that the activation of the dropout layers will lead to a different set of neuron weights 
 even the data used for the training of the neural network are identical. 
 
\begin{figure}[h!]
 \centering
 \subfigure[Calibration No. 23 (TXE)]
 {\includegraphics[width=0.45\textwidth]{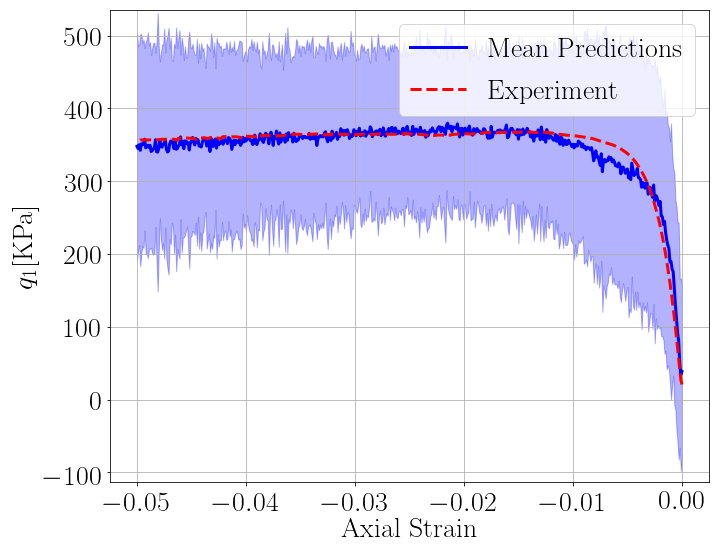}}
\hspace{0.01\textwidth}
 \subfigure[Calibration No. 29 (TXE)]
{\includegraphics[width=0.45\textwidth]{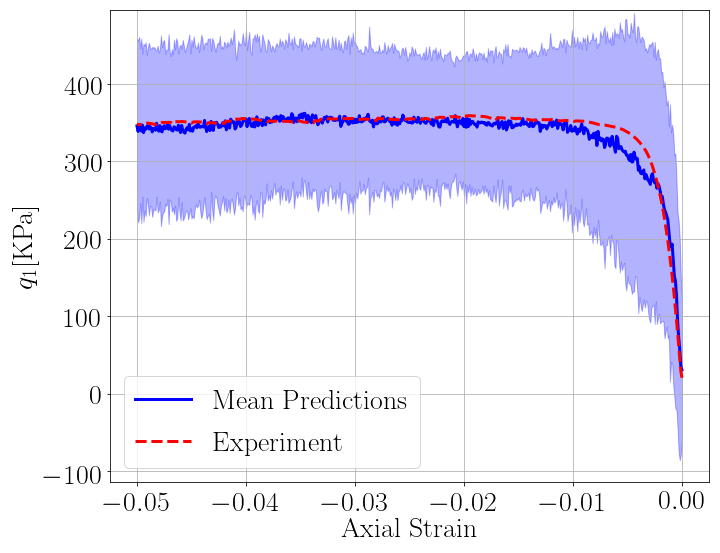}}
\hspace{0.01\textwidth}
 \subfigure[Blind Test  No. 50 (TXC)]
{\includegraphics[width=0.45\textwidth]{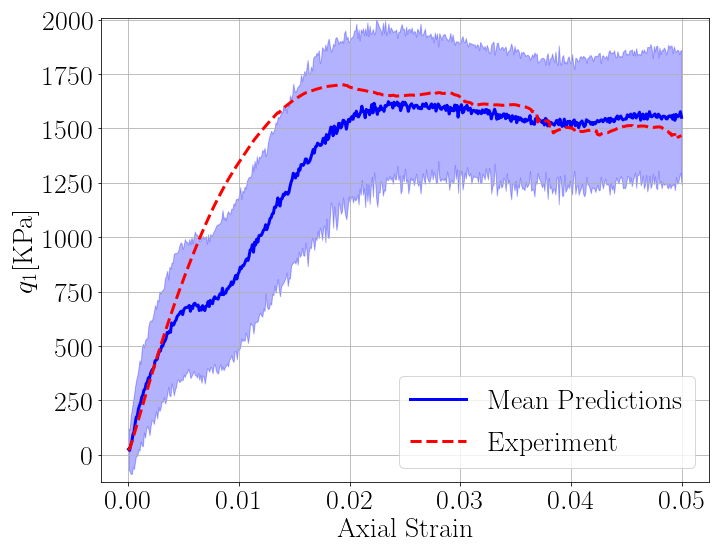}}
\hspace{0.01\textwidth}
 \subfigure[Blind Test  No. 56 (TXC)]
{\includegraphics[width=0.45\textwidth]{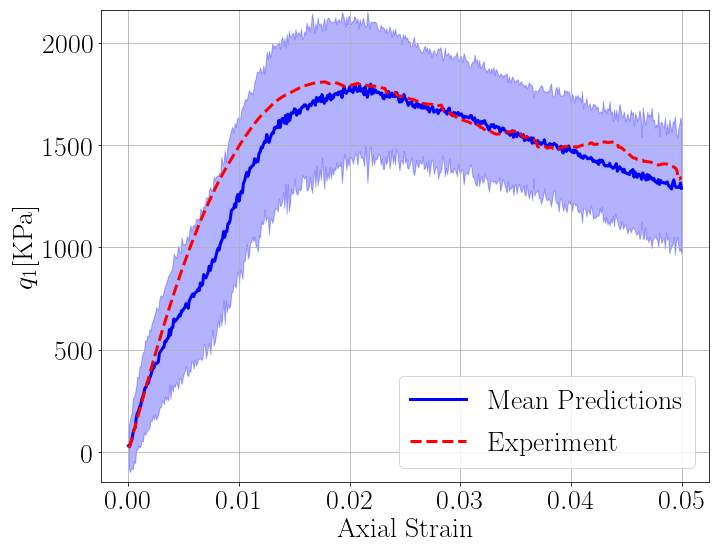}}
    \caption{Difference between the major and minor principal stress vs. axial strain. Results are obtained for active dropout layers. Shaded area includes predictions within $95\%$ confidence interval. Compressive strain has positive sign convention.}
    \label{fig:q1-UQ}
\end{figure}

\begin{figure}[h!]
 \centering
 \subfigure[Calibration No. 23 (TXE)]
 {\includegraphics[width=0.45\textwidth]{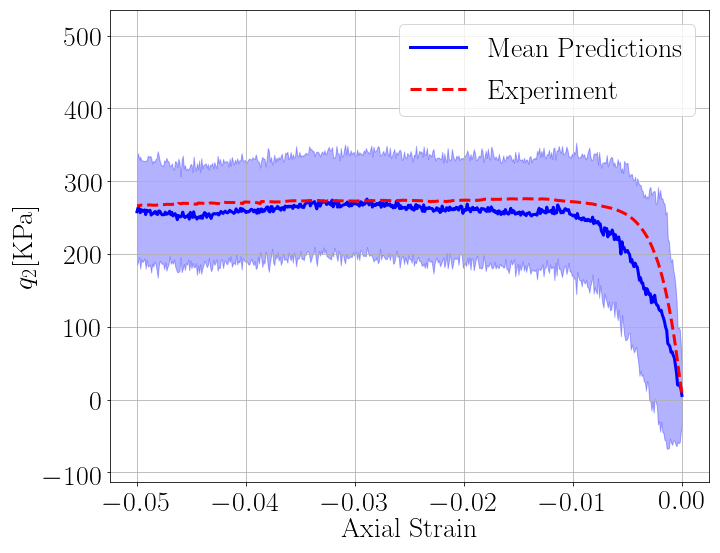}}
\hspace{0.01\textwidth}
 \subfigure[Calibration No. 29 (TXE)]
{\includegraphics[width=0.45\textwidth]{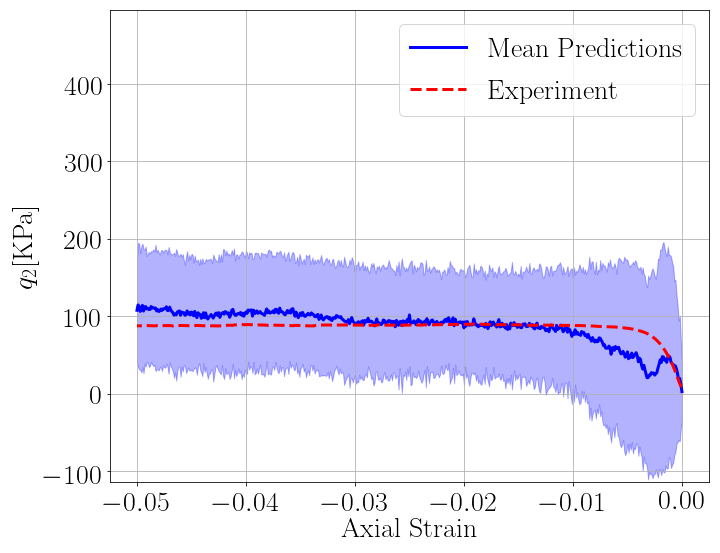}}
\hspace{0.01\textwidth}
 \subfigure[Blind Test  No. 50 (TXC)]
{\includegraphics[width=0.45\textwidth]{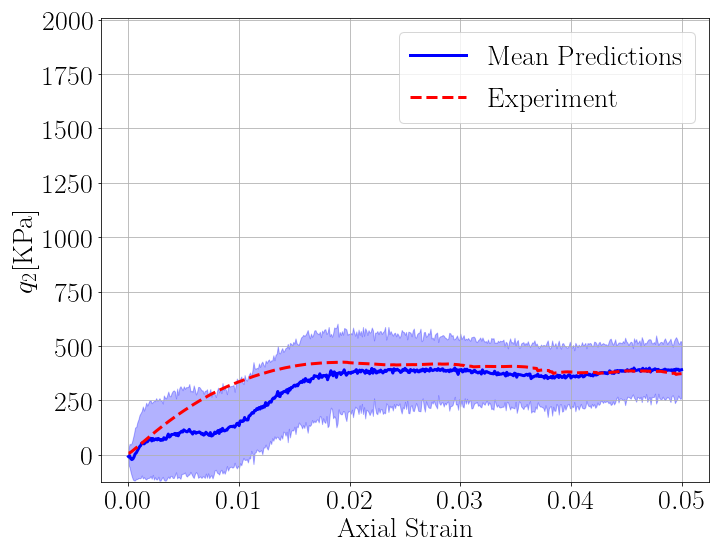}}
\hspace{0.01\textwidth}
 \subfigure[Blind Test  No. 56 (TXC)]
{\includegraphics[width=0.45\textwidth]{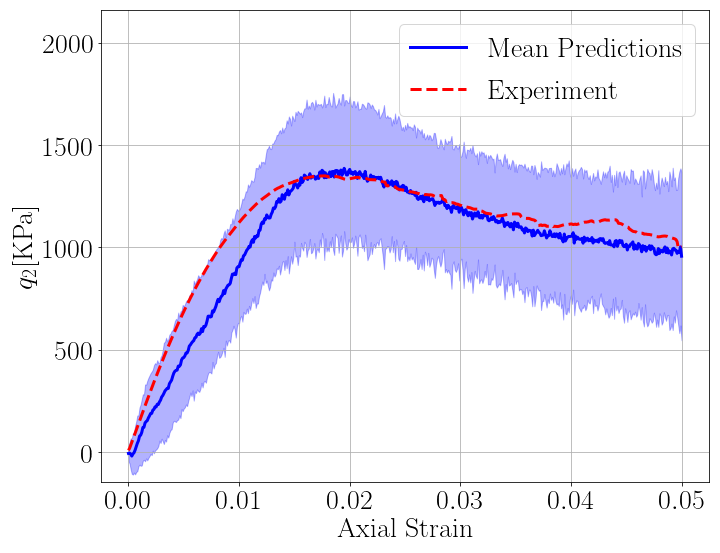}}
    \caption{Difference between the major and immediate principal stress vs. axial strain. Results are obtained for active dropout layers. Shaded area includes predictions within $95\%$ confidence interval. Compressive strain has positive sign convention.}
    \label{fig:q2-UQ}
\end{figure}

\begin{figure}[h!]
 \centering
 \subfigure[Calibration No. 23 (TXE)]
 {\includegraphics[width=0.45\textwidth]{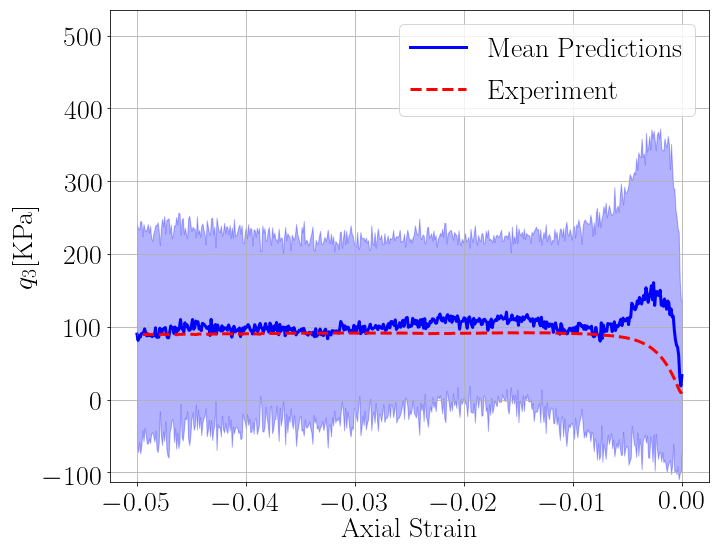}}
\hspace{0.01\textwidth}
 \subfigure[Calibration No. 29 (TXE)]
{\includegraphics[width=0.45\textwidth]{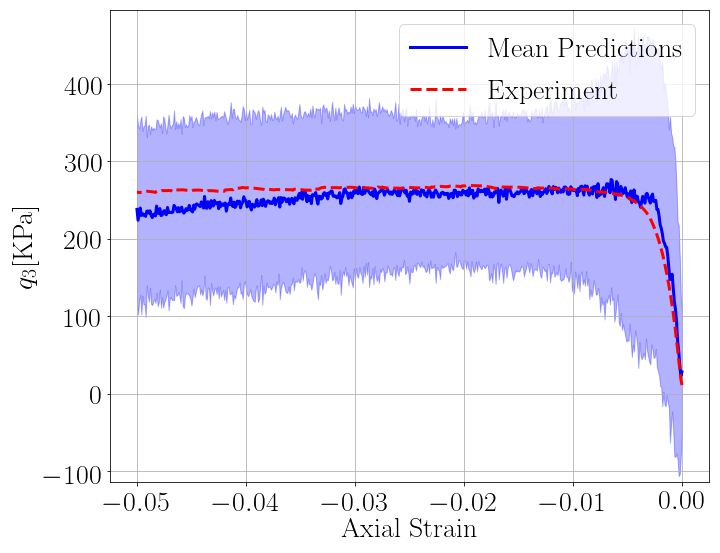}}
\hspace{0.01\textwidth}
 \subfigure[Blind Test  No. 50 (TXC)]
{\includegraphics[width=0.45\textwidth]{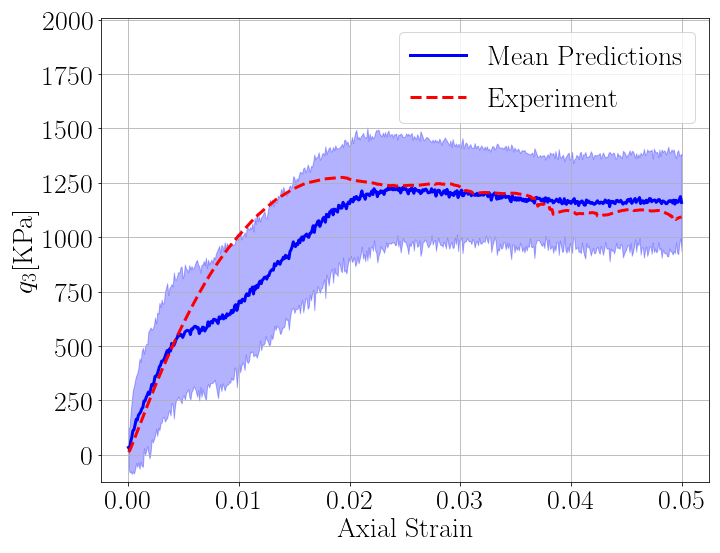}}
\hspace{0.01\textwidth}
 \subfigure[Blind Test  No. 56 (TXC)]
{\includegraphics[width=0.45\textwidth]{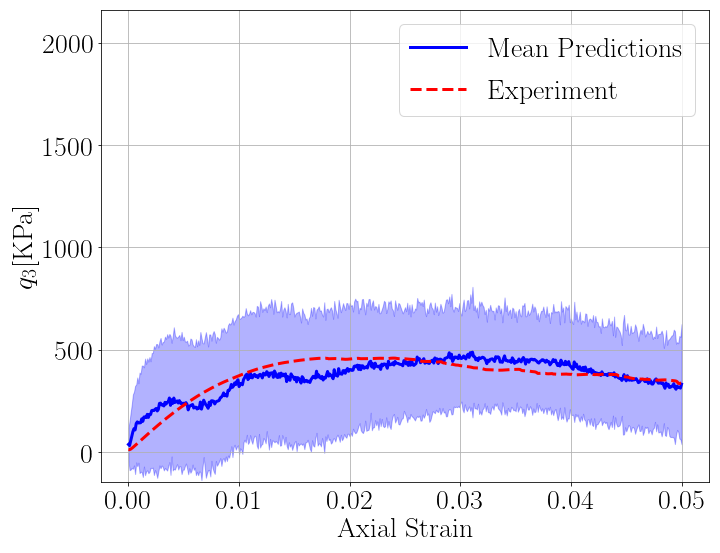}}
    \caption{Difference between the immediate and minor principal stress vs. axial strain. Results are obtained for active dropout layers. Shaded area includes predictions within $95\%$ confidence interval. Compressive strain has positive sign convention.}
    \label{fig:q3-UQ}
\end{figure}

Figs. \ref{fig:q1-UQ}, \ref{fig:q2-UQ}, and \ref{fig:q3-UQ} show the confidence interval for 200
Monte Carlo predictions of principal stress differences, $q_{1}$, $q_{2}$ and $q_{3}$ vs. axial strain for 4 selected triaxial extension and compression loading paths. 

In most of the cases shown in Figs. \ref{fig:q1-UQ}, \ref{fig:q2-UQ}, and \ref{fig:q3-UQ}, the mean paths of the stochastic predictions generated by the dropout layer is able to match qualitatively with the experimental benchmarks. 
Furthermore, in most cases, the principal stress differences observed from experiments are within the $95 \%$ confidence interval. It should nevertheless be noted that 
the blind test is less accurate than the calibration cases, indicating that the neural networks may have been over-fitted. 

\begin{figure}[h!]
 \centering
 \subfigure[Calibration No. 23 (TXE)]
 {\includegraphics[width=0.45\textwidth]{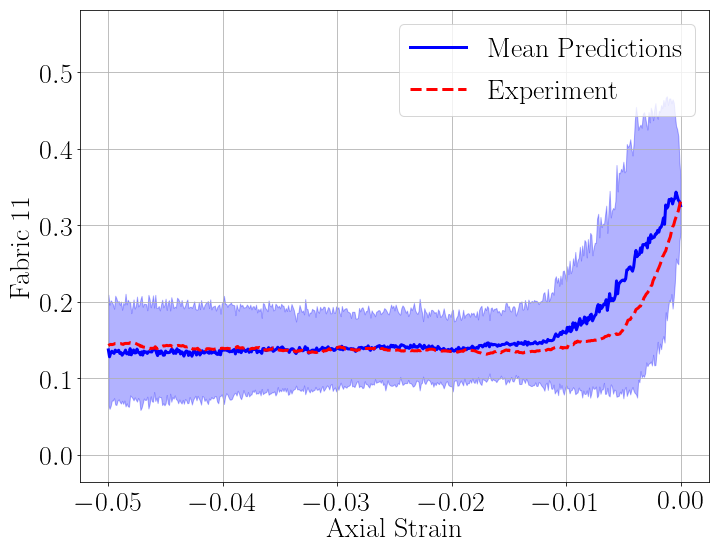}}
\hspace{0.01\textwidth}
 \subfigure[Calibration No. 29 (TXE)]
{\includegraphics[width=0.45\textwidth]{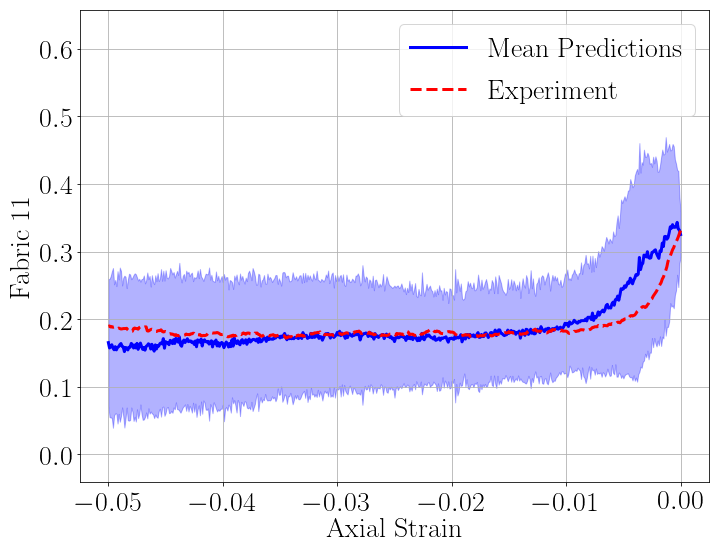}}
\hspace{0.01\textwidth}
 \subfigure[Blind Test  No. 50 (TXC)]
{\includegraphics[width=0.45\textwidth]{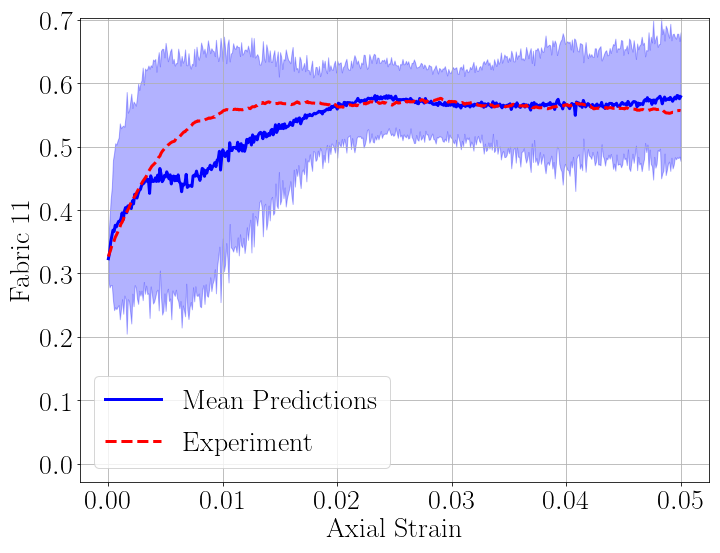}}
\hspace{0.01\textwidth}
 \subfigure[Blind Test  No. 56 (TXC)]
{\includegraphics[width=0.45\textwidth]{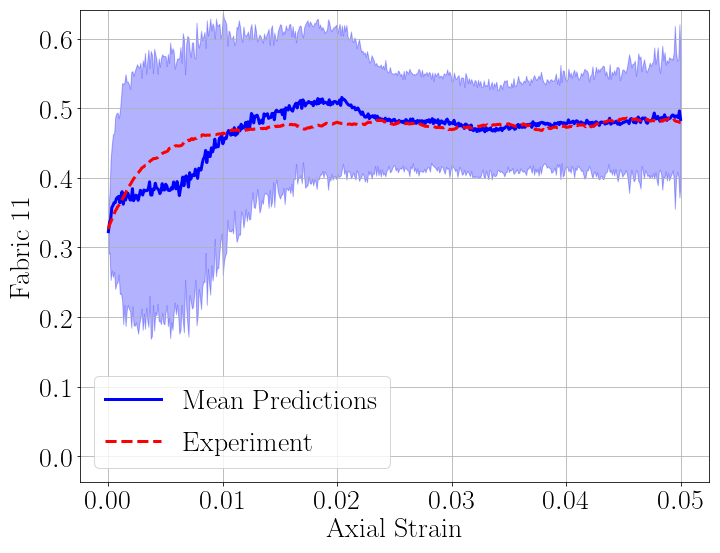}}
    \caption{Component 11 of fabric tensor vs. axial strain. Results are obtained for active dropout layers. Shaded area includes predictions within $95\%$ confidence interval. Compressive strain has positive sign convention.}
    \label{fig:fabric11-UQ}
\end{figure}

\begin{figure}[h!]
 \centering
 \subfigure[Calibration No. 23 (TXE)]
 {\includegraphics[width=0.45\textwidth]{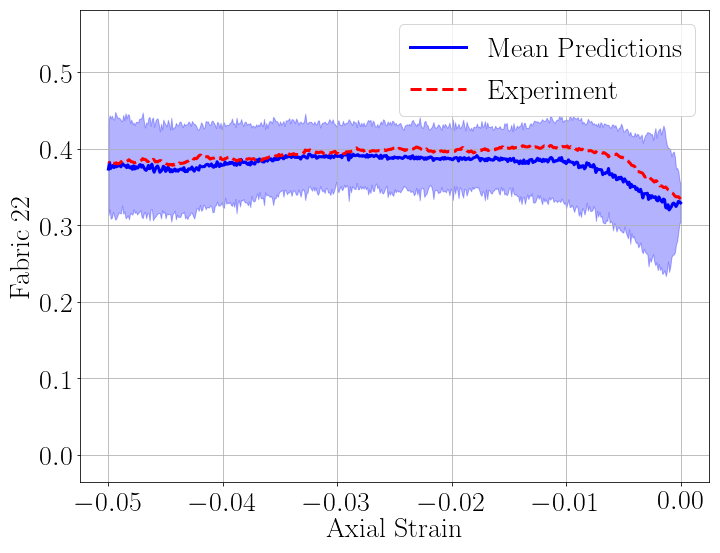}}
\hspace{0.01\textwidth}
 \subfigure[Calibration No. 29 (TXE)]
{\includegraphics[width=0.45\textwidth]{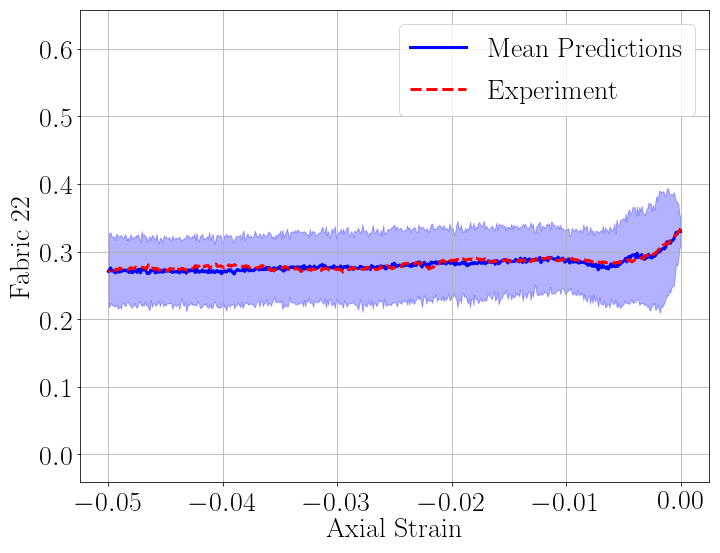}}
\hspace{0.01\textwidth}
 \subfigure[Blind Test  No. 50 (TXC)]
{\includegraphics[width=0.45\textwidth]{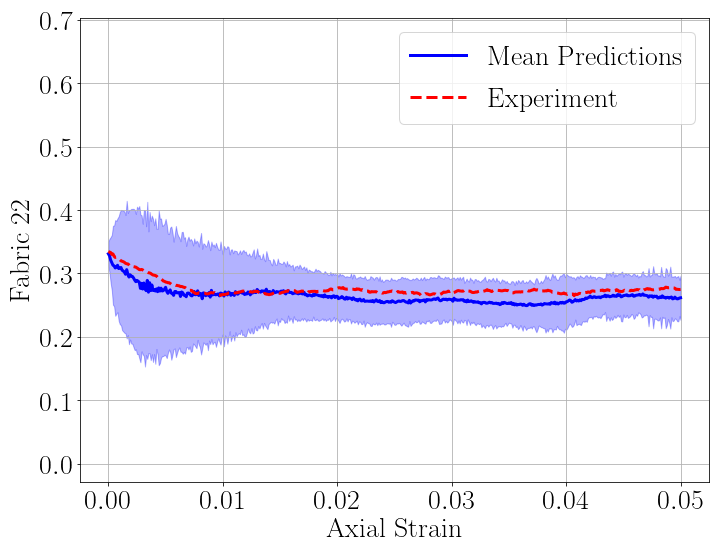}}
\hspace{0.01\textwidth}
 \subfigure[Blind Test  No. 56 (TXC)]
{\includegraphics[width=0.45\textwidth]{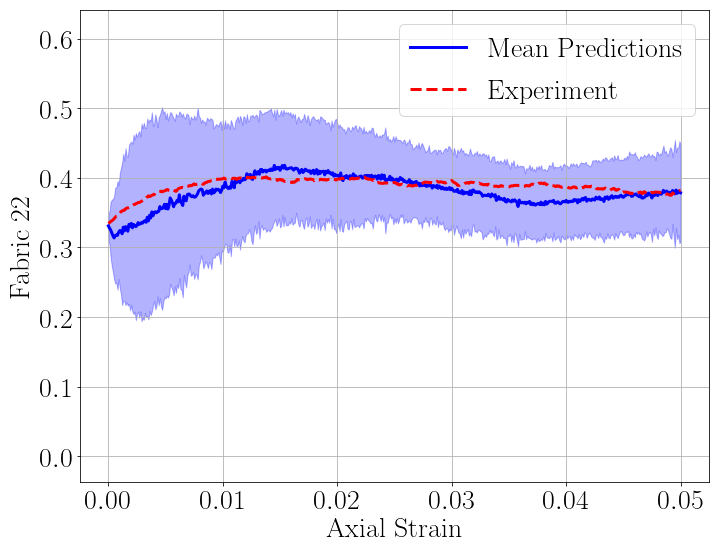}}
    \caption{Component 22 of fabric tensor vs. axial strain. Results are obtained for active dropout layers. Shaded area includes predictions within $95\%$ confidence interval. Compressive strain has positive sign convention.}
    \label{fig:fabric22-UQ}
\end{figure}

\begin{figure}[h!]
 \centering
 \subfigure[Calibration No. 23 (TXE)]
 {\includegraphics[width=0.45\textwidth]{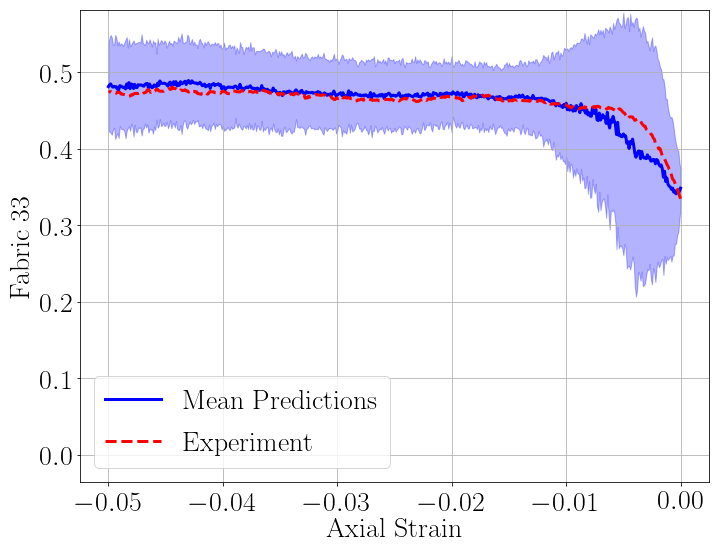}}
\hspace{0.01\textwidth}
 \subfigure[Calibration No. 29 (TXE)]
{\includegraphics[width=0.45\textwidth]{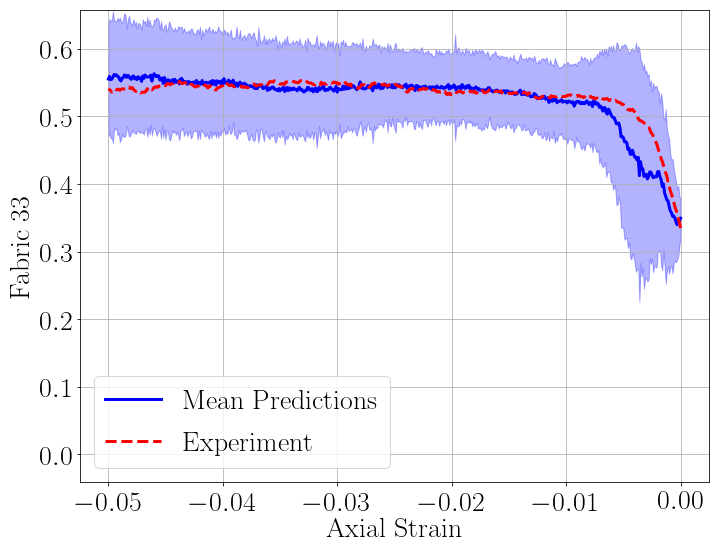}}
\hspace{0.01\textwidth}
 \subfigure[Blind Test  No. 50 (TXC)]
{\includegraphics[width=0.45\textwidth]{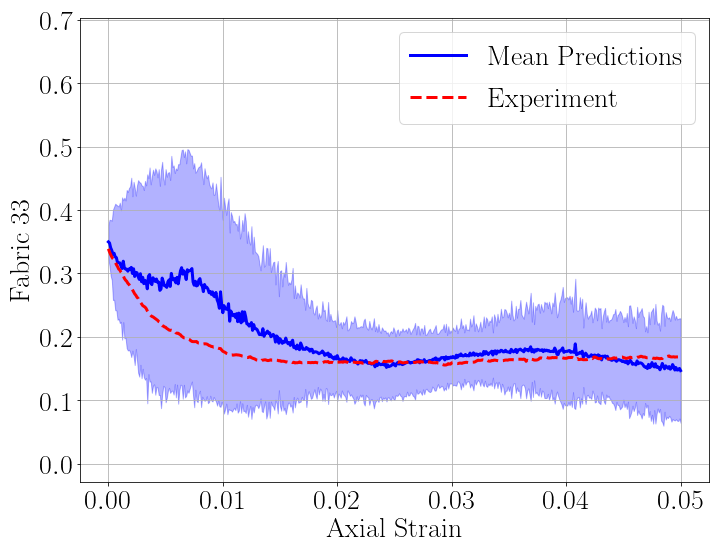}}
\hspace{0.01\textwidth}
 \subfigure[Blind Test  No. 56 (TXC)]
{\includegraphics[width=0.45\textwidth]{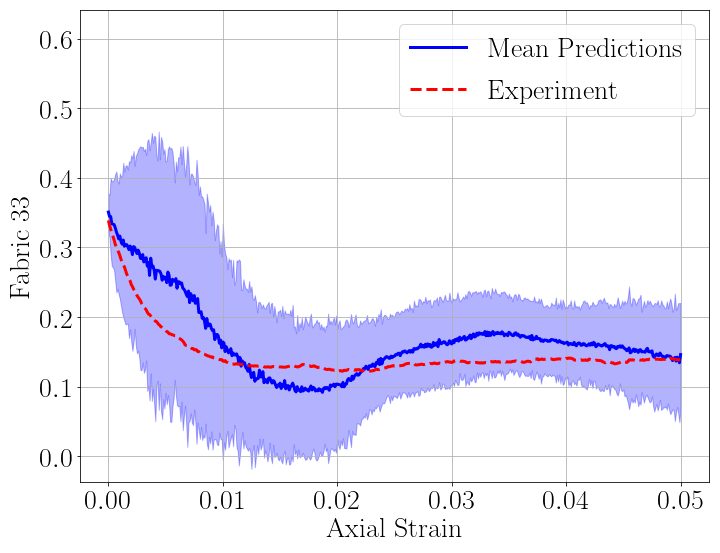}}
    \caption{Component 33 of fabric tensor vs. axial strain. Results are obtained for active dropout layers. Shaded area includes predictions within $95\%$ confidence interval. Compressive strain has positive sign convention.}
    \label{fig:fabric33-UQ}
\end{figure}

To examine how uncertainty is propagated in the causal graph, we plot the diagonal components of the fabric tensor and the results are shown in 
Figs. \ref{fig:fabric11-UQ},  \ref{fig:fabric22-UQ},  and \ref{fig:fabric33-UQ}. 
For brevity, the off-diagonal components of the fabric tensor, 
which are much smaller than the diagonal components, are not provided here. 
Comparing the $95\%$ confidence interval of the fabric tensor
 and that of the principal stress difference, one can easily see that the predictions
of stress tend to be more accurate when the fabric tensor can be more precisely determined with a narrower confidence interval. 

\section{Conclusions}
\label{sec:conclusion}
In this paper, we introduce, for the first time, a data-driven framework
 that combines 1) the 
causal discovery algorithm that detects unknown causal relations, 
2) the Bayesian approximation for uncertainty quantification enabled by the dropout technique, and 
3) the recurrent neural network technique to analyze, interpret, 
and forecast the path-dependent responses of granular materials. 
Numerical experiments conducted on idealized granular system 
have indicated that the data-driven framework 
is able to investigate and discover new hidden causal relationships and
propagate uncertainty generated from a sequence of structured 
neural network predictions
within a casual graph. This approach has potentials to 
help modelers and experimentalists to spot hidden mechanisms
not apparent to human eyes as well as deduce complex 
casual relationships in a high-dimensional parametric space 
where intuition and domain knowledge are not sufficient 
due to the dimensionality of the data. 
Further work may include improvement and comparisons of different 
causal inferences, extension to recover causal relations when both instantaneous and lagged causal relations exist, as well as the applications to more complex granular systems 
where particles are of different shapes and properties.

\section{Availability of data, material, and code for reproducing results}
The causal discovery algorithm can be found at \citet{xu2020}. The recurrent neural network is built via Tensorflow and the code to complete the training and the generation of the forecast engine can be found at \citet{bahmani2020}. The discrete element simulations data can be found in the Mendeley data repositories \citep{wang2019discrete,hypopl_data}.

\section{Acknowledgments}
The members of the Columbia research group involved in this research are supported  
by National Science Foundation under grant contracts CMMI-1846875 and OAC-1940203,  
the Earth Materials and Processes
program from the US Army Research Office under grant contract 
W911NF-18-2-0306, and the Dynamic Materials and Interactions Program from 
the Air Force Office of Scientific 
Research under grant contracts FA9550-17-1-0169. 
The members of the Johns Hopkins University are supported by National Science Foundation under grant contract 1940107.  
These supports are gratefully acknowledged. 
 The authors would also like to thank  Dr. Kun Wang from Los Alamos National Laboratory for providing the data for the traction-separation law. 

The views and conclusions contained in this document are those of the authors, 
and should not be interpreted as representing the official policies, either expressed or implied, 
of the sponsors, including the Army Research Laboratory or the U.S. Government. 
The U.S. Government is authorized to reproduce and distribute reprints for 
Government purposes notwithstanding any copyright notation herein.

\subsection{Author statement}
Xiao Sun and Bahador Bahmani contribute equally as first authors.
All authors have contributed to the planning/writing/reviewing/editing of this manuscript.

\subsection{Declaration of Competing Interest}
The authors confirm that there are no relevant financial or 
non-financial competing interests to report.

\begin{appendix}
\normalsize{

\section{Appendix: Proof of Theorem 1}
\label{sec:proof}
\noindent {\bf Theorem 1 }	
	Given Assumptions 1-3, for every $V_i, V_j\in \mV_{-\vec{U}}$, $V_i$ and $V_j$ are not adjacent in $G$ if and only if they are independent conditional on some subset of $\{V_k\mid V_k\in \mV_{-\vec{U}}, k\neq i, k\neq j \}\cup \{\vec{U}\}$. 
	
\begin{proof}
 From equation \eqref{eq:eq1}, any variable $V_i$ in $\mV_{-\vec{U}}$ can be written as a function of $\{\theta_i(\vec{U})\}_{i=1}^{m-1} $ and $\{\epsilon_i\}_{i=1}^{m-1}$, where $m-1$ is the number of vertices included in  $\mV_{-\vec{U}}$ since $\mV$ includes $m$ vertices. 
 Therefore, the distribution of $\mV_{-\vec{U}}$ at each value of $\vec{U}$ is determined by the distribution of $\epsilon_1, \dots, \epsilon_{m-1}$ and the values of $\{\theta_i(\vec{U})\}_{i=1}^{m-1}$. For any $V_i, V_j \in \mV_{-\vec{U}}$ and $S\subseteq \{V_k\mid V_k \in \mV_{-\vec{U}}, k\neq i, k\neq j \}$, $p(V_i, V_j \mid S  \cup \{ \vec{U}\})$ is determined by $\prod_{i=1}^{m-1}p(\epsilon_i)$ and $\{\theta_i(\vec{U})\}_{i=1}^{m-1} $. Since $\prod_{i=1}^{m-1}p(\epsilon_i)$  does not change with $\vec{U}$, 
 we have
 \begin{eqnarray}
 p(V_i, V_j \mid S \cup \{\theta_i(\vec{U})\}_{i=1}^{m-1} \cup \{ \vec{U}\})=p(V_i, V_j \mid S \cup \{\theta_i(\vec{U})\}_{i=1}^{m-1}). 
 \end{eqnarray}
 Denote $ \indep$ to indicate independence, it follows that
 \begin{eqnarray}
 \label{eq:p1}
 \vec{U}  \indep (V_i, V_j) \mid S \cup \{\theta_i(\vec{U})\}_{i=1}^{m-1}. 
  \end{eqnarray}
 Applying the weak union property of conditional independence, we have $\vec{U}  \indep V_i \mid \{V_j\} \cup S \cup \{\theta_i(\vec{U})\}_{i=1}^{m-1}. $
 
 Suppose that $V_i$ and $V_j$ are not adjacent in $G$, then they are not adjacent in $G^{aug}$. There exists a set $S\subseteq \{V_k\mid V_k \in \mV_{-\vec{U}}, k\neq i, k\neq j \}$ such that $S\cup \{\theta_i(\vec{U})\}_{i=1}^{m-1}$ d-separates $V_i$ and $V_j$. Because of Assumption 1, we have  
 \begin{eqnarray}
 \label{eq:p2}
V_i \indep V_j \mid S\cup \{\theta_i(\vec{U})\}_{i=1}^{m-1}.
 \end{eqnarray}
Since all $\theta_i(\vec{U})$ are deterministic functions of $\vec{U}$, we have $p(V_i, V_j\mid S \cup \vec{U}) = p(V_i, V_j\mid S \cup  \{\theta_i(\vec{U})\}_{i=1}^{m-1} \cup \vec{U})$. 

Equations \eqref{eq:p1} and \eqref{eq:p2} imply that $V_i \indep (\vec{U}, V_j) \mid S \cup  \{\theta_i(\vec{U})\}_{i=1}^{m-1}$. By the weak union property of conditional independence, we have $V_i \indep  V_j \mid S \cup  \{\theta_i(\vec{U})\}_{i=1}^{m-1}\cup \{\vec{U}\}$.  Since all $\theta_i(\vec{U})$ are deterministic functions of $\vec{U}$, it follows that $V_i \indep  V_j \mid S \cup \{\vec{U}\}$. 

Now we prove that if $V_i$ and $V_j$ are conditionally independent given a subset $S$ of  $\{V_k\mid V_k\in \mV_{-\vec{U}}, k\neq i, k\neq j \}\cup \{\vec{U}\}$, $V_i$ and $V_j$ are not adjacent in $G$. Because of Assumption 2 (faithfullness),  $V_i$ and $V_j$ are not adjacent in $G^{aug}$. Therefore, they are not adjacent in $G$. 
 
\end{proof}

\section{Appendix: Graph metric definitions}
\label{sec:graph_metrics}

In this section, we provide brief review of the the terminology of the graph measures obtained 
from the grain connectivity graph generated in each time step of a discrete element simulation. 
These graph measures are  
used to create the  knowledge graph for the 
machine learning constitutive law in Section \ref{sec:bulkplasticity}. 

The following graph metrics were calculated using the open-source software networkX \citep{hagberg2008exploring} for exploration and analysis of graph networks. 

\begin{definition}
The \textbf{degree assortativity coefficient} measures the similarity of the connections in a graph with respect to the node degree. \label{def:degrAssort} 
\end{definition}

\begin{definition}
The \textbf{graph transitivity} is the fraction of all possible triangles present in the graph over the number of triads. Possible triangles are identified by the number of triads -- two edges with a shared vertex. \label{def:transty}
 \end{definition}
 
 \begin{definition}
The \textbf{density} for undirected graphs is defined as:

\begin{equation}
d=\frac{2 m}{n(n-1)},
\end{equation} 

where $n$ in the number of nodes and $m$ is the number of edges of the graph.
\label{def:graphDensity}
 \end{definition}
 
\begin{definition}
The \textbf{average clustering coefficient} of the graph is defined as:
\begin{equation}
C=\frac{1}{n} \sum_{v \in G} c_{n},
\end{equation} 

where $n$ in the number of nodes and is $c_n$ is the clustering coefficient of node $n$ defined as:

\begin{equation}
c_{n}=\frac{2 T(n)}{\operatorname{deg}(n)(\operatorname{deg}(n)-1)},
\end{equation} 

where $T(n)$ is the number of triangles passing through node $n$ and $\operatorname{deg}(n)$ is the degree of node $n$.
\label{def:average_clustering}
 \end{definition}

 \begin{definition}
 A \textbf{clique} is a subset of nodes of an undirected graph such that every two distinct nodes in the clique are adjacent. The \textbf{graph clique number} is the size of largest clique in the graph. \label{def:clique} 
\end{definition}
 
  \begin{definition}
The \textbf{efficiency} of a pair of nodes is defined as the reciprocal of the shortest path distance between the nodes. 
The \textbf{local efficiency} of a node in the graph is the average global efficiency of the subgraph induced by the neighbours of the node. 
The \textbf{average local efficiency}, used in this work, is the average of the local efficiency calculated for every node in the graph.
 \label{def:efficiency} 
\end{definition}
 
\section{Appendix: Loading conditions in bulk plasticity experiments}
The database used for causal discovery and training of the neural network forecast engines for numerical example in Section \ref{sec:bulkplasticity} 
includes 60 true triaxial numerical experiments conducted via the  YADE' DEM simulator. 
These experiments differ according to the applied axial strain rate $\dot{\epsilon}_{11}$, initial confining pressure $p_0$, initial void ratio $e_0$, and a parameter $b = \frac{\sigma_{22}-\sigma_{33}}{\sigma_{11}-\sigma_{33}}$ that controls applied stress conditions. In all 60 cases we set $\dot{\sigma}_{33}=\dot{\sigma}_{12}=\dot{\sigma}_{23}=\dot{\sigma}_{13}=0$. The setup of them are listed below. The tests with the bold font are the one discussed in Section \ref{sec:bulkplasticity}. The first 30 test (labelled T0-T29) are used to train the neural network, while T30-T59 are used for forward predictions. 

\begin{description}
	\item T0 $\dot{\epsilon}_{11}<0$, $b=0$, $p_0 = -300 kPa$, $e_0 = 0.539$.
	 \item T1  $\dot{\epsilon}_{11}<0$, $b=0$, $p_0 = -400 kPa$, $e_0 = 0.536$.
	 \item T2  $\dot{\epsilon}_{11}<0$, $b=0$, $p_0 = -500 kPa$, $e_0 = 0.534$.
	 \item T3  $\dot{\epsilon}_{11}>0$, $b=0$, $p_0 = -300 kPa$, $e_0 = 0.539$.
	 \item T4  $\dot{\epsilon}_{11}>0$, $b=0$, $p_0 = -400 kPa$, $e_0 = 0.536$.
	 \item T5  $\dot{\epsilon}_{11}>0$, $b=0$, $p_0 = -500 kPa$, $e_0 = 0.534$.
	 \item T6  $\dot{\epsilon}_{11}<0$, $b=0.5$, $p_0 = -300 kPa$, $e_0 = 0.539$.
	 \item T7  $\dot{\epsilon}_{11}<0$, $b=0.5$, $p_0 = -400 kPa$, $e_0 = 0.536$.
	 \item T8  $\dot{\epsilon}_{11}<0$, $b=0.5$, $p_0 = -500 kPa$, $e_0 = 0.534$.
	 \item T9  $\dot{\epsilon}_{11}>0$, $b=0.5$, $p_0 = -300 kPa$, $e_0 = 0.539$.
	 \item T10  $\dot{\epsilon}_{11}>0$, $b=0.5$, $p_0 = -400 kPa$, $e_0 = 0.536$.
	 \item T11  $\dot{\epsilon}_{11}>0$, $b=0.5$, $p_0 = -500 kPa$, $e_0 = 0.534$.
	 \item T12  $\dot{\epsilon}_{11}<0$, $b=0.1$, $p_0 = -300 kPa$, $e_0 = 0.539$.
	 \item T13  $\dot{\epsilon}_{11}<0$, $b=0.1$, $p_0 = -400 kPa$, $e_0 = 0.536$.
	 \item T14  $\dot{\epsilon}_{11}<0$, $b=0.1$, $p_0 = -500 kPa$, $e_0 = 0.534$.
	 \item T15  $\dot{\epsilon}_{11}>0$, $b=0.1$, $p_0 = -300 kPa$, $e_0 = 0.539$.
	 \item T16  $\dot{\epsilon}_{11}>0$, $b=0.1$, $p_0 = -400 kPa$, $e_0 = 0.536$.
	 \item T17  $\dot{\epsilon}_{11}>0$, $b=0.1$, $p_0 = -500 kPa$, $e_0 = 0.534$.
	 \item T18  $\dot{\epsilon}_{11}<0$, $b=0.25$, $p_0 = -300 kPa$, $e_0 = 0.539$.
	 \item T19  $\dot{\epsilon}_{11}<0$, $b=0.25$, $p_0 = -400 kPa$, $e_0 = 0.536$.
	 \item T20  $\dot{\epsilon}_{11}<0$, $b=0.25$, $p_0 = -500 kPa$, $e_0 = 0.534$.
	 \item T21  $\dot{\epsilon}_{11}>0$, $b=0.25$, $p_0 = -300 kPa$, $e_0 = 0.539$.
	 \item T22  $\dot{\epsilon}_{11}>0$, $b=0.25$, $p_0 = -400 kPa$, $e_0 = 0.536$.
	 \item \textbf{T23}  $\dot{\epsilon}_{11}>0$, $b=0.25$, $p_0 = -500 kPa$, $e_0 = 0.534$.
	 \item T24  $\dot{\epsilon}_{11}<0$, $b=0.75$, $p_0 = -300 kPa$, $e_0 = 0.539$.
	 \item T25  $\dot{\epsilon}_{11}<0$, $b=0.75$, $p_0 = -400 kPa$, $e_0 = 0.536$.
	 \item T26  $\dot{\epsilon}_{11}<0$, $b=0.75$, $p_0 = -500 kPa$, $e_0 = 0.534$.
	 \item T27  $\dot{\epsilon}_{11}>0$, $b=0.75$, $p_0 = -300 kPa$, $e_0 = 0.539$.
	 \item T28  $\dot{\epsilon}_{11}>0$, $b=0.75$, $p_0 = -400 kPa$, $e_0 = 0.536$.
	 \item \textbf{T29}  $\dot{\epsilon}_{11}>0$, $b=0.75$, $p_0 = -500 kPa$, $e_0 = 0.534$.
	 \item T30  $\dot{\epsilon}_{11}<0$, $b=0$, $p_0 = -350 kPa$, $e_0 = 0.539$.
	 \item T31  $\dot{\epsilon}_{11}<0$, $b=0$, $p_0 = -450 kPa$, $e_0 = 0.536$.
	 \item T32  $\dot{\epsilon}_{11}<0$, $b=0$, $p_0 = -550 kPa$, $e_0 = 0.534$.
	 \item T33  $\dot{\epsilon}_{11}>0$, $b=0$, $p_0 = -350 kPa$, $e_0 = 0.539$.
	 \item T34  $\dot{\epsilon}_{11}>0$, $b=0$, $p_0 = -450 kPa$, $e_0 = 0.536$.
	 \item T35  $\dot{\epsilon}_{11}>0$, $b=0$, $p_0 = -550 kPa$, $e_0 = 0.534$.
	 \item T36  $\dot{\epsilon}_{11}<0$, $b=0.5$, $p_0 = -350 kPa$, $e_0 = 0.539$.
	 \item T37  $\dot{\epsilon}_{11}<0$, $b=0.5$, $p_0 = -450 kPa$, $e_0 = 0.536$.
	 \item T38  $\dot{\epsilon}_{11}<0$, $b=0.5$, $p_0 = -550 kPa$, $e_0 = 0.534$.
	 \item T39  $\dot{\epsilon}_{11}>0$, $b=0.5$, $p_0 = -350 kPa$, $e_0 = 0.539$.
	 \item T40  $\dot{\epsilon}_{11}>0$, $b=0.5$, $p_0 = -450 kPa$, $e_0 = 0.536$.
	 \item T41  $\dot{\epsilon}_{11}>0$, $b=0.5$, $p_0 = -550 kPa$, $e_0 = 0.534$.
	 \item T42  $\dot{\epsilon}_{11}<0$, $b=0.1$, $p_0 = -350 kPa$, $e_0 = 0.539$.
	 \item T43  $\dot{\epsilon}_{11}<0$, $b=0.1$, $p_0 = -450 kPa$, $e_0 = 0.536$.
	 \item T44  $\dot{\epsilon}_{11}<0$, $b=0.1$, $p_0 = -550 kPa$, $e_0 = 0.534$.
	 \item T45  $\dot{\epsilon}_{11}>0$, $b=0.1$, $p_0 = -350 kPa$, $e_0 = 0.539$.
	 \item T46  $\dot{\epsilon}_{11}>0$, $b=0.1$, $p_0 = -450 kPa$, $e_0 = 0.536$.
	 \item T47  $\dot{\epsilon}_{11}>0$, $b=0.1$, $p_0 = -550 kPa$, $e_0 = 0.534$.
	 \item T48  $\dot{\epsilon}_{11}<0$, $b=0.25$, $p_0 = -350 kPa$, $e_0 = 0.539$.
	 \item T49  $\dot{\epsilon}_{11}<0$, $b=0.25$, $p_0 = -450 kPa$, $e_0 = 0.536$.
	 \item \textbf{T50}  $\dot{\epsilon}_{11}<0$, $b=0.25$, $p_0 = -550 kPa$, $e_0 = 0.534$.
	 \item T51  $\dot{\epsilon}_{11}>0$, $b=0.25$, $p_0 = -350 kPa$, $e_0 = 0.539$.
	 \item T52  $\dot{\epsilon}_{11}>0$, $b=0.25$, $p_0 = -450 kPa$, $e_0 = 0.536$.
	 \item T53  $\dot{\epsilon}_{11}>0$, $b=0.25$, $p_0 = -550 kPa$, $e_0 = 0.534$.
	 \item T54  $\dot{\epsilon}_{11}<0$, $b=0.75$, $p_0 = -350 kPa$, $e_0 = 0.539$.
	 \item T55  $\dot{\epsilon}_{11}<0$, $b=0.75$, $p_0 = -450 kPa$, $e_0 = 0.536$.
	 \item \textbf{T56}  $\dot{\epsilon}_{11}<0$, $b=0.75$, $p_0 = -550 kPa$, $e_0 = 0.534$.
	 \item T57  $\dot{\epsilon}_{11}>0$, $b=0.75$, $p_0 = -350 kPa$, $e_0 = 0.539$.
	 \item T58  $\dot{\epsilon}_{11}>0$, $b=0.75$, $p_0 = -450 kPa$, $e_0 = 0.536$.
	 \item T59  $\dot{\epsilon}_{11}>0$, $b=0.75$, $p_0 = -550 kPa$, $e_0 = 0.534$.	
\end{description}
}

\section{Appendix: Extension to lagged causal relationships}
\begin{figure}[h!]
	\centering
		\subfigure[Repetitive causal graph.]
{	\includegraphics[width=8cm, height=3.2cm]{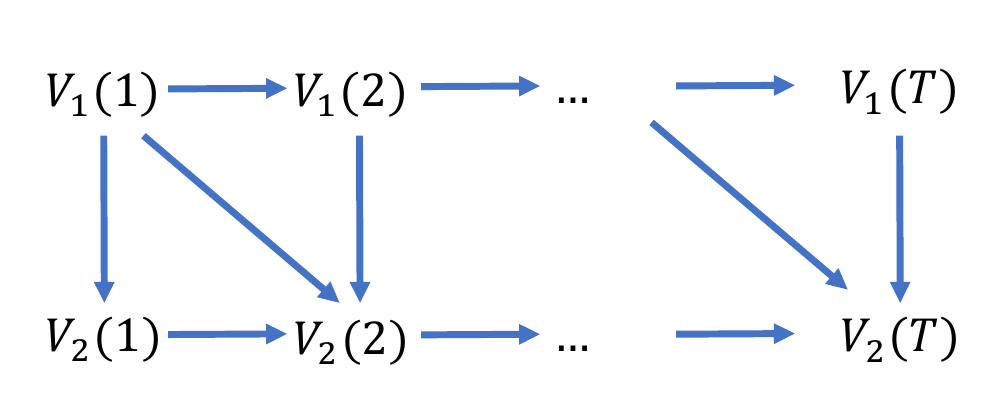}}
	\subfigure[Unit causal graph.]
{	\includegraphics[width=4cm, height=3cm]{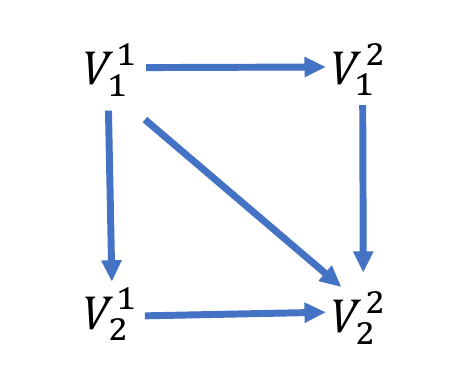}}
	\caption{An illustration of lagged causal relationships with two vertices in a DAG and the lag time being 1 ($L=1$).}
	\label{fig::appendix}
\end{figure}

In this section, we briefly discuss an extension of the proposal causal discovery approach to allow both  instantaneous and lagged causal relations. Without loss of generality, we use the example of traction-separation law to illustrate the method.  Recall that   $\mV_{-\vec{U}}$  includes all other variables in $\mV$ excluding $\vec{U}$ (e.g., porosity, fabric tensor). Assume that there are $m$ vertices in $\mV_{-\vec{U}}$, and we denote the values of vertices at the $t$th time point to be $\mV_{-\vec{U}}(t)=(V_1(t), \dots, V_m(t))$, where $t=1, \dots, T$. Since there exist both  instantaneous and lagged causal relations over these variables, we assume that the largest lag time to be $L$. Furthermore, we denote the $i$th variable from the $l$th time point to the $(T-L+l-1)$th time point to be 
$V_i^l=(V_i(l), V_i(l+1), \dots, V_i(T-L+l-1))$,   $i=1, \dots, m$ and $l=1, \dots, L+1$. 
Then we  introduce a new set of variables $\tilde{\mV}_{-\vec{U}}=\{\tilde{\mV}^1\}_{l=1}^{L+1}$, 
where $\tilde{\mV}^l  =\{V_1^l, V_2^l, \dots, V_m^l\}$. Fig. \ref{fig::appendix} illustrates the lagged causal relationships using a DAG with only two vertices. Fig. \ref{fig::appendix}(a) shows the repetitive causal graph over two time series $\mV(t)=(V_1(t), V_2(t))$, $t=1, \dots, T$ when the lag time $L=1$. Fig. \ref{fig::appendix}(b) shows the unit causal graph over the newly introduced variables $\tilde{\mV}=\tilde{\mV}^1 \cup \tilde{\mV}^2$. In this case, our goal is to not only recover the instantaneous causal relations between $V_1^2$ and $V_2^2$ as what did in the main manuscript, but also the lagged causal relations from $V_1^1$ to $V_2^2$, and from  $V_1^1$ to $V_1^2$.

Recall that $\vec{U}$ is the known input (i.e., displacement jump), and we have the prior knowledge that the dynamic changes in $\vec{U}$ can cause changes in other variables, not vice versa. Therefore, $\vec{U}$ can be used as a surrogate variable to help identify the causal relations. To recover the causal skeleton for both instantaneous and lagged causal relations, we modify  Algorithm 1 as follows. Firstly, a complete undirected graph $U_G$ is built with the variable  $\vec{U}$ and the newly introduced variables  $\tilde{\mV}_{-\vec{U}}$. Then for each $i$, we test for the marginal and conditional independence between $V_i^{L+1}$ and $\vec{U}$, $i=1, \dots, m$.  If they are independent, the edge between $V_i^{L+1}$ and $\vec{U}$ is removed; meanwhile the edge between $V_i^{l}$ and $\vec{U}$ is also removed, $l=1, \dots, L$. Next, we recover lagged causal relations by testing the marginal and conditional independence between the variables in $\tilde{\mV}^{L+1}$ and the variables in $\tilde{\mV}^{L-l+1}$ for each $l$th lagged relation, $l=1, \dots, L$. In particular, if $V_i^{L+1}$ and $V_j^{L-l+1}$ are independent, their edge is removed from $U_G$; meanwhile the edge between $V_i^{L-k+1}$ and $V_j^{L-l+1-k}$ is removed, $k=0, \dots, L-l$. Lastly, we recover the instantaneous causal relations by testing for the marginal 
and conditional independence between $V_i^{L+1}$  and $V_j^{L+1}$, $i\neq j$. If they are independent, their edge is removed; meanwhile the edge between $V_i^{k}$  and $V_j^{k}$ is removed, $k=1, \dots, L$. 
After the causal skeleton is determined, we can apply Algorithm 2 in the paper to recover the causal directions for instantaneous causal relations. For lagged ones,  they follow the rule that past causes future.

\end{appendix}

\section{Acknowledgments}
The authors are supported by 
by the NSF CAREER grant from Mechanics of Materials and Structures program
at National Science Foundation under grant contracts CMMI-1846875 and OAC-1940203, 
the Dynamic Materials and Interactions Program from the Air Force Office of Scientific 
Research under grant contracts FA9550-17-1-0169 and FA9550-19-1-0318.
These supports are gratefully acknowledged. 
The views and conclusions contained in this document are those of the authors, 
and should not be interpreted as representing the official policies, either expressed or implied, 
of the sponsors, including the Army Research Laboratory or the U.S. Government. 
The U.S. Government is authorized to reproduce and distribute reprints for 
Government purposes notwithstanding any copyright notation herein.

\bibliographystyle{plainnat}
\bibliography{main}

\end{document}